\documentclass{article}

\usepackage[main, final]{neurips_2026}

\usepackage[utf8]{inputenc}
\usepackage[T1]{fontenc}
\usepackage[colorlinks=true,
            allcolors=blue!55!black]{hyperref}

\usepackage{url}
\usepackage{booktabs}
\usepackage{amsfonts}
\usepackage{nicefrac}
\usepackage{microtype}
\usepackage[table]{xcolor}
\usepackage{xspace}
\usepackage{enumitem}
\usepackage{amsmath,amssymb}
\usepackage{multirow}
\usepackage{pifont}
\providecommand{\cmark}{\ding{51}}
\providecommand{\xmark}{\ding{55}}
\usepackage{cleveref}
\usepackage{algorithm}
\usepackage{algpseudocode}
\usepackage{graphicx}
\usepackage{adjustbox}
\usepackage{subcaption}
\usepackage{tabularx}
\usepackage{placeins}
\usepackage{tikz}
\usepackage{wrapfig}
\usetikzlibrary{arrows.meta,calc,positioning}
\usepackage{listings}

\lstdefinelanguage{Coq}{
  morekeywords={Module, Type, End, Definition, Fixpoint, Parameter, Axiom,
                Theorem, Lemma, Proof, Qed, Admitted, forall, fun, match,
                with, if, then, else, let, in, return, true, false, nil,
                Some, None, intros, unfold, simpl, reflexivity, induction,
                apply, rewrite, destruct, split, constructor, inversion,
                assumption, tauto, Require, Import, Inductive, Fixpoint},
  sensitive=true,
  morecomment=[s]{(*}{*)},
  morestring=[b]",
}
\usepackage{needspace}
\usepackage{etoolbox}
\BeforeBeginEnvironment{lstlisting}{\ifinner\else\par\needspace{3.5\baselineskip}\fi}

\providecommand{\aknew}[1]{#1}

\definecolor{revmaroon}{HTML}{800000}
\providecommand{\rev}[1]{#1}
\providecommand{\del}[1]{}


\definecolor{cldarkgreen}{rgb}{0.0, 0.6, 0.2}  %
\newif\ifcomments
\commentsfalse
\ifcomments
    \providecommand{\shubham}[1]{{\color{blue}{/* shubham: #1 */}}}
    \providecommand{\mert}[1]{{\color{olive}{/* mert: #1 */}}}
    \providecommand{\shu}[1]{{\color{magenta}{/* shu: #1 */}}}
    \providecommand{\ion}[1]{{\color{cyan}{/* ion: #1 */}}}
    \providecommand{\ak}[1]{{\color{orange}{/* alex: #1 */}}}
    
    \providecommand{\mohsen}[1]{{\color{orange}{/* mohsen: #1 */}}}
    \providecommand{\accheng}[1]{{\color{red}{/* accheng: #1 */}}}
    \providecommand{\cl}[1]{{\textcolor{cldarkgreen}{/* cl: #1 */}}}
\else
    \providecommand{\shubham}[1]{}
    \providecommand{\mert}[1]{}
    \providecommand{\shu}[1]{} 
    \providecommand{\ion}[1]{}
    \providecommand{\ak}[1]{}
    \providecommand{\mohsen}[1]{}
    \providecommand{\accheng}[1]{}
    \providecommand{\cl}[1]{}
\fi

\newcommand{\sys}{IDS\xspace}

\usepackage{tipa}

\newcommand{\coq}{Rocq\xspace}

\usepackage{xcolor}
\definecolor{softred}{HTML}{CC0000}
\definecolor{softgreen}{HTML}{2AAA2A}
\definecolor{softyellow}{HTML}{B8860B}

\definecolor{tropicalrainforest}{rgb}{0.0, 0.46, 0.37}

\definecolor{fignumcolor}{HTML}{3B78D9}
\DeclareRobustCommand{\fignum}[1]{\tikz[baseline=(c.base)]{\node[circle, fill=fignumcolor, text=white, inner sep=0.4pt, minimum size=8pt, font=\sffamily\bfseries\tiny] (c) {#1};}}

\crefformat{section}{\S#2#1#3}
\crefformat{subsection}{\S#2#1#3}
\crefformat{subsubsection}{\S#2#1#3}
\Crefformat{section}{\S#2#1#3}
\Crefformat{subsection}{\S#2#1#3}
\Crefformat{subsubsection}{\S#2#1#3}
\crefrangeformat{section}{\S#3#1#4--\S#5#2#6}
\crefrangeformat{subsection}{\S#3#1#4--\S#5#2#6}
\crefmultiformat{section}{\S\S#2#1#3}{ and~#2#1#3}{, #2#1#3}{, and~#2#1#3}
\crefmultiformat{subsection}{\S\S#2#1#3}{ and~#2#1#3}{, #2#1#3}{, and~#2#1#3}

\crefformat{appendix}{App.~#2#1#3}
\Crefformat{appendix}{App.~#2#1#3}
\crefrangeformat{appendix}{App.~#3#1#4--#5#2#6}
\crefmultiformat{appendix}{App.~#2#1#3}{ and~#2#1#3}{, #2#1#3}{, and~#2#1#3}
\crefformat{subappendix}{App.~#2#1#3}
\Crefformat{subappendix}{App.~#2#1#3}
\crefformat{subsubappendix}{App.~#2#1#3}
\crefmultiformat{subappendix}{App.~#2#1#3}{ and~#2#1#3}{, #2#1#3}{, and~#2#1#3}
\crefrangeformat{subappendix}{App.~#3#1#4--#5#2#6}
\Crefformat{subsubappendix}{App.~#2#1#3}

\crefformat{figure}{Fig.~#2#1#3}
\Crefformat{figure}{Fig.~#2#1#3}
\crefrangeformat{figure}{Figs.~#3#1#4--#5#2#6}
\crefmultiformat{figure}{Figs.~#2#1#3}{ and~#2#1#3}{, #2#1#3}{ and~#2#1#3}
\crefformat{table}{Table~#2#1#3}
\Crefformat{table}{Table~#2#1#3}
\crefrangeformat{table}{Tables~#3#1#4--#5#2#6}
\crefmultiformat{table}{Tables~#2#1#3}{ and~#2#1#3}{, #2#1#3}{ and~#2#1#3}

\newcommand{\cut}[1]{\unskip}

\usepackage{mathpartir}

\usepackage{graphicx}

\usepackage{amsfonts}
 \usepackage{amsthm} 
\theoremstyle{definition}  %
\newtheorem{definition}{Definition}

\usepackage{xcolor}
\usepackage[normalem]{ulem}
\usepackage{cancel}
\usepackage{url}
\usepackage{mmm}

\usepackage{amssymb}
\usepackage{amsmath}

\usepackage{microtype}

\def\t{\phantom{X}}

\def\<{\langle}
\def\>{\rangle}

\usepackage{listings}
\usepackage{array}

\usepackage{tikz}
\usetikzlibrary{positioning} 
\usetikzlibrary{arrows} 
\usetikzlibrary{calc}
\usetikzlibrary{backgrounds}
\usetikzlibrary{shapes}

\usepackage{stmaryrd}

\def\defeq{\triangleq}

\def\self{\mathit{self}}

\def\ret{\mathsf{ret} \ }
\def\lforall{\mathsf{forall} \ }
\def\lif{\mathsf{if} \ }
\def\lelse{\mathsf{else} \ }

\def\llet{\mathsf{let} \ }
\def\max{\mathsf{max}}

\def\List{\mathsf{List}}
\def\llookup{\mathsf{lookup} \ }

\def\RYW{\mathit{RYW}}
\def\MR{\mathit{MR}}

\def\MW{\mathit{MW}}
\def\CC{\mathit{CC}}
\def\LCC{\mathit{LCC}}
\def\Rel{\mathit{Rel}}

\def\ext{\mathsf{ext}}

\def\CState{\mathsf{CState}}
\def\RState{\mathsf{RState}}

\def\true{\mathsf{true}}
\def\Set{\mathsf{Set}}
\def\Nat{\mathsf{Nat}}
\def\L{L}
\def\lab{\mathit{label}}
\def\clientLabs{\mathit{client\text{-}labels}}

\def\cInit{\mathsf{c\text{-}init}}
\def\rInit{\mathsf{r\text{-}init}}

\def\get{\mathsf{get}}
\def\getGuard{\mathsf{get\text{-}guard}}
\def\getReq{\mathsf{get\text{-}req}}
\def\getRes{\mathsf{get\text{-}res}}

\def\lput{\mathsf{put}}
\def\putGuard{\mathsf{put\text{-}guard}}
\def\putReq{\mathsf{put\text{-}req}}

\def\Type{\mathsf{Type}}

\def\Bool{\mathsf{Bool}}

\def\PutReq{\mathit{Put\text{-}req}}
\def\PutReqPayload{\mathsf{Put\text{-}req\text{-}payload}}

\def\GetReq{\mathit{Get\text{-}req}}
\def\GetReqPayload{\mathsf{Get\text{-}req\text{-}payload}}
\def\GetRes{\mathit{Get\text{-}res}}
\def\GetResPayload{\mathsf{Get\text{-}res\text{-}payload}}

\def\Unit{\mathsf{Unit}}

\newcommand\sPut[3][]{%
  \mathit{put}%
  \ifthenelse{\equal{#1}{}}{%
    \ifthenelse{\equal{#2}{}}{}{
    ({#2},{#3})}%
  }{%
    ^{#1}({#2},{#3})%
  }%
}  
  
\newcommand\sGet[3][]{%
   \ifthenelse{\equal{#2}{}}{}{#2\leftarrow}%
   \mathit{get}
   \ifthenelse{\equal{#1}{}}{}{^{#1}}%
   \ifthenelse{\equal{#3}{}}{}{({#3})}}

  \newcommand\sSkip{\mathit{skip}}

\newcommand{\name}[1]{\textsc{#1}}

\newcommand\indrule[3][]{%
  \inferrule[#1]{#2}{#3}%
  }

\newcommand\figurealgtextsize{%
  \setlength{\abovedisplayskip}{0pt}%
  \setlength{\abovedisplayshortskip}{0pt}%
  \setlength{\belowdisplayskip}{0pt}%
  \setlength{\belowdisplayshortskip}{0pt}}
\newcommand\figureruletextsize{%
  \setlength{\abovedisplayskip}{0pt}%
  \setlength{\abovedisplayshortskip}{0pt}%
  \setlength{\belowdisplayskip}{0pt}%
  \setlength{\belowdisplayshortskip}{0pt}
}
\newcommand\figuredefstextsize{%
  \setlength{\abovedisplayskip}{0pt}%
  \setlength{\abovedisplayshortskip}{0pt}%
  \setlength{\belowdisplayskip}{0pt}%
  \setlength{\belowdisplayshortskip}{0pt}
}

\usepackage{siunitx}  %

\usepackage{balance}

\newcommand\world[3]{%
   {\begin{array}{l} ( %
      #1, \\ \phantom{(}
      #2, \\ \phantom{(}
      #3          
   ) \end{array}}
}  

\newcommand\note[1]{\text{\small#1}}
\newcommand\bnote[1]{\textbf{\small#1}}

\usepackage{tikz}
\usetikzlibrary{positioning, arrows.meta}

\DeclareMathAlphabet{\mathpzc}{OT1}{pzc}{m}{it}

\DeclareUnicodeCharacter{00AC}{\ensuremath{\neg}}                     %
\DeclareUnicodeCharacter{00B1}{\ensuremath{\pm}}                      %
\DeclareUnicodeCharacter{00B2}{\ensuremath{^2}}                       %
\DeclareUnicodeCharacter{00B3}{\ensuremath{^3}}                       %
\DeclareUnicodeCharacter{00B7}{\ensuremath{\cdot}}                    %
\DeclareUnicodeCharacter{00B9}{\ensuremath{^1}}                       %
\DeclareUnicodeCharacter{00D7}{\ensuremath{\times}}                   %
\DeclareUnicodeCharacter{02B0}{\ensuremath{^h}}                       %
\DeclareUnicodeCharacter{02B3}{\ensuremath{^r}}                       %
\DeclareUnicodeCharacter{02E2}{\ensuremath{^s}}                       %
\DeclareUnicodeCharacter{20F0}{\ensuremath{^*}}                       %
\DeclareUnicodeCharacter{0302}{\ensuremath{\hat{\phantom{x}}}}        %
\DeclareUnicodeCharacter{0332}{\ensuremath{\underline{\phantom{x}}}}  %
\DeclareUnicodeCharacter{0308}{\ensuremath{\ddot{\phantom{x}}}}       %
\DeclareUnicodeCharacter{0393}{\ensuremath{\Gamma}}                   %
\DeclareUnicodeCharacter{0394}{\ensuremath{\Delta}}                   %
\DeclareUnicodeCharacter{0398}{\ensuremath{\Theta}}                   %
\DeclareUnicodeCharacter{039B}{\ensuremath{\Lambda}}                  %
\DeclareUnicodeCharacter{039E}{\ensuremath{\Xi}}                      %
\DeclareUnicodeCharacter{03A0}{\ensuremath{\Pi}}                      %
\DeclareUnicodeCharacter{03A3}{\ensuremath{\Sigma}}                   %

\DeclareUnicodeCharacter{03A4}{\ensuremath{\mathrm{T}}}  %

\DeclareUnicodeCharacter{03A5}{\ensuremath{\Upsilon}}   %
\DeclareUnicodeCharacter{03A6}{\ensuremath{\Phi}}       %
\DeclareUnicodeCharacter{03A8}{\ensuremath{\Psi}}       %
\DeclareUnicodeCharacter{03A9}{\ensuremath{\Omega}}     %
\DeclareUnicodeCharacter{03B1}{\ensuremath{\alpha}}     %
\DeclareUnicodeCharacter{03B2}{\ensuremath{\beta}}      %
\DeclareUnicodeCharacter{03B3}{\ensuremath{\gamma}}     %
\DeclareUnicodeCharacter{03B4}{\ensuremath{\delta}}     %
\DeclareUnicodeCharacter{03B5}{\ensuremath{\epsilon}}   %
\DeclareUnicodeCharacter{03B6}{\ensuremath{\zeta}}      %
\DeclareUnicodeCharacter{03B7}{\ensuremath{\eta}}       %
\DeclareUnicodeCharacter{03B8}{\ensuremath{\theta}}     %
\DeclareUnicodeCharacter{03B9}{\ensuremath{\iota}}      %
\DeclareUnicodeCharacter{03BA}{\ensuremath{\kappa}}     %
\DeclareUnicodeCharacter{03BB}{\ensuremath{\lambda}}    %
\DeclareUnicodeCharacter{03BC}{\ensuremath{\mu}}        %
\DeclareUnicodeCharacter{03BD}{\ensuremath{\nu}}        %
\DeclareUnicodeCharacter{03BE}{\ensuremath{\xi}}        %
\DeclareUnicodeCharacter{03C0}{\ensuremath{\pi}}        %
\DeclareUnicodeCharacter{03C1}{\ensuremath{\rho}}       %
\DeclareUnicodeCharacter{03C2}{\ensuremath{\varsigma}}  %
\DeclareUnicodeCharacter{03C3}{\ensuremath{\sigma}}     %
\DeclareUnicodeCharacter{03C4}{\ensuremath{\tau}}       %
\DeclareUnicodeCharacter{03C5}{\ensuremath{\upsilon}}   %
\DeclareUnicodeCharacter{03C6}{\ensuremath{\varphi}}    %
\DeclareUnicodeCharacter{03C7}{\ensuremath{\chi}}       %
\DeclareUnicodeCharacter{03C8}{\ensuremath{\psi}}       %
\DeclareUnicodeCharacter{03C9}{\ensuremath{\omega}}     %
\DeclareUnicodeCharacter{03D1}{\ensuremath{\vartheta}}  %

\DeclareUnicodeCharacter{0432}{\ensuremath{\mathrm{PDF\;TODO}}}  %
\DeclareUnicodeCharacter{0435}{\ensuremath{\mathrm{PDF\;TODO}}}  %
\DeclareUnicodeCharacter{0438}{\ensuremath{\mathrm{PDF\;TODO}}}  %
\DeclareUnicodeCharacter{043C}{\ensuremath{\mathrm{PDF\;TODO}}}  %
\DeclareUnicodeCharacter{0440}{\ensuremath{\mathrm{PDF\;TODO}}}  %
\DeclareUnicodeCharacter{0442}{\ensuremath{\mathrm{PDF\;TODO}}}  %
\DeclareUnicodeCharacter{041F}{\ensuremath{\mathrm{PDF\;TODO}}}  %
\DeclareUnicodeCharacter{0BA8}{\ensuremath{\mathrm{PDF\;TODO}}}  %
\DeclareUnicodeCharacter{0BBF}{\ensuremath{\mathrm{PDF\;TODO}}}  %
\DeclareUnicodeCharacter{1100}{\ensuremath{\mathrm{PDF\;TODO}}}  %
\DeclareUnicodeCharacter{11F9}{\ensuremath{\mathrm{PDF\;TODO}}}  %

\DeclareUnicodeCharacter{1D48}{\ensuremath{^d}}  %
\DeclareUnicodeCharacter{1D50}{\ensuremath{^m}}  %
\DeclareUnicodeCharacter{1D56}{\ensuremath{^p}}  %
\DeclareUnicodeCharacter{1D62}{\ensuremath{_i}}  %
\DeclareUnicodeCharacter{1D63}{\ensuremath{_r}}  %
\DeclareUnicodeCharacter{1D64}{\ensuremath{_u}}  %
\DeclareUnicodeCharacter{1D9C}{\ensuremath{^c}}  %

\DeclareUnicodeCharacter{2032}{\ensuremath{\text{\textasciiacute}}}  %

\DeclareUnicodeCharacter{2070}{\ensuremath{^0}}  %
\DeclareUnicodeCharacter{2074}{\ensuremath{^4}}  %
\DeclareUnicodeCharacter{2075}{\ensuremath{^5}}  %
\DeclareUnicodeCharacter{2076}{\ensuremath{^6}}  %
\DeclareUnicodeCharacter{2077}{\ensuremath{^7}}  %
\DeclareUnicodeCharacter{2078}{\ensuremath{^8}}  %
\DeclareUnicodeCharacter{2079}{\ensuremath{^9}}  %
\DeclareUnicodeCharacter{207A}{\ensuremath{^+}}  %
\DeclareUnicodeCharacter{207B}{\ensuremath{^-}}  %
\DeclareUnicodeCharacter{207F}{\ensuremath{^n}}  %
\DeclareUnicodeCharacter{2080}{\ensuremath{_0}}  %
\DeclareUnicodeCharacter{2081}{\ensuremath{_1}}  %
\DeclareUnicodeCharacter{2082}{\ensuremath{_2}}  %
\DeclareUnicodeCharacter{2083}{\ensuremath{_3}}  %
\DeclareUnicodeCharacter{2084}{\ensuremath{_4}}  %
\DeclareUnicodeCharacter{2085}{\ensuremath{_5}}  %
\DeclareUnicodeCharacter{2086}{\ensuremath{_6}}  %
\DeclareUnicodeCharacter{2087}{\ensuremath{_7}}  %
\DeclareUnicodeCharacter{2088}{\ensuremath{_8}}  %
\DeclareUnicodeCharacter{2089}{\ensuremath{_9}}  %
\DeclareUnicodeCharacter{208A}{\ensuremath{_+}}  %
\DeclareUnicodeCharacter{208B}{\ensuremath{_-}}  %
\DeclareUnicodeCharacter{2090}{\ensuremath{_a}}  %
\DeclareUnicodeCharacter{2091}{\ensuremath{_e}}  %
\DeclareUnicodeCharacter{2095}{\ensuremath{_h}}  %
\DeclareUnicodeCharacter{2096}{\ensuremath{_k}}  %
\DeclareUnicodeCharacter{2097}{\ensuremath{_l}}  %
\DeclareUnicodeCharacter{2098}{\ensuremath{_m}}  %
\DeclareUnicodeCharacter{2099}{\ensuremath{_n}}  %
\DeclareUnicodeCharacter{209A}{\ensuremath{_p}}  %
\DeclareUnicodeCharacter{2092}{\ensuremath{_o}}  %
\DeclareUnicodeCharacter{209B}{\ensuremath{_s}}  %
\DeclareUnicodeCharacter{209C}{\ensuremath{_t}}  %
\DeclareUnicodeCharacter{2093}{\ensuremath{_x}}  %

\DeclareUnicodeCharacter{2113}{\ensuremath{\ell}} %

\DeclareUnicodeCharacter{2115}{\ensuremath{\mathbbm{N}}}  %
\DeclareUnicodeCharacter{211A}{\ensuremath{\mathbbm{Q}}}  %
\DeclareUnicodeCharacter{211D}{\ensuremath{\mathbbm{R}}}  %
\DeclareUnicodeCharacter{2124}{\ensuremath{\mathbbm{Z}}}  %

\DeclareUnicodeCharacter{2135}{\ensuremath{\aleph}}           %
\DeclareUnicodeCharacter{2190}{\ensuremath{\leftarrow}}       %
\DeclareUnicodeCharacter{2192}{\ensuremath{\rightarrow}}      %
\DeclareUnicodeCharacter{2194}{\ensuremath{\leftrightarrow}}  %

\DeclareUnicodeCharacter{21A0}{\ensuremath{\twoheadrightarrow}}  %

\DeclareUnicodeCharacter{21A6}{\ensuremath{\mapsto}}  %
\DeclareUnicodeCharacter{21A6}{\ensuremath{\mapsto}}  %

\DeclareUnicodeCharacter{21BE}{\ensuremath{\restriction}}  %

\DeclareUnicodeCharacter{21D0}{\ensuremath{\Leftarrow}}       %
\DeclareUnicodeCharacter{21D2}{\ensuremath{\Rightarrow}}      %
\DeclareUnicodeCharacter{21D4}{\ensuremath{\Leftrightarrow}}  %

\DeclareUnicodeCharacter{2200}{\ensuremath{\forall}}      %
\DeclareUnicodeCharacter{2203}{\ensuremath{\exists}}      %
\DeclareUnicodeCharacter{2204}{\ensuremath{\not\exists}}  %
\DeclareUnicodeCharacter{2205}{\ensuremath{\emptyset}}    %
\DeclareUnicodeCharacter{2208}{\ensuremath{\in}}          %
\DeclareUnicodeCharacter{2209}{\ensuremath{\not\in}}      %
\DeclareUnicodeCharacter{220B}{\ensuremath{\ni}}          %
\DeclareUnicodeCharacter{220C}{\ensuremath{\not\ni}}      %

\DeclareUnicodeCharacter{220E}{{\tiny \ensuremath{\blacksquare}}} %

\DeclareUnicodeCharacter{220F}{{\tiny \ensuremath{\prod}}}  %
\DeclareUnicodeCharacter{2211}{\ensuremath{\sum}}           %
\DeclareUnicodeCharacter{25CB}{\ensuremath{\circ}}          %
\DeclareUnicodeCharacter{2022}{\ensuremath{\bullet}}        %
\DeclareUnicodeCharacter{221E}{\ensuremath{\infty}}         %
\DeclareUnicodeCharacter{2223}{\ensuremath{\mid}}           %
\DeclareUnicodeCharacter{2225}{\ensuremath{\parallel}}      %
\DeclareUnicodeCharacter{2227}{\ensuremath{\wedge}}         %
\DeclareUnicodeCharacter{2228}{\ensuremath{\vee}}           %

\DeclareUnicodeCharacter{2229}{\ensuremath{\cap}}           %

\DeclareUnicodeCharacter{222A}{\ensuremath{\cup}}           %
\DeclareUnicodeCharacter{222B}{\ensuremath{\int}}           %
\DeclareUnicodeCharacter{2234}{\ensuremath{\therefore}}     %

\DeclareUnicodeCharacter{2237}{\ensuremath{\dblcolon}} %

\newcommand{\dotminus}{\mathop{\mbox{$-^{\hspace{-.5em}\cdot}\,$}}}
\DeclareUnicodeCharacter{2238}{\ensuremath{\dotminus}}  %

\DeclareUnicodeCharacter{223C}{\ensuremath{\sim}}       %
\DeclareUnicodeCharacter{2243}{\ensuremath{\simeq}}     %
\DeclareUnicodeCharacter{2245}{\ensuremath{\cong}}      %
\DeclareUnicodeCharacter{2248}{\ensuremath{\approx}}    %
\DeclareUnicodeCharacter{2250}{\ensuremath{\doteq}}     %

\usepackage{mathtools}
\DeclareUnicodeCharacter{2254}{\ensuremath{\coloneqq}}  %

\newcommand{\eqq}{\stackrel{\text{\tiny ?}}{=}}
\DeclareUnicodeCharacter{225F}{\ensuremath{\eqq}}  %

\DeclareUnicodeCharacter{2260}{\ensuremath{\neq}}         %
\DeclareUnicodeCharacter{2261}{\ensuremath{\equiv}}       %
\DeclareUnicodeCharacter{2262}{\ensuremath{\not\equiv}}   %
\DeclareUnicodeCharacter{2264}{\ensuremath{\le}}          %
\DeclareUnicodeCharacter{2265}{\ensuremath{\ge}}          %
\DeclareUnicodeCharacter{226E}{\ensuremath{\nless}}       %
\DeclareUnicodeCharacter{226F}{\ensuremath{\ngtr}}        %
\DeclareUnicodeCharacter{2270}{\ensuremath{\nleq}}        %
\DeclareUnicodeCharacter{2271}{\ensuremath{\ngeq}}        %
\DeclareUnicodeCharacter{227A}{\ensuremath{\prec}}        %
\DeclareUnicodeCharacter{2280}{\ensuremath{\nprec}}        %
\DeclareUnicodeCharacter{227C}{\ensuremath{\preceq}}      %
\DeclareUnicodeCharacter{2282}{\ensuremath{\subset}}      %
\DeclareUnicodeCharacter{2284}{\ensuremath{\not\subset}} %
\DeclareUnicodeCharacter{2283}{\ensuremath{\supset}}      %
\DeclareUnicodeCharacter{2286}{\ensuremath{\subseteq}}    %
\DeclareUnicodeCharacter{228E}{\ensuremath{\uplus}}       %
\DeclareUnicodeCharacter{2291}{\ensuremath{\sqsubseteq}}  %
\DeclareUnicodeCharacter{228F}{\ensuremath{\sqsubset}}  %
\DeclareUnicodeCharacter{2293}{\ensuremath{\sqcap}}       %
\DeclareUnicodeCharacter{2294}{\ensuremath{\sqcup}}       %
\DeclareUnicodeCharacter{2295}{\ensuremath{\oplus}}       %
\DeclareUnicodeCharacter{22A2}{\ensuremath{\vdash}}       %
\DeclareUnicodeCharacter{22A4}{\ensuremath{\top}}         %
\DeclareUnicodeCharacter{22A5}{\ensuremath{\bot}}         %
\DeclareUnicodeCharacter{22C0}{\ensuremath{\bigwedge}}    %
\DeclareUnicodeCharacter{22C2}{\ensuremath{\bigcap}}      %
\DeclareUnicodeCharacter{22C3}{\ensuremath{\bigcup}}      %
\DeclareUnicodeCharacter{22EE}{\ensuremath{\vdots}}       %
\DeclareUnicodeCharacter{22EF}{\ensuremath{\cdots}}       %

\DeclareUnicodeCharacter{22F0}{\ensuremath{\iddots}} %

\DeclareUnicodeCharacter{230A}{\ensuremath{\lfloor}}  %
\DeclareUnicodeCharacter{230B}{\ensuremath{\rfloor}}  %
\DeclareUnicodeCharacter{2308}{\ensuremath{\lceil}}   %
\DeclareUnicodeCharacter{2309}{\ensuremath{\rceil}}   %

\DeclareUnicodeCharacter{2329}{\ensuremath{\langle}}  %
\DeclareUnicodeCharacter{232A}{\ensuremath{\rangle}}  %

\DeclareUnicodeCharacter{23CE}{\ensuremath{\Return}}  %

\DeclareUnicodeCharacter{2460}{\ensuremath{\text{\ding{172}}}} %
\DeclareUnicodeCharacter{2461}{\ensuremath{\text{\ding{173}}}} %
\DeclareUnicodeCharacter{2462}{\ensuremath{\text{\ding{174}}}} %
\DeclareUnicodeCharacter{2463}{\ensuremath{\text{\ding{175}}}} %
\DeclareUnicodeCharacter{2464}{\ensuremath{\text{\ding{176}}}} %
\DeclareUnicodeCharacter{2465}{\ensuremath{\text{\ding{177}}}} %
\DeclareUnicodeCharacter{2466}{\ensuremath{\text{\ding{178}}}} %
\DeclareUnicodeCharacter{2467}{\ensuremath{\text{\ding{179}}}} %
\DeclareUnicodeCharacter{2468}{\ensuremath{\text{\ding{180}}}} %

\DeclareUnicodeCharacter{25C5}{\ensuremath{\triangleleft}}  %

\DeclareUnicodeCharacter{2660}{\ensuremath{\spadesuit}} %
\DeclareUnicodeCharacter{266D}{\ensuremath{\flat}}      %
\DeclareUnicodeCharacter{266F}{\ensuremath{\sharp}}     %

\DeclareUnicodeCharacter{1D4D0}{\ensuremath{\mathcal{A}}} 
\DeclareUnicodeCharacter{1D4D1}{\ensuremath{\mathcal{B}}}  
\DeclareUnicodeCharacter{1D4D2}{\ensuremath{\mathcal{C}}}  
\DeclareUnicodeCharacter{1D4D3}{\ensuremath{\mathcal{D}}}  
\DeclareUnicodeCharacter{1D4D4}{\ensuremath{\mathcal{E}}}  
\DeclareUnicodeCharacter{1D4D5}{\ensuremath{\mathcal{F}}}  
\DeclareUnicodeCharacter{1D4D6}{\ensuremath{\mathcal{G}}}  
\DeclareUnicodeCharacter{1D4D7}{\ensuremath{\mathcal{H}}}  
\DeclareUnicodeCharacter{1D4D8}{\ensuremath{\mathcal{I}}}  
\DeclareUnicodeCharacter{1D4D9}{\ensuremath{\mathcal{J}}}  
\DeclareUnicodeCharacter{1D4DA}{\ensuremath{\mathcal{K}}}  
\DeclareUnicodeCharacter{1D4DB}{\ensuremath{\mathcal{L}}}  
\DeclareUnicodeCharacter{1D4DC}{\ensuremath{\mathcal{M}}}  
\DeclareUnicodeCharacter{1D4DD}{\ensuremath{\mathcal{N}}}  
\DeclareUnicodeCharacter{1D4DE}{\ensuremath{\mathcal{O}}}  
\DeclareUnicodeCharacter{1D4DF}{\ensuremath{\mathcal{P}}}

\DeclareUnicodeCharacter{1D4E0}{\ensuremath{\mathcal{Q}}}  
\DeclareUnicodeCharacter{1D4E1}{\ensuremath{\mathcal{R}}}  
\DeclareUnicodeCharacter{1D4E2}{\ensuremath{\mathcal{S}}}  
\DeclareUnicodeCharacter{1D4E3}{\ensuremath{\mathcal{T}}}  
\DeclareUnicodeCharacter{1D4E4}{\ensuremath{\mathcal{U}}}  
\DeclareUnicodeCharacter{1D4E5}{\ensuremath{\mathcal{V}}}  
\DeclareUnicodeCharacter{1D4E6}{\ensuremath{\mathcal{W}}}  
\DeclareUnicodeCharacter{1D4E7}{\ensuremath{\mathcal{X}}}  
\DeclareUnicodeCharacter{1D4E8}{\ensuremath{\mathcal{Y}}}  
\DeclareUnicodeCharacter{1D4E9}{\ensuremath{\mathcal{Z}}}

\DeclareUnicodeCharacter{2713}{\ensuremath{\checkmark}}  %

\DeclareUnicodeCharacter{27E6}{\ensuremath{\llbracket}}  %
\DeclareUnicodeCharacter{27E7}{\ensuremath{\rrbracket}}  %

\DeclareUnicodeCharacter{27E8}{\ensuremath{\langle}}          %
\DeclareUnicodeCharacter{27E9}{\ensuremath{\rangle}}          %
\DeclareUnicodeCharacter{27F6}{\ensuremath{\longrightarrow}}  %

\DeclareUnicodeCharacter{2983}{\ensuremath{\mathrm{PDF\;TODO}}}  %
\DeclareUnicodeCharacter{2984}{\ensuremath{\mathrm{PDF\;TODO}}}  %
\DeclareUnicodeCharacter{2985}{\ensuremath{\mathrm{PDF\;TODO}}}  %
\DeclareUnicodeCharacter{2986}{\ensuremath{\mathrm{PDF\;TODO}}}  %

\DeclareUnicodeCharacter{2987}{\ensuremath{\llparenthesis}}  %
\DeclareUnicodeCharacter{2988}{\ensuremath{\rrparenthesis}}  %

\DeclareUnicodeCharacter{2216}{\ensuremath{\setminus}}

\DeclareUnicodeCharacter{2A06}{\ensuremath{\bigcup}}  %

\DeclareUnicodeCharacter{D7B0}{\ensuremath{\mathrm{PDF\;TODO}}} %

\DeclareUnicodeCharacter{1D7D9}{\ensuremath{\mathbbm{1}}}  %

\DeclareUnicodeCharacter{25C2}{\ensuremath{\triangleleft}}  
\DeclareUnicodeCharacter{25B9}{\ensuremath{\triangleright}}
\DeclareUnicodeCharacter{2286}{\ensuremath{\subseteq}}    %
\DeclareUnicodeCharacter{2217}{\em}

\DeclareUnicodeCharacter{1D62}{\ensuremath{_i}}  
\DeclareUnicodeCharacter{2C7C}{\ensuremath{_j}}
\DeclareUnicodeCharacter{2096}{\ensuremath{_k}}

\DeclareUnicodeCharacter{276A}{\mbox{$($}}  
\DeclareUnicodeCharacter{276B}{\mbox{$)$}}

\DeclareUnicodeCharacter{261E}{\todo}      %

\DeclareUnicodeCharacter{1D539}{\mathbb{B}}      %
\DeclareUnicodeCharacter{2102}{\mathbb{C}}      %
\DeclareUnicodeCharacter{2098}{_m}      %
\DeclareUnicodeCharacter{2115}{\mathbb{N}}      %

\DeclareUnicodeCharacter{2713}{\ensuremath{\checkmark}}  %

\DeclareUnicodeCharacter{25CB}{\ensuremath{\circ}}        %
\DeclareUnicodeCharacter{2A1D}{\ensuremath{\bowtie}}        %
\DeclareUnicodeCharacter{22C8}{\ensuremath{\bowtie}}        %

\DeclareUnicodeCharacter{2287}{\ensuremath{\supseteq}}        %

\DeclareUnicodeCharacter{21DD}{\ensuremath{\rightsquigarrow}}        %

\DeclareUnicodeCharacter{1D408}{\ensuremath{\mathbb{I}}} %

\DeclareUnicodeCharacter{2217}{\ensuremath{\ast}}      %

\DeclareUnicodeCharacter{2298}{\ensuremath{\oslash}}      %

\title{Inductive Deductive Synthesis: \\Co-Evolving Code-and-Proof Generation}

\author{%
  Shubham Agarwal$^{*,1}$, Alexander Krentsel$^{*,1}$, Shu Liu$^{*,1}$, Mert Cemri$^{*,1}$,\\ \bfseries Audrey Cheng$^{1}$,
  Rui Meng$^{2}$, Tomas Pfister$^{2}$, Chun-Liang Li$^{2}$, Sylvia Ratnasamy$^{1}$, \\ \bfseries Aditya Parameswaran$^{1}$,
  Matei Zaharia$^{1}$, Ion Stoica$^{1}$, Mohsen Lesani$^{3}$ \\[0.3em]
  $^{1}$UC Berkeley \quad $^{2}$Google \quad $^{3}$UC Santa Cruz
}

\begin{document}

\maketitle

{\renewcommand{\thefootnote}{$\ast$}\footnotetext{Equal contribution.\quad\textsuperscript{\normalfont$\dagger$}Code available at \url{https://github.com/skydiscover-ai/skydiscover}.}}

\begin{abstract}
AI agents increasingly excel at generating, testing, and refining code.
However, they fall short on tasks requiring formal guarantees of full coverage that testing alone cannot provide.
Distributed systems are a prime example: properties such as consistency between reads and writes must hold under every possible interleaving of events.
Mechanized formal verification can guarantee such correctness, but typically demands months to years of expert effort.
As evidence, even SOTA coding agents (Codex with GPT-5.4 and Claude Code with Opus 4.6) succeed on only 2/7 distributed key-value-store specifications.
In this paper, we present the first effective approach to addressing this gap, Inductive Deductive Synthesis (\sys), which \del{jointly and }incrementally \rev{co-evolves}\del{synthesizes} implementation and proof, and learns from failed attempts to systematically try promising strategies.
Built as an agentic LLM system, \sys achieves 7/7 in about \rev{$6.9$}\del{$6.8$} hours and \$$106$ per spec on average, roughly $200\times$ faster than expert effort and $17\%$ cheaper than SOTA agents.
\sys further incorporates performance feedback into the same loop, yielding implementations up to $3\times$ faster than published verified systems.\textsuperscript{\normalfont$\dagger$}
\end{abstract}

\section{Introduction}
\label{sec:intro}
Coding agents have made great strides at writing code for many tasks, even running tests to automatically refine their own output~\cite{cursor, claude-code, codex, swe-agent, reflexion}. However, they fall short on tasks requiring correctness \textit{guarantees} across every possible behavior, rather than just the cases reached during testing.
Distributed systems are a prime example: properties such as consistency between subsequent writes and reads must hold under every possible interleaving of messages, failures, and concurrent updates, a combinatorially large space that no realistic test suite can cover~\cite{taxdc}. The consequences of incorrect behavior are severe, ranging from a replicated store that loses data~\cite{network_reliable}, to a file system that drops a write~\cite{all_not_equal}, to a confidential store that leaks a key.

To prevent such catastrophic failures, formal verification techniques can provide exhaustive correctness guarantees. These techniques allow a developer to (1) state a specification, (2) write an implementation, and (3) develop a machine-checkable proof that the implementation satisfies the specification on every possible input~\cite{coq, lean, verus, dafny}.
However, adoption of formal methods over the past two decades has been limited due to the immense manual labor required: verifying real-world systems takes months to years of expert effort~\cite{sel4,compcert,fscq,ironfleet,verdi,chapar}.

Naturally, one may ask, can we simply provide a coding agent with a specification and formal verification tooling, and ask it to build complex systems that provably satisfy that specification? Our experiments demonstrate that the answer is no.
We show in \cref{sec:eval} that Codex (GPT-5.4)~\cite{codex} and Claude Code (Opus 4.6)~\cite{claude-code} fail to generate correct systems for $5$ of $7$ widely studied distributed key-value-store consistency specifications (e.g., causal consistency, monotonic reads).\del{ The failure mode we observe is \del{a structural one}\rev{structural}:}
Following typical human patterns, agents treat verification as a downstream check on code they have already written. This approach (a) requires writing the whole proof at once, which is exceptionally difficult even for humans\footnote{Past work~\cite{chapar} reports 9--12 months of expert effort to write a proof for a key-value (KV) store.}, and (b) defers all correctness feedback to the end of an implementation, depriving the agent of valuable guiding signals.
\rev{If the implementation is wrong, no amount of proof effort can succeed; repairing the full implementation after proof failure is only partly helpful, as agents  often lack the context to tell which design choices to rule out.}

\looseness=-1 To address these challenges, we introduce \emph{Inductive Deductive Synthesis} (IDS), a technique that enables agents to synthesize provably correct complex systems from a specification. 
\begin{wrapfigure}{r}{0.625\linewidth}
\centering
\vspace{-0.8\baselineskip}
\includegraphics[width=\linewidth]{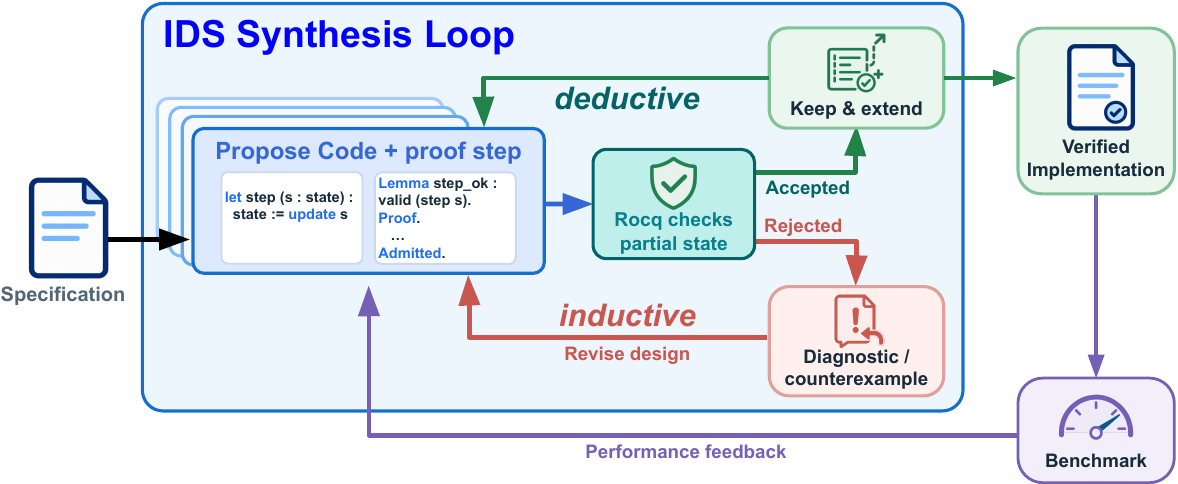}
\caption{\small Code and proof advance jointly and incrementally with IDS; the proof assistant grades each partial step. 
}
\label{fig:ids-intro}
\vspace{-0.8\baselineskip}
\end{wrapfigure}
Our key insight is to synthesize both the implementation and its proof \textit{jointly and incrementally} (\cref{fig:ids-intro}), while learning from failures at each step to guide the search toward promising solutions and away from dead ends. As detailed in \cref{sec:background}, every implementation decision (e.g., adding a data structure or modifying control flow) is paired with a \rev{co-evolving}\del{corresponding} proof update (e.g., adding new constraints or splitting lemmas).

\looseness=-1 IDS offers two benefits: (1) each individual joint step is considerably simpler to \textit{prove}, decreasing the complexity of proof generation, and (2) the intermediate \textit{partial} proofs can be verified with a formal checker to assess the viability of the current design. A negative result rules out a dead-end implementation path before further work is wasted and serves as a counterexample to improve the design going forward. This is akin to ``chain-of-thought''~\cite{chain-of-thought}, except that the intermediate states are formally encoded and verified. Beyond correctness, this iterative \rev{co-evolution} loop also optimizes performance: as soon as a candidate's implementation is complete, it is benchmarked on a distributed testbed; the performance measurements then further guide the search toward efficient implementations.

\looseness=-1 We instantiate \sys as a multi-agent system (\cref{sec:design}), and we evaluate \sys{} on 7 distributed key-value-store consistency specifications: Chapar's published causal-consistency spec~\cite{chapar} plus six new \emph{\sys{} suite} specs we release with the system (\cref{app:suite}). \sys{} produces a complete verified implementation on all 7 in about \rev{$6.9$}\del{$6.8$} hours and \$$106$ per spec on average, with no human intervention or fine-tuning. \rev{Writing the six new specifications took an expert a few days (\cref{sec:discussion}); the cost of a verified system thus moves from months of proof effort to a few days of specification plus hours of synthesis.} On the two specs our vanilla agent baseline completed, \sys{} runs $1.6\times$ faster and $17\%$ cheaper. The discovered implementations match or beat hand-written expert references on every spec and reach up to $3\times$ the throughput of Chapar's published vector-clock reference, a margin we attribute to a design-space search broader than what manual proof allows.

\looseness=-1 As a general prover, \sys{} achieves state-of-the-art performance on four cross-language verification benchmarks (DafnyBench~\cite{dafnybench}, miniCodeProps~\cite{minicodeprops}, Verus-Bench~\cite{autoverus}, CoqStoq~\cite{rango}), beating agentic baselines that share its prompt and underlying model by $+10\%$ to $+51\%$. \rev{It also leads on three proof benchmarks from real projects, including the Linux kernel: $39/60$ on VerKer~\cite{tu2026agentic} (prior $18/60$), $432/500$ on VeriSoftBench~\cite{xin2026verisoftbench} (prior $205/500$), and $838/849$ on VeruSAGE-Bench~\cite{yang2025verusage} (prior $797/849$).} 
On three\del{ further} code-and-proof synthesis benchmarks, \sys{} \del{again leads}\rev{leads}:
$176/189$ on VERINA~\cite{verina} (prior $38/189$), $65/77$ on AlgoVeri~\cite{algoveri} (prior $31/77$), and a perfect $62/62$ on CloverBench~\cite{clover}.

\looseness=-1 \noindent In summary, we make the following contributions in this work:
\begin{itemize}[leftmargin=*,nosep]
\item \textbf{Inductive Deductive Synthesis (IDS):} the first general technique for jointly \del{synthesizing}\rev{co-evolving} code and machine-checked proof under a partial-proof oracle, with feedback from failures and performance benchmarks in the same loop.
\item \textbf{IDS-based multi-agent system:} the first agentic system to autonomously generate \textit{verified} distributed systems. Solves $7/7$ consistency specifications where SOTA agents (GPT-5.4, Opus 4.6) solve only $2/7$, with up to $3\times$ the throughput of published verified systems.
\item \textbf{\sys{} suite:} six new \coq{} specifications for distributed key-value-store consistency models, released as an open benchmark for verified-synthesis research.
\end{itemize}

\section{Related Work}
\label{sec:related}

Verified code generation with an LLM has three components: (1) a specification of desired guarantees, (2) an implementation, and (3) a machine-checkable proof that the implementation satisfies the specification. Prior work falls into four threads: three each address one component (specification generation, program generation with an evaluator, automated proof generation), and verified synthesis attempts (2) and (3) jointly. \sys{} pursues this last class on distributed-systems specifications.

\looseness=-1 \textbf{Specification generation. \ }
A complementary line uses LLMs to \del{recover}\rev{extract} 
formal specifications from code or intent~\cite{nl2postcond,autospec};
Clover~\cite{clover} generates formal annotations from natural language specifications, and then checks the consistency of code with the formal and natural specifications \del{of}\rev{on} Dafny benchmarks.
\rev{KBSpec~\cite{wang2026kbspec} generates a Java Modeling Language specification from a knowledge base of documentation and verifier feedback, and SLD-Spec~\cite{chen2025sld} generates an ACSL specification for C~\cite{baudin2024acsl} via program slicing and logical deletion.}
\sys{} takes such specifications as input rather than producing them.
\rev{Specification generation from reference code is an avenue of further investigation (\cref{sec:discussion,app:limitations-impacts}).}

\textbf{Program generation with an evaluator. \ }
A separate strand of work treats program construction as search guided by evaluator feedback. 
Coding agents get feedback from given or self-generated tests~\cite{swe-agent,reflexion},
or from communication with other specialized agents~\cite{metagpt,chatdev,autogen}.
Algorithmic discovery systems use a scoring function to drive evolution~\cite{funsearch,alphaevolve} or deep reinforcement learning~\cite{alphadev}. 
\sys{} shares the evaluator-guided search scheme;
however, instead of tests or scores, it uses a proof assistant.
In contrast to sampled signals (tests over traces, specific inputs, or estimated scores),
type-checked proofs guarantee correctness on all inputs.

\textbf{Automated proof generation. \ }\looseness=-1
Most prior LLM-for-verification work fixes the implementation and specification, asking the model only for the proof.
For SMT-backed languages such as Dafny~\cite{dafny} and Verus~\cite{verus}, \rev{recent systems synthesize loop invariants and assertions for the verifier. They do this either via compiler-feedback loops~\cite{autoverus} or targeted \rev{assertion}\del{invariant} generation~\cite{laurel}}.
\del{For tactic-driven assistants such as Lean~\cite{lean} and \coq{}~\cite{coq}, prover systems have been trained on databases of prior formal proofs~\cite{coqgym}, natural-language proofs~\cite{wang2024theoremllama}, and conjectured theorems~\cite{dong2025stp}.
Approaches differ in scope: some generate full proofs in one pass~\cite{deepseek-prover-v2,kimina-prover,baldur}; others decompose the proof into sub-goals~\cite{li2026goedel,zhao2024subgoalxl,dsp}; tactic-level methods iteratively predict and repair the next proof step from the current proof state and verifier feedback~\cite{hubert2025olympiad,leandojo,copra,apollo,ruida2025ma}.}
\rev{For tactic-driven assistants (Lean~\cite{lean}, \coq{}~\cite{coq}\rev{, Isabelle}), provers trained on prior formal proofs~\cite{coqgym}, natural-language proofs~\cite{wang2024theoremllama}, and conjectured theorems~\cite{dong2025stp} \rev{either generate full proofs, in one pass~\cite{deepseek-prover-v2,kimina-prover} or with repair~\cite{baldur,apollo,ruida2025ma}, decompose into sub-goals~\cite{li2026goedel,zhao2024subgoalxl,dsp}, or predict and repair one tactic at a time from proof state and verifier feedback~\cite{hubert2025olympiad,leandojo,copra}.}\del{either generate full proofs in one pass~\cite{deepseek-prover-v2,kimina-prover,baldur}, decompose into sub-goals~\cite{li2026goedel,zhao2024subgoalxl,dsp}, or predict and repair one tactic at a time from proof state and verifier feedback~\cite{hubert2025olympiad,leandojo,copra,apollo,ruida2025ma}.}}
\rev{AutoRocq~\cite{tu2026agentic} and ProofCoop~\cite{kaufman2026proofcoop} likewise take the implementation as given, the latter with an ensemble of custom-trained models. \sys{} instead starts from no implementation, so proof obligations shift as code evolves; it \rev{co-evolves code and proof}\del{generates code and proof jointly through co-evolution} with off-the-shelf agents, and further optimizes performance.}

\rev{\textbf{Verified synthesis. \ }\looseness=-1
Producing a program and its proof from a specification alone dates to deductive~\cite{manna-waldinger,synquid} and counterexample-guided inductive synthesis~\cite{sketch,cegis}\rev{ and their combination~\cite{seshia2015combining}, but these methods'}\del{, and their combination~\cite{seshia2015combining} but their} inference rules and search spaces confined results to small functions. LLMs rekindle this dream. VERINA~\cite{verina} benchmarks spec, code, and proof generation and finds automatic provers wanting; AlphaVerus~\cite{alphaverus} bootstraps verified Verus code at function level. At the other extreme, hand-built verified systems (IronFleet's Paxos~\cite{ironfleet}, Verdi's Raft~\cite{verdi}, FSCQ's crash-safe file system~\cite{fscq}, Anvil's Kubernetes controller~\cite{anvil}, Chapar's KV store~\cite{chapar}, and Iris-based work~\cite{iris,grove,perennial}) show that machine-checked guarantees are achievable at systems scale, but at months to years of expert effort. 
\sys{} shows that verified synthesis need not be small-scale or slow: it automatically generates implementations and their machine-checkable correctness proofs for distributed key-value stores in hours rather~than~months.}

\del{\textbf{Verified synthesis. \ }\looseness=-1
The long-standing goal of producing a program and its correctness proof from a specification alone goes back to deductive synthesis~\cite{manna-waldinger,synquid} and counterexample-guided inductive synthesis~\cite{sketch,cegis}, but the inference rules and candidate-search space confined results to small individual functions. The LLM era rekindles this dream. VERINA~\cite{verina} provides a benchmark to evaluate specification, code, and proof generation, evaluates existing tools, and underscores significant challenges for automatic provers.
Recent LLM-driven verified-synthesis work targets SMT-backed languages: AlphaVerus~\cite{alphaverus} bootstraps verified Verus code via self-improving translation, scaling to function-level tasks. At the other extreme, hand-built verified distributed systems, such as IronFleet's Paxos-replicated state machines~\cite{ironfleet}, Verdi's Raft~\cite{verdi}, FSCQ's crash-safe file system~\cite{fscq}, Anvil's verified Kubernetes controllers~\cite{anvil}, Chapar's causally-consistent KV stores~\cite{chapar}, and Iris-based concurrent and distributed verification~\cite{iris,grove,perennial} demonstrate that machine-checkable guarantees \emph{are} achievable at systems scale, but at months to years of expert proof effort. \sys{} shows that verified synthesis need not be small-scale or slow: it automatically generates implementations and their machine-checkable correctness proofs for distributed key-value stores in hours rather~than~months.}

\section{Background and Overview}
\label{sec:background}

We begin with a short introduction to formal verification.
A formally verified system has three pieces. The \emph{specification} is a precise mathematical statement of correctness; the \emph{implementation} is the code that runs; and the \emph{proof} is a machine-checkable argument that, on every input, the implementation matches the specification. Historically, humans wrote all three; \sys{} takes a human-written spec and automatically produces both implementation and proof, using LLM agents driven by the proof assistant \coq{}~\cite{coq}. We show a simple example: a \emph{counter}, a state machine with two operations (\texttt{inc} to increment the count and \texttt{read} to return it) starting from an initial state \texttt{init}.

\providecolor{specbg}{HTML}{F7F8FA}
\providecolor{specrule}{HTML}{C8CDD3}
\providecolor{speckwd}{HTML}{1F4E79}
\providecolor{partialhi}{HTML}{FFF3CD}
\providecolor{completehi}{HTML}{D4EDDA}

\begin{figure}[tb]
\centering
\lstset{
  language=Coq,
  basicstyle=\ttfamily\scriptsize,
  keywordstyle=\bfseries\color{speckwd},
  showstringspaces=false,
  columns=fullflexible,
  breaklines=true,
  numbers=none,
  frame=single,
  framerule=0.3pt,
  framesep=3pt,
  rulecolor=\color{specrule},
  xleftmargin=4pt,
  xrightmargin=2pt,
  backgroundcolor=\color{specbg},
  aboveskip=0pt,
  belowskip=0pt,
}
\begin{minipage}[t]{0.32\textwidth}
\centering
\textbf{Specification}\\[2pt]
\begin{lstlisting}
Module Type CounterSpec.

  Parameter t    : Type.
  Parameter init : t.
  Parameter inc  : t -> t.
  Parameter read : t -> nat.

  Axiom read_init :
    read init = 0.

  Axiom read_inc :
    forall s,
    read (inc s) =
      S (read s).

End CounterSpec.
\end{lstlisting}
\end{minipage}\hfill
\begin{minipage}[t]{0.32\textwidth}
\centering
\textbf{Partial impl.\ \& proof}\\[2pt]
\begin{lstlisting}
Definition t := list unit.
Definition init : t := nil.
\end{lstlisting}
\begin{lstlisting}[backgroundcolor=\color{partialhi}]
Definition inc (s : t) : t.
Admitted.
\end{lstlisting}
\begin{lstlisting}
Definition read (s : t) :=
  length s.

Theorem read_init :
  read init = 0.
Proof. reflexivity. Qed.
Theorem read_inc :
  forall s,
  read (inc s) = S (read s).
\end{lstlisting}
\begin{lstlisting}[backgroundcolor=\color{partialhi}]
Admitted.
\end{lstlisting}
\end{minipage}\hfill
\begin{minipage}[t]{0.32\textwidth}
\centering
\textbf{Complete impl.\ \& proof}\\[2pt]
\begin{lstlisting}
Definition t := list unit.
Definition init : t := nil.
\end{lstlisting}
\begin{lstlisting}[backgroundcolor=\color{completehi}]
Definition inc (s : t) :=  tt::s.
\end{lstlisting}
\begin{lstlisting}
Definition read (s : t) :=
  length s.
Theorem read_init :
  read init = 0.
Proof. reflexivity. Qed.
Theorem read_inc :
  forall s,
  read (inc s) = S (read s).
\end{lstlisting}
\begin{lstlisting}[backgroundcolor=\color{completehi}]
Proof. intros s. unfold read, inc.
  simpl. reflexivity. Qed.
\end{lstlisting}
\end{minipage}
\vspace{-3pt}
\caption{Counter \textit{specification} (left), \textit{partial} synthesis (center, yellow = \texttt{Admitted}), and \textit{complete} synthesis (right, green = filled in). \coq{} accepts both files.}
\label{fig:counter-listing}
\vspace{-8pt}
\end{figure}

\looseness=-1 \cref{fig:counter-listing} shows the counter's \textit{specification} (left), a \textit{partial synthesis} (center), and a \textit{complete synthesis} (right). The spec states two properties, or ``axioms,'' about \texttt{init}, \texttt{inc}, and \texttt{read}: reading the initial state returns zero, and reading after an increment returns one more than reading before. In the partial synthesis, \texttt{inc}'s body and the \texttt{read\_inc} theorem are deferred via \texttt{Admitted} (a placeholder for unfinished work), but \coq{} still accepts the file: the chosen representation (a list whose length encodes the count) is consistent with the work so far. The complete synthesis fills in the deferred pieces.
The representation is a free choice; any state satisfying the axioms works, though \texttt{nat} (natural numbers) is more efficient.

\textbf{\coq{}'s type-checker as an oracle, including for partial work. \ }
\coq{}'s checker \textit{accepts} a declaration/proof if and only if it satisfies its stated type/proposition, and rejects with a precise diagnostic otherwise. There are no false positives or false negatives, in contrast to tests (only the inputs tried), static analysis (over-approximate), or LLM-as-judge (guesses). The verdict also extends to \emph{partial} code, as the center column of \cref{fig:counter-listing} illustrates: unproven obligations are stated as \emph{deferred holes} (\texttt{Admitted} lemmas or stub function bodies), and the file still type-checks.
(\cref{app:progression} walks a longer \texttt{all\_less\_than} example through this progression.)
\sys{} leverages the type-checker as an oracle to drive a \emph{deductive synthesis}~\cite{manna-waldinger}: at every step, the agent extends the partial implementation and proof, then asks \coq{} whether the partial state still type-checks, \rev{thus co-evolving implementation and proof}. A yes keeps the design on track; a no rules it out before further work is committed. \aknew{If progress stalls (measured by the number of partial proofs left empty), IDS reverts to an earlier state.}

\textbf{From counter to distributed stores. \ }
\sys{} applies this loop at scale.
For distributed systems, the specification grows from simple axioms to conditions over executions involving messages, clients, replicas, and failures.
For example, \texttt{read\_inc} in a distributed setting becomes read-your-writes: a client that increments then reads must observe its own increments.
The implementation grows from a few-line module to a multi-replica protocol with larger state, multiple message types, and handlers.
For example, a replica can no longer store one \texttt{nat}; it must keep a vector with one entry per client, tracking the increments received from that client. On a request carrying the client's increment count, the replica can then tell whether it is up-to-date enough to serve, providing read-your-writes.
The proof grows from simple tactics to an inductive simulation argument (an induction linking implementation and spec states) with dozens of helper lemmas.
Such implementations span a large, subtle design space with widely varying performance, explored by decades of research. \cref{sec:design} describes the agent that drives this process, \cref{app:suite} gives the full specifications, and \cref{sec:eval} reports results.
\vspace{-1em}

\section{\sys{} Agentic Architecture}
\label{sec:design}

\begin{figure}[t]
   \centering
   \includegraphics[width=\linewidth]{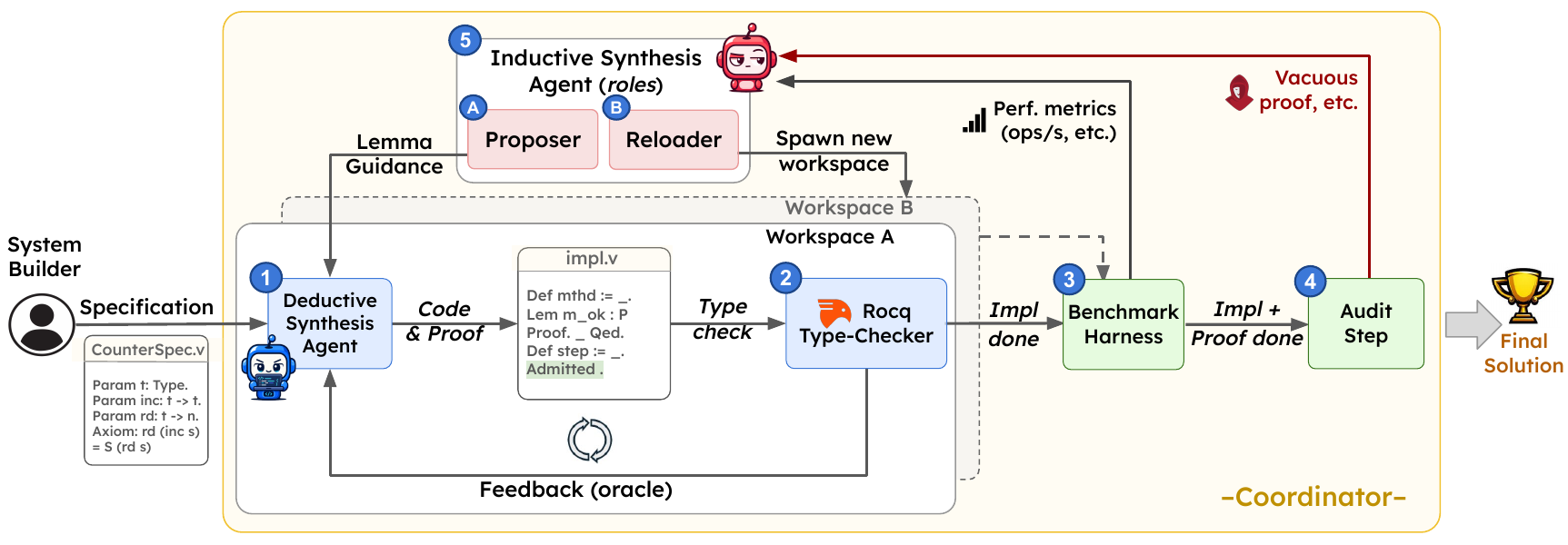}
   \vspace{-3pt}
   \caption{\sys{}' agentic architecture; code and proof advance jointly throughout. A coordinator runs parallel \fignum{1} Deductive Synthesis Agents (DSAs); each step is graded by \fignum{2} \coq{}'s type-checker, completed implementations are \fignum{3} benchmarked, and closed proofs are \fignum{4} audited for non-vacuity. On failure, the \fignum{5} Inductive Synthesis Agent (ISA) intervenes: \fignum{A} \emph{proposer} adds helper lemmas on tactical stalls; \fignum{B} \emph{reloader} respawns a DSA with a fresh design on strategic dead-ends.
}
   \label{fig:architecture}
\vspace{-8pt}
\end{figure}

\textbf{Overview. \ }
We now describe how we realize IDS as an agentic architecture (\cref{fig:architecture}). Given a system interface and property specifications, the framework orchestrates multiple LLM agents guided by verification and performance feedback to generate an efficient, mechanically-checked implementation. This architecture relies on a powerful synergy: deductive synthesis constructs the code and proof under a given strategy, while inductive synthesis discovers better strategies from failed attempts.

Inspired by deductive synthesis~\cite{manna-waldinger,manna2006synthesis}, \sys{} introduces Deductive Synthesis Agents (DSAs, \fignum{1}) to incrementally synthesize implementations from specifications. Guided by a strategy, a DSA recursively decomposes a component and its specification into sub-components using placeholders. 
\looseness=-1 The DSA constructs proofs assuming these sub-components are correct, repeating this process down to trivial elements. The \coq{} type-checker (\fignum{2}) continuously verifies both complete and partial proofs throughout this decomposition.

While this deductive approach succeeds when guided by a sound search strategy (e.g., a high-level design or proof approach in the prompt), a DSA can stall under a dead-end strategy. Inspired by inductive synthesis~\cite{cegis,sketch,clarke2000counterexample}, \sys{} introduces an Inductive Synthesis Agent (ISA, \fignum{5}) that learns from failed strategies to propose more promising ones. The ISA receives feedback when a DSA stalls or yields poor performance. It then intervenes in two roles: a \emph{proposer} for tactical, local adjustments (\fignum{A}), and a \emph{reloader} for strategic, global shifts to entirely new designs (\fignum{B}).

\looseness=-1 \textbf{Coordinator. \ } The coordinator launches and orchestrates the DSAs and the ISA, maintaining overall system state and managing a parallel pool of workers. While DSAs can report localized errors, they cannot detect strategic dead-ends on their own. The coordinator monitors progress to identify stagnating agents, recording failed strategies and invoking the ISA to propose new ones before respawning the DSA. 
 
To ensure the synthesized system is not only correct but also performant, the coordinator benchmarks each candidate eagerly, even before its proof completes. It extracts the implementation to executable code and runs the benchmark harness (\fignum{3}, \cref{sec:eval-setup}). The measurements are fed back to the ISA so that its future strategies guide the DSA toward more efficient implementations. Once a proof closes (i.e., the proof completes with no remaining \texttt{Admitted} placeholders), the coordinator audits the result to ensure it is fully verified, non-trivial, and implements the expected interface (\fignum{4}). \rev{The audit is a deterministic script, not a model (\cref{app:auditor}). With it, $20$ of $21$ runs on the seven key-value store specifications yield a genuine proof and the other does not close; without it ($-$A in \cref{tab:eval-ablation}), $7$ are genuine, $9$ leave admits or axioms, and $5$ are vacuous.} Finally, \del{it }\rev{the coordinator} returns the most efficient verified solution found within the time budget. 

\textbf{DSA: Deductive Synthesis Agent. \ }
\label{sec:design-worker}
Off-the-shelf LLMs, \rev{having been} pretrained on a large corpus of programs and proofs, are well-equipped to guide synthesis search,
as we empirically demonstrate in \cref{sec:eval-synthesis}. To harness this capability, we design the DSA as a generic LLM coding agent operating under strict constraints: implementations must be well-typed before proving, specifications cannot be altered, \rev{\texttt{Axiom}, \texttt{Hypothesis}, and \texttt{Parameter} are forbidden at every step (\cref{app:worker}), }final outputs must contain no unproven assumptions, and system state must remain bounded.

At each node of the search tree, the agent can take the following actions to advance the synthesis: (1) define partial implementations, (2) close open proof branches, or (3) decompose complex lemmas into simpler helper lemmas. After each step, the implementation and proof are passed to \coq{}'s type-checker. If there are no errors, the state is saved, building a search tree of well-typed partial implementations and proofs. If errors occur, the agent can take the following actions: (1) repair the failed step by attempting a different approach, or (2) revert to an earlier node in the search tree.

\textbf{ISA: Inductive Synthesis Agent. \ } The DSA cannot reliably step back and pivot a failing strategy on its own. The ISA fills this gap as a stateless LLM agent that reasons over a strategy summary and proposes new approaches across two decision horizons: \begin{itemize}[leftmargin=*,nosep,topsep=2pt] 
\item \textbf{Proposer (tactical).} Triggered when the strategy is promising but the DSA is stuck on a specific proof. The proposer offers finer-grained assistance, suggesting helper lemmas and structural decompositions from a holistic view of the proof state.
\item \textbf{Reloader (strategic).} Triggered when the strategy is a dead-end or slow. Rather than allowing the DSA to endlessly apply small fixes, the reloader intervenes by respawning the DSA with an entirely new high-level design. \rev{Before a reload, the DSA reuses its saved states (repair first, then revert); the reloader writes the new design from the record of past attempts (\cref{app:worker,app:resetter}).}
\end{itemize}
\rev{Within a run, \sys{} keeps the full history of attempts in persistent files (a design log of every approach tried and why it failed, plus the current status and strategy guidance), which the agents selectively inspect on each call rather than carrying in a single ballooning context session.}

\section{Evaluation}
\label{sec:eval}

We evaluate \sys{} on 7 \coq{} specifications of distributed key-value-store consistency: Chapar's published causal-consistency spec~\cite{chapar} and 6 new \emph{\sys{} suite} specs co-developed with a formal-verification and distributed-systems expert (\cref{app:suite}). We extract synthesized implementations from Gallina to OCaml and benchmark them on a 5-VM Google Cloud cluster under a distributed runtime against hand-written expert references. For synthesis, we compare against SOTA coding agents, Codex (GPT-5.4)~\cite{codex} and Claude Code (Opus 4.6)~\cite{claude-code}, under the same prompt and budget. We answer three questions in the following subsections:
\begin{itemize}[leftmargin=*,nosep,itemsep=2pt,topsep=2pt]
\item \textbf{Q1. Does \sys{} handle hard specifications?} \sys{} succeeds on all $7$ in about \rev{$6.9$}\del{$6.8$} hours and \$$106$ per spec on average (max $11$h, \$$155$), versus $2$ of $7$ each for Codex and Claude Code under the same prompt and budget (\cref{sec:eval-synthesis}).
\item \textbf{Q2. Are the synthesized implementations performant?} They match or beat every expert-implemented reference: \rev{up to }$3\times$ throughput on Chapar's vector-clock reference, $1.4\times$ on \sys{} suite CC and monotonic reads, comparable on read-your-writes (within \rev{$21\%$}\del{$20\%$}), and sustained throughput on MW and RYW+MW where the reference times out (\cref{sec:eval-perf}).
\item \textbf{Q3. Which of \sys{}' design decisions drive these results?} The joint code-and-proof co-design, the ISA's proposer and reloader roles, the audit step, the \coq{} feedback to the DSA, and the performance feedback to the ISA each contribute consequentially. \rev{The proof-engineering guidance in the DSA's prompt is an efficiency aid: removing it has minimal effect on \rev{efficacy}\del{correctness} but slows the search.} The DSA alone also sets a new state-of-the-art on \del{four }\rev{seven} public proof benchmarks across four languages (\cref{sec:eval-ablations}).
\end{itemize}

\subsection{Experimental Setup}
\label{sec:eval-setup}

\textbf{Specifications. \ }
Each specification is a high-level reference implementation of a key-value store; its state type and operations define the consistency notion. \sys{} takes a specification and produces a refining implementation paired with a machine-checked proof of correctness on every input.

Chapar~\cite{chapar} provides a published causal-consistency specification\del{ (CC)}. The other six form the \emph{\sys{} suite} benchmark dataset that we release alongside the system (\cref{app:suite}): read-your-writes (RYW), monotonic writes (MW), monotonic reads (MR), their composition (RYW+MW), and two causal-consistency variants: \sys{} suite CC and LCC (\emph{labeled} causal consistency, where each client subscribes to a set of topic labels and sees causality enforced only on those labels). The \sys{} suite spans the space of session consistency guarantees. To the best of our knowledge, no prior benchmark provides verified \coq{} artifacts across this range.
Difficulty grows in three tiers: simple (RYW, MW), specs whose proofs need many cases (MR, RYW+MW), and specs that also manage explicit dependency sets (\rev{IDS suite CC}\del{CC}, LCC).
A side-by-side summary of all seven specifications appears in \cref{app:consistency-specs}.

\textbf{Harness and metrics. \ }
Verified \coq{} implementations are extracted to OCaml (via \coq{}'s built-in extraction, an automatic compilation pass from Gallina to OCaml) and run under a distributed runtime on a 5-VM Google Cloud cluster. \rev{As in prior verified system works (Verdi~\cite{verdi} and Chapar~\cite{chapar}), for verification guarantees to hold, we assume that extraction, the OCaml toolchain and the runtime are trusted (\cref{app:limitations-impacts}).} Each run executes $4$ parallel workers $\times$ $1{,}000$ random operations; we sweep across put rates, and run a separate scaling sweep at put rate~$=50\%$ over $N \in \{1{,}000, 2{,}000, 5{,}000, 20{,}000\}$. \cref{app:perf-harness} covers the cluster topology and toolchain.
Every method runs under the same per-spec wall-clock and dollar budget. Runs that exhaust either budget are considered failed runs. 
Each cell aggregates three independent runs, and we report pass rate 
(successes out of three), 
time-to-finish, dollar cost, throughput, p99 latency, peak memory, and ops-per-worker scaling, all as medians. \rev{Reported time-to-finish includes every per-step \coq{} check; each step recompiles only the edited file,
taking about $4$ seconds on Chapar, versus $12$--$17$ seconds for a full build (\cref{app:worker}).}

\textbf{Baselines. \ }
The synthesis baselines are Codex (GPT-5.4)~\cite{codex} and Claude Code (Opus 4.6)~\cite{claude-code}. Both run under the same prompt and budget as \sys{}' DSA, so the comparison isolates the agentic architecture. Codex is also \sys{}' primary backend. 
For performance comparisons, the references are the hand-written expert implementations: 
Chapar's two published references (vector-clock and list-based)~\cite{chapar}, and the expert-written reference released alongside each \sys{} suite specification, except LCC, a custom spec with no expert reference.
For the cross-language evaluation, 
we adopt \del{four }\rev{seven} public verification benchmarks (DafnyBench, Verus-Bench, miniCodeProps, CoqStoq\rev{, VerKer, VeriSoftBench, VeruSAGE-Bench}) across the languages Dafny, Verus, Lean, and \coq{}. 
We compare against the strongest published prior \del{tool }\rev{method} for each benchmark.
\cref{app:benchmarks} covers three more (VERINA, AlgoVeri, CloverBench). These cross-language benchmarks target function- and algorithm-level tasks, smaller and simpler in scope than the real-world distributed systems above\rev{, though VerKer, VeriSoftBench, and VeruSAGE-Bench draw their tasks from the Linux kernel, $23$ formal-methods repositories, and eight real-world Verus projects}.

\definecolor{rowhl}{HTML}{EFF3F8}
\definecolor{passgood}{HTML}{C8E6C9}    %
\definecolor{passok}{HTML}{E6F4E6}      %
\definecolor{passfail}{HTML}{F8D7DA}    %
\definecolor{rulecol}{HTML}{B0B7BF}     %

\begin{table}[t]
\centering
\footnotesize
\setlength{\tabcolsep}{2pt}
\renewcommand{\arraystretch}{1.0}
\caption{\sys{} succeeds on \textbf{all seven specifications} ($\geq\,2/3$ runs) versus only \textbf{2 of 7 each} for Codex and Claude Code, while using \textbf{less total wall-clock and fewer total dollars}; on every spec where any method succeeds, \sys{} is also \textbf{faster and cheaper}. 
\textbf{pass rate}~$=$ \# successes out of $3$ (green: $\geq\,2/3$; red: $<\,2/3$); \textbf{hours}~$=$ median wall-clock until a successful run (\coq{} \texttt{Qed} accepted, audit step passed, extracted OCaml runs on the harness); \textbf{cost}~$=$ median dollars per run. All methods share the same model, verifier, pass criterion, and per-spec budget; relative std across runs is $\leq 7\%$.}
\label{tab:eval-synthesis}
\begin{tabular*}{\linewidth}{@{\extracolsep{\fill}} l ccc !{\color{rulecol}\vrule} ccc !{\color{rulecol}\vrule} ccc @{}}
\toprule
 & \multicolumn{3}{c}{Codex} & \multicolumn{3}{c}{Claude Code} & \multicolumn{3}{c}{\textbf{\sys{}}} \\
\cmidrule(lr){2-4} \cmidrule(lr){5-7} \cmidrule(lr){8-10}
Specification & Pass rate$\uparrow$ & Hours$\downarrow$ & Cost (\$)$\downarrow$ & Pass rate$\uparrow$ & Hours$\downarrow$ & Cost (\$)$\downarrow$ & Pass rate$\uparrow$ & Hours$\downarrow$ & Cost (\$)$\downarrow$ \\
\midrule
\rowcolor{rowhl}
\multicolumn{10}{l}{\emph{Chapar}} \\
Causal Consistency      & \cellcolor{passfail}0/3 & 12 & 160 & \cellcolor{passfail}0/3 & 11 & 165 & \cellcolor{passgood}\textbf{3/3} & \textbf{10} & \textbf{148} \\
\midrule
\rowcolor{rowhl}
\multicolumn{10}{l}{\emph{\sys{} suite}} \\
Read-Your-Writes        & \cellcolor{passgood}\textbf{3/3} & 4 & 60 & \cellcolor{passgood}\textbf{3/3} & 3 & 70 & \cellcolor{passgood}\textbf{3/3} & \textbf{2} & \textbf{52} \\
Monotonic Reads         & \cellcolor{passfail}0/3 & 12 & 130 & \cellcolor{passfail}0/3 & 10 & 140 & \cellcolor{passgood}\textbf{3/3} & \textbf{7} & \textbf{102} \\
Monotonic Writes        & \cellcolor{passok}2/3 & 5 & 65 & \cellcolor{passgood}\textbf{3/3} & 4 & 70 & \cellcolor{passgood}\textbf{3/3} & \textbf{3} & \textbf{58} \\
RYW + MW                & \cellcolor{passfail}0/3 & 9 & 125 & \cellcolor{passfail}1/3 & 6 & 110 & \cellcolor{passgood}\textbf{3/3} & \textbf{5} & \textbf{93} \\
Causal Consistency      & \cellcolor{passfail}0/3 & 12 & 160 & \cellcolor{passfail}0/3 & 11 & 170 & \cellcolor{passok}\textbf{2/3} & \textbf{11} & \textbf{155} \\
LCC                     & \cellcolor{passfail}0/3 & 13 & 180 & \cellcolor{passfail}0/3 & 11 & 160 & \cellcolor{passgood}\textbf{3/3} & \textbf{10} & \textbf{136} \\
\midrule
\textbf{Total across specs} & \cellcolor{passfail}2/7 & 67 & \$880 & \cellcolor{passfail}2/7 & 56 & \$885 & \cellcolor{passgood}\textbf{7/7} & \textbf{48} & \textbf{\$744} \\
\bottomrule
\end{tabular*}
\vspace{-8pt}
\end{table}

\subsection{Synthesis Correctness}
\label{sec:eval-synthesis}

\looseness=-1 \sys{} returns verified implementations for all seven specifications, a $3.5\times$ pass rate over the $2$ out of $7$ that each of Codex and Claude Code reach under the same prompt and budget. Even on the two specs (RYW and MW) where both baselines succeed, \sys{} is $1.6\times$ faster and $17\%$ cheaper (Table~\ref{tab:eval-synthesis}).
\sys{} succeeds by storing the data in a way that lets the proof split into smaller, manageable cases (e.g., one entry per key on Chapar CC, or one entry per (key, client) pair on monotonic reads); the baselines keep the spec's default layout, never backtrack, and stall on the same hard cases. We also evaluate two LLM-only baselines that ask Codex for an implementation only (the standard coding-agent task), with \mbox{best-of-$N{=}100$} generation, on $4$ properties (RYW, MW, MR, CC). \emph{Setting~1 (spec-given):} Codex receives the formal \coq{} specification. \emph{Setting~2 (vibe coding):} Codex receives only a natural-language paragraph of the property. We then check each selected implementation two ways: an adversarial multi-client trace flags implementations that violate the spec on a single execution, and a refinement proof against the spec catches the rest. Spec-given Codex passes both checks on $1/4$ properties; vibe coding on $0/4$ (\cref{app:llm-only-synthesis}). Even given the formal spec and $100$ candidates per property, the LLM alone cannot produce implementations that are guaranteed correct on every input. \rev{Conversely, given the spec and \sys{}' synthesized implementation, Claude Code closes the proof on only $4$ of $7$ specs, failing Chapar CC, \sys{} suite CC, and LCC (\cref{tab:proof-only}).} \rev{Implementation and proof are thus both hard, even though proofs make up $87$--$97\%$ of each final artifact's lines (\cref{tab:synth-artifacts}).} \rev{Only RYW passes in Setting~1, and only RYW passes a majority of runs without joint discovery ($-$J, \cref{sec:eval-ablations}). The simplest spec closes both times; everything harder needs implementation and proof to~advance~together.}

\textbf{Chapar causal consistency. \ }
\sys{} closes Chapar's published causal-consistency proof, originally a major hand-written verification~\cite{chapar}, in $10$ hours and \$148 (roughly $200\times$ faster than the cited 9--12 months of expert effort), while both agent baselines fail \rev{despite spending more time and money}\del{to finish within $1.2\times$ our wall-clock and cost}.
\sys{} succeeded by changing how the data was stored. Its first attempt kept each replica's state as one big object, so proving correctness required reasoning about every key and message together, and the proof got stuck. \sys{} backtracked, gave each key its own small table, and the proof then split into one easy case per key. The full trajectory and the per-family lemma counts are given in \cref{app:synth-results}.

\textbf{\sys{} suite specifications. \ }
\sys{} handles the four harder \sys{} suite specifications (monotonic reads, RYW+MW, LCC, and \sys{} suite CC) in $5$ to $11$ hours per spec, with $3$-of-$3$ pass rates on the first three and $2$-of-$3$ on \sys{} suite CC. In comparison, Codex never succeeds, and Claude Code succeeds only once, on RYW+MW. The same Chapar pattern recurs: the first design gets stuck on the spec's default layout, then \sys{} backtracks and rebuilds the data so the proof splits into smaller cases, \del{for instance one entry per (key, client) on monotonic reads, or }\rev{e.g.,} one case per message on the two CC variants\del{. Per-specification details are reported in \cref{app:synth-results}}\rev{ (\cref{app:synth-results})}.

\rev{\textbf{Failures and cost. \ } \sys{} recovers from tactical stalls, strategic dead-ends, and vacuous solutions; its only failed run ($1$ of $21$, on \sys{} suite CC) ended at the budget.
Because \coq{} checks every proof and the audit filters trivial ones, a failure yields no artifact rather than a false positive (\cref{tab:failure-modes}). Cost scales with the complexity of the guarantee, not artifact size: the session guarantees (RYW, MW, RYW+MW, MR) take $2$--$7$ hours and \$$52$--$102$, the causal ones (LCC, Chapar CC, \sys{} suite CC) $10$--$11$ hours and \$$136$--$155$. LCC is 29\% larger than Chapar CC ($3{,}924$ vs.\ $3{,}037$ lines) yet closes in the same $10$ hours for less (\$$136$ vs.\ \$$148$; \cref{tab:eval-synthesis,tab:synth-artifacts}).}

\begin{figure}[t]
\centering
\includegraphics[width=0.92\linewidth]{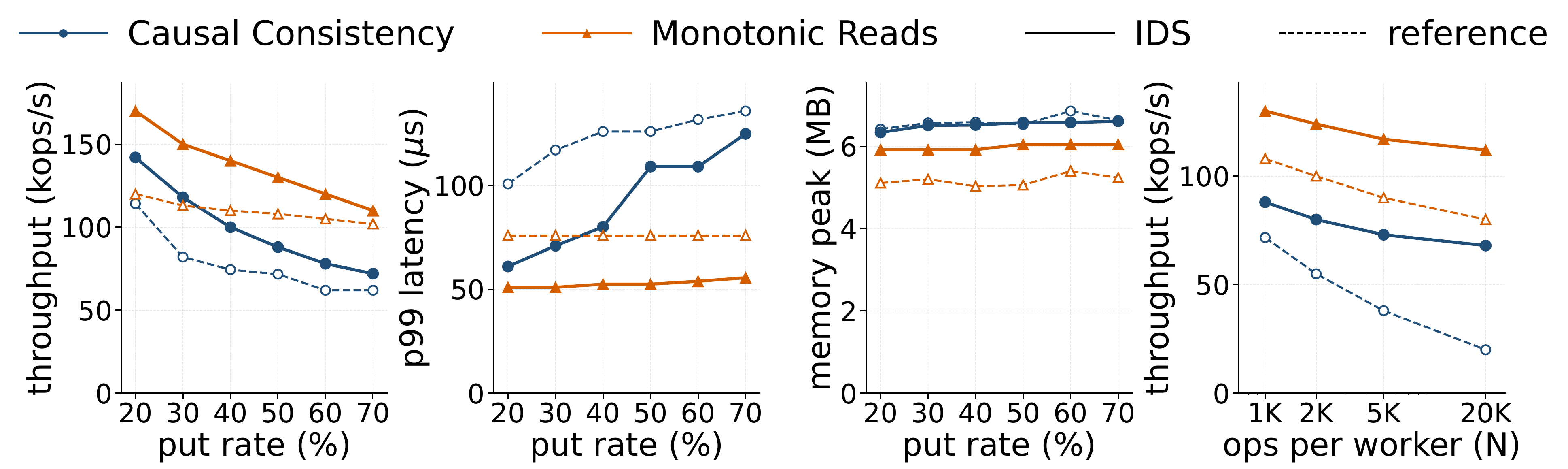}
\vspace{-3pt}
\caption{Throughput, p99 latency, and peak memory vs.\ put rate, and throughput vs.\ ops-per-worker $N$, for \sys{} suite CC and monotonic reads. Solid $=$ \sys{}; dashed $=$ reference (function-as-map for both). The per-spec breakdown for all seven specifications appears in
\cref{app:perf-curves}.}
\label{fig:eval-perf}
\vspace{-8pt}
\end{figure}

\subsection{Runtime Performance}
\label{sec:eval-perf}

\sys{}' synthesized implementations match or beat every prior reference: \rev{up to }$3\times$ throughput on Chapar's vector-clock reference, $1.4\times$ on \sys{} suite CC and monotonic reads, comparable on read-your-writes (within \rev{$21\%$}\del{$20\%$}), and $152$k and $139$k ops/s on MW and RYW+MW where the reference times out at every put rate (\cref{app:perf-curves}).\del{ We compare on throughput, p99 latency, peak memory, and ops-per-worker scaling.} \cref{fig:eval-perf} shows \sys{} suite CC and monotonic reads vs.\ the references\del{, and} 
(full per-spec breakdown
in \cref{app:perf-curves}). The gap follows from one mechanism: 
\sys{}' data-store representations are bounded,
while the references' grow with workload size. These representations emerged from the joint code+proof loop: the proof obligation drove the agent to them (\cref{sec:eval-synthesis}). The performance feedback from the benchmark harness then steers the agent toward the fastest among multiple verifying candidates (\cref{sec:eval-ablations}).\looseness=-1

\looseness=-1 \textbf{\sys{} suite CC. \ }
Both implementations enforce causality with one timestamp per client; the data store differs. The reference's function-from-key-to-value extracts to OCaml as a chain that grows per put (every get walks it), while \sys{} uses a balanced-tree map where a lookup costs at most the tree depth. \sys{} reaches up to $1.4\times$ peak throughput at low put rate where gets dominate, with p99 latency up to $1.7\times$ lower. 
The gap narrows at high put rate as broadcast becomes the bottleneck; tail latency \del{tracks the same effect, rising }\rev{rises} with put rate on both implementations as broadcast queue waits grow. Peak memory is comparable at $1$k ops per worker, as vector-clock state is bounded by client count rather than put rate; larger workloads would push it upward. On ops-per-worker scaling, the reference's throughput collapses by more than half from $1$k to $20$k ops per worker while \sys{} loses only about a quarter.

\looseness=-1 \textbf{Other \sys{} suite specifications. \ }
Unlike CC, these specs only need per-client guarantees, so they do not pay CC's per-write cross-client coordination cost; their tail latency stays flat across put rate (\cref{fig:eval-perf}, MR). On the other \sys{} suite specs, \sys{} ranges from comparable performance on read-your-writes (chain stays short) to a large gap with the reference timing out at every put rate on the monotonic-write specs (MW, RYW+MW). 
The four with prior references (RYW, MW, MR, RYW+MW) share the closure-chain mechanism: the reference's function-from-key-to-value extracts as a chain that grows per put; \sys{} replaces it with a flat list or balanced tree, and the speedup tracks how often the chain is walked. 
On the monotonic reads benchmark, a get in the reference walks every operation, so \sys{} outperforms it by $1.4\times$.

\subsection{Ablations and Generalization}
\label{sec:eval-ablations}

\begin{figure}[t]
\centering
\captionsetup{justification=raggedright,singlelinecheck=false}
\begin{minipage}[b]{0.335\linewidth}
\centering
\providecolor{rowhl}{HTML}{EFF3F8}
\centering
\scriptsize
\setlength{\tabcolsep}{1.1pt}
\renewcommand{\arraystretch}{1.00}

\begin{tabular}{@{}l ccccccc@{}}
\toprule
         & full & $-$J & $-$P & $-$R & $-$A & $-$VF & \rev{$-$G} \\
\midrule
\rowcolor{rowhl}
Chapar CC& 3/3 & 0/3 & 0/3 & 0/3 & 1/3 & 0/3 & \rev{3/3} \\
RYW      & 3/3 & 2/3 & 2/3 & 3/3 & 3/3 & 1/3 & \rev{3/3} \\
MW       & 3/3 & 1/3 & 2/3 & 2/3 & 2/3 & 0/3 & \rev{3/3} \\
RYW+MW   & 3/3 & 0/3 & 1/3 & 2/3 & 1/3 & 0/3 & \rev{3/3} \\
MR       & 3/3 & 1/3 & 0/3 & 1/3 & 0/3 & 0/3 & \rev{3/3} \\
\sys{} suite CC& 2/3 & 0/3 & 0/3 & 0/3 & 0/3 & 0/3 & \rev{2/3} \\
LCC      & 3/3 & 0/3 & 0/3 & 0/3 & 0/3 & 0/3 & \rev{2/3} \\
\midrule
\rowcolor{rowhl}
VERINA   & 176 & 160 & 150 & 167 & 176 & 110 & \rev{171} \\
\bottomrule
\end{tabular}
\captionof{table}{Component ablations: \# successes from $3$ runs ($189$ tasks for VERINA).}
\label{tab:eval-ablation}

\end{minipage}\hfill
\begin{minipage}[b]{0.32\linewidth}
\centering
\includegraphics[width=\linewidth]{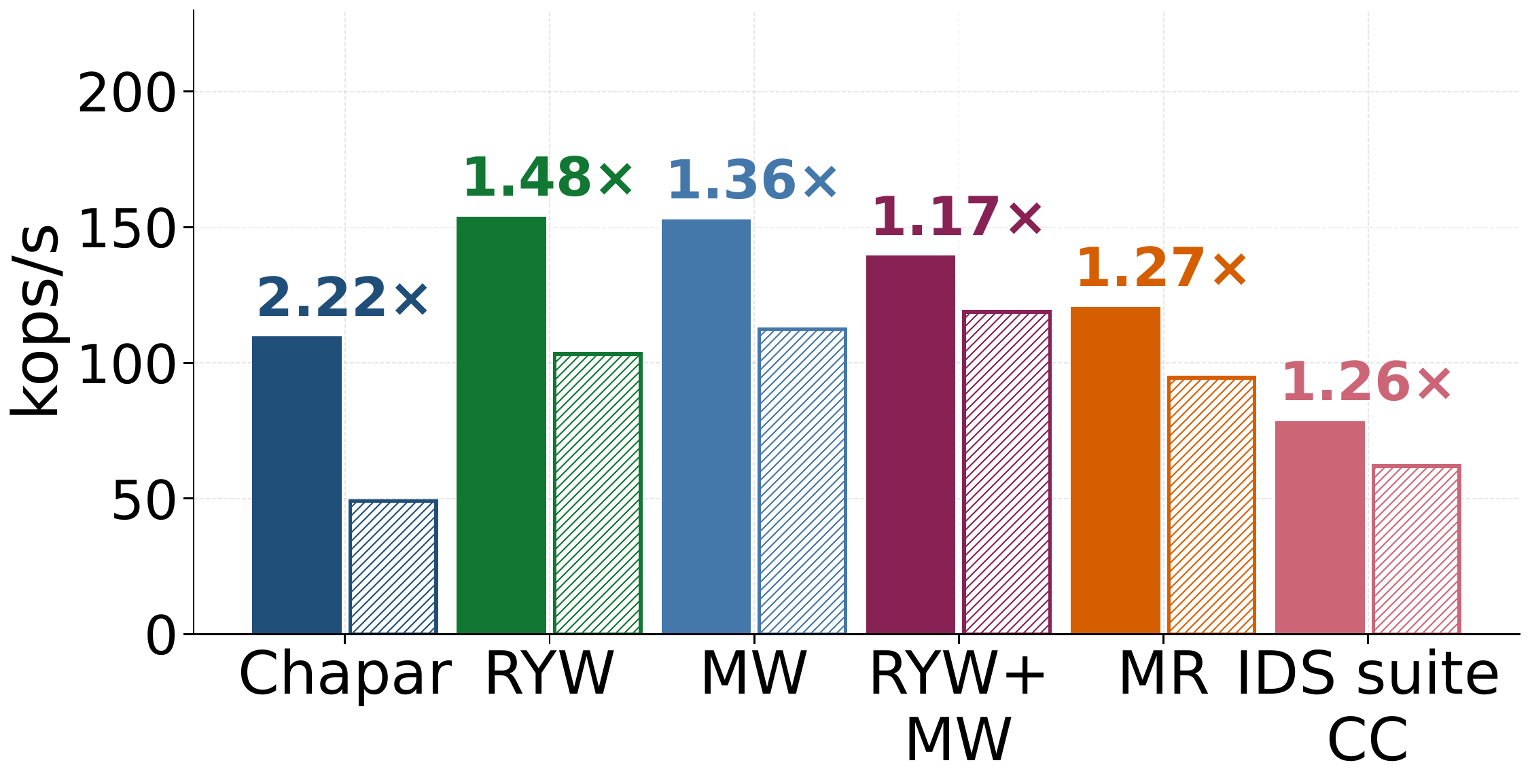} \\[1.3em]
\captionof{figure}{Performance-feedback ablation at $\textit{put}=60\%$: \sys{} (solid) vs.\ no-feedback (hatched).}
\label{fig:eval-bench}
\end{minipage}\hfill
\begin{minipage}[b]{0.30\linewidth}
\centering
\includegraphics[width=\linewidth]{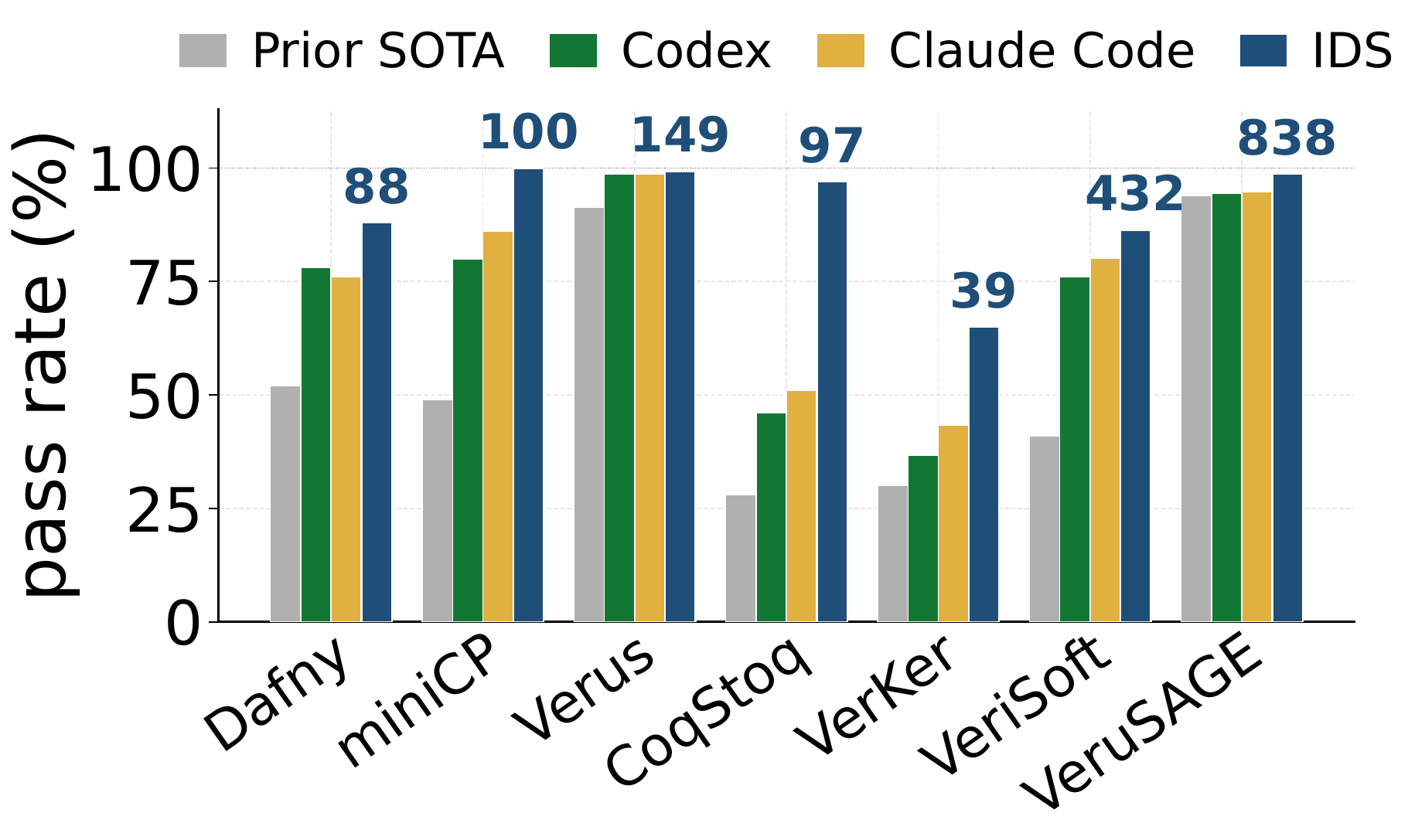} %
\captionof{figure}{Cross-language success rate (\%) on \del{4 }\rev{seven} verification benchmarks;
labels are \# of successes.}
\label{fig:eval-crosslang}
\end{minipage}
\vspace{-8pt}
\end{figure}

The joint step-wise discovery of code and proof is the key design decision behind \sys{}.
We ablate this joint design itself ($-$J), the ISA's proposer ($-$P) and reloader ($-$R), the coordinator's audit step ($-$A), and the \coq{} feedback to the DSA ($-$VF)\rev{, as well as the proof-engineering guidance in the DSA's prompt ($-$G)} across the seven key-value store specs and VERINA's $189$ Lean tasks (Table~\ref{tab:eval-ablation}).\del{ We also evaluate the DSA alone and \sys{}' prover component decoupled from the rest of the system on four cross-language proof benchmarks (DafnyBench, miniCodeProps, Verus-Bench, CoqStoq), where it sets a new state-of-the-art on each.}

\looseness=-1 \textbf{Architecture ablations. \ }
Without joint discovery, only RYW succeeds in a majority of runs: Chapar CC, RYW+MW, \sys{} suite CC, and LCC drop to $0$ out of $3$, and VERINA falls to $85\%$ ($-8\%$ from full \sys{}' $93\%$). This is because a fixed implementation limits the proof phase. On Chapar CC, $-$J commits to the global per-replica state closest to the spec. Full \sys{} reloads to per-key entries when the proof stalls. 
The audit step catches trivial solutions that pass \coq{}'s type-checker: 
on \sys{} suite CC, an agent without auditing once shipped a \texttt{put-guard} returning \texttt{false} unconditionally, with the theorem being trivially satisfied. On the other hand, cells under $-$A count only audited successful runs. 
Ablations of the proposer and reloader each drop pass rate on the four hardest specs (Chapar CC, MR, \sys{} suite CC, LCC), taking VERINA to $79\%$ ($-14\%$) under $-$P and to $88\%$ ($-5\%$) under $-$R. 
The \coq{} feedback to the DSA is the most consequential single component: 
replacing the structured \coq{} diagnostic (goal, hypotheses, tactic backtrace)
with only an accept/reject drops every spec to at most $1$ of $3$ successful runs and VERINA to $58\%$ ($-35\%$). 
\rev{Removing the proof-engineering guidance from the DSA's prompt ($-$G; its tactical playbook, \cref{app:worker}) has minimal effect on efficacy ($19$ vs.\ $20$ of $21$ runs; VERINA $90\%$, $-3\%$): the agent uses \coq{} feedback to rediscover the needed tactics itself, only slower, raising mean time from \rev{$6.9$}\del{$6.8$} to $8.4$ hours and mean cost from \$$106$ to \$$144$. This guidance is thus an efficiency aid.}

\textbf{Performance-feedback ablation. \ }
Without performance feedback ($-$PF), the agent commits to the first implementation and proof that passes the type-checker and misses the course-corrections seen in \cref{sec:eval-perf}. 
As \cref{fig:eval-bench} shows,
full \sys{} is on average $1.42\times$ faster than $-$PF across the six specs with reference baselines (Chapar CC, RYW, MW, RYW+MW, MR, \sys{} suite CC).

\looseness=-1 \textbf{DSA as a general prover. \ }
The DSA's effectiveness is not specific to \sys{}' synthesis loop or to \coq{}: 
applied alone to public verification benchmarks across four languages, it sets a new state-of-the-art on each (\cref{fig:eval-crosslang}, Table~\ref{tab:bench-results}).
It saturates miniCodeProps ($100$ of $100$) and Verus-Bench ($149$ of $150$), and beats prior SOTA by $36$ to $69\%$ on DafnyBench and CoqStoq. 
\sys{}' agentic architecture brings further gain:
full \sys{} adds $+10\%$ on DafnyBench, $+20\%$ on miniCodeProps, and $+51\%$ on CoqStoq over Codex under the same prompt, while a baseline-model swap (Codex $\to$ Claude under the same prompt, no \sys{} architecture) gains at most $6\%$ on any of these four benchmarks. 
\rev{On VerKer, VeriSoftBench, and VeruSAGE-Bench, \sys{} solves $39$ of $60$ (AutoRocq~\cite{tu2026agentic} $18$), $432$ of $500$ (prior SOTA $205$), and $838$ of $849$ (prior SOTA $797$). On VeruSAGE-Bench, \sys{} leads, confirming that it generalizes to Rust systems such as OS kernels, memory allocators, and serialization code.}
\sys{} also reaches state-of-the-art on all three code-and-proof benchmarks: $176/189$ on VERINA (prior SOTA $38$), $65/77$ on AlgoVeri ($2\times$ prior SOTA $31$), and a full $62/62$ on CloverBench (Table~\ref{tab:bench-results}; detail in \cref{app:benchmarks}).
\rev{Beyond per-task benchmarks, \sys{} synthesizes the lock of IronFleet~\cite{ironfleet} in Dafny, verified by Z3 rather than \coq{}: from the unmodified spec, it regenerates and proves all $15$ files with no axioms or assumptions (\cref{app:ironfleet}).}

\section{Discussion and Conclusion}
\label{sec:conclusion}
\label{sec:discussion}

\rev{\textbf{IDS is language- and problem-agnostic. \ } \looseness=-1 IDS needs only (i) a checker that accepts partial proofs, (ii) a programmatic interface an agent can drive, and (iii) extraction to executable code for performance feedback. Any language with a native verifier qualifies (we already run on Dafny, Lean, Verus, and \coq{}; \cref{sec:eval-ablations,app:worker}), as does any language with a reusable formal semantics in a proof assistant (e.g., CompCert's C semantics~\cite{compcert}). A domain needs only a formal specification and, optionally, a performance benchmark. We chose \coq{} for its mature ecosystem, and key-value store specifications\rev{ because the literature already provided one,}\del{: the literature already provided one complete formal specification,} which we extend with six more (\cref{app:suite}).}

\del{\aknew{\textbf{IDS is language- and problem-agnostic. \ } IDS as a technique depends on neither the verification backend nor the problem domain. The backend need only support (i) mechanical checking of partial proofs, (ii) a programmatic interface an LLM agent can drive, and (iii) executable extraction of verified implementations for performance feedback; Lean~\cite{lean} and Verus~\cite{verus} both qualify, and we expect IDS to transfer to either without structural changes. \rev{As with any machine-checked verification, code in a given language needs a verifier: a native checker, which IDS uses directly, swapping only the verifier and surface tools (Dafny, Lean, Verus, and \coq{}; \cref{sec:eval-ablations,app:worker}), or a formal semantics in a proof assistant, often already built and reusable (e.g., CompCert's C semantics in \coq{}~\cite{compcert}).} The problem domain need only admit a machine-checkable formal specification and a way to benchmark candidate implementations. We chose \coq{} for the maturity of its ecosystem, and we chose distributed key-value stores because the literature already provided a complete formal specification for one consistency model; following its structure, we contribute six additional specifications covering other consistency properties (\cref{app:suite}).}}

\aknew{\looseness -2 \textbf{Limitations. \ } Our work has three main limitations. First, \sys{} requires a formal \del{\coq{} }specification as input; writing complete formal specifications for distributed systems requires substantial expert effort\del{, a challenge we return to below}.
\rev{Still, this effort is small relative to the proof effort \sys{} replaces: our six specifications (\cref{app:consistency-specs}) took a few days and are reusable, and the audit rules (\cref{app:auditor}), which took less effort, are shared infrastructure rather than a per-task cost.} Second, the seven specifications we evaluate cover correctness-critical behavior but not the
\del{\rev{broader}}%
full operational envelope of a production KV store; they do not define interfaces for scaling out (adding or removing nodes at runtime), reconfiguration, fault recovery, or informal but practically necessary behavior such as logging and observability endpoints. Given a richer specification that captures these aspects, we see no reason IDS would not apply unchanged, but evaluating this is left to future work. Third, our empirical evaluation \rev{of system-level synthesis }covers the distributed key-value store problem \rev{and the IronFleet lock~\cite{ironfleet} }only, leaving validation on other domains (e.g., OS protocols, cryptographic primitives) to future work. \cref{app:limitations-impacts} discusses additional limitations and broader impacts.}

\rev{\textbf{The specification bottleneck. \ } While \sys{} automates implementation and proof synthesis, writing the specification remains the largest open problem: capturing intended behavior in machine-checkable form is hard, and a guarantee is only as strong as the spec itself~\cite{krentsel2026realityfinalverifierkey}. We envision two directions for closing this gap: \emph{adversarial specification synthesis}, in which an agent interrogates a human stakeholder in natural language, surfacing ambiguities until a formal spec emerges, and \emph{specification extraction}, where an agent recovers the design-agnostic correctness properties of a deployed system so \sys{} can synthesize faster implementations that satisfy them. We leave both to future work.}

\del{\aknew{\textbf{The specification bottleneck. \ } While \sys{} automates the synthesis of verified implementations, writing the formal specification itself remains the largest open problem: accurately capturing a system's intended behavior in a machine-checkable form is hard, and any correctness guarantee is only as strong as the specification it is checked against. We envision two directions for closing this gap. First, \emph{adversarial specification synthesis}, in which an LLM agent collaborates with a human stakeholder in natural language, iteratively probing intent and surfacing ambiguities until a formal specification emerges. Second, \emph{specification extraction from existing systems}, in which an agent recovers the correctness properties of a deployed implementation while leaving the design space open, so that \sys{} can synthesize alternative implementations that satisfy the same properties with better performance. We are actively pursuing both directions, which we plan to report on in future work.}}

\rev{\textbf{Memory across runs. \ } \sys{} treats each run as independent, but lessons from one spec can carry over to related ones. To test this, we record what worked on RYW (the winning abstraction, invariants, and lemma decomposition) in a \emph{skill card} and give it to the agent when it starts RYW+MW. The card only guides the search; \coq{} still checks every proof. With this guidance, RYW+MW still passes in $3/3$ trials, but considerably faster: in $3$ hours for \$$62$ instead of $5$ hours for \$$93$ (\cref{tab:skill-card}). This works because key-value store consistency guarantees share structure. Storing such cards in a vector database would let \sys{} reuse them across many specs; we leave this to future work.}

\textbf{Conclusion. \ } \looseness=-1 Verified systems construction has traditionally required months to years of expert effort;\del{. \sys{} condenses this timeline to mere hours by replacing human labor with compute. By providing an LLM coding agent with fast and exact correctness feedback without false positives, \sys{} transforms the joint code-and-proof construction into an automated search problem.}
\rev{\sys{} condenses it to hours by giving an LLM coding agent fast, exact correctness feedback and turning joint code-and-proof construction into a search problem.}
This shifts vibe coding to verified coding: \rev{systems built by LLMs and machine-checked against a formal specification, thus eliminating silent bugs against the specification.}
\del{by using LLMs to directly build type-safe systems that are machine-checked against a formal specification, effectively eliminating silent bugs and unverified assumptions.}%
Crucially, \del{this methodology}\rev{\sys{}} \rev{can apply} to any domain with a machine-checkable correctness oracle, including OS kernels, compilers, cryptographic protocols, and hardware. Ultimately, \sys{} shows that verified software generation is transitioning from a human-labor bottleneck to a compute-driven approach, paving the way for the broad adoption of verified system synthesis.

\section*{Acknowledgments}
This research is supported by NSF (IFML) CCF-2019844 and gifts from Accenture, AMD, Anyscale, Broadcom Inc., Google, IBM, Intel, Intesa Sanpaolo, Lambda, Mibura Inc., Samsung SDS, and SAP. We thank our Sky Lab and NetSys colleagues at UC Berkeley for fruitful discussions that shaped~this~paper. The authors declare no competing interests.

{\small
\bibliographystyle{plain}
\bibliography{references,specs/refs}
}

\clearpage
\appendix
\crefalias{section}{appendix}
\crefalias{subsection}{subappendix}
\crefalias{subsubsection}{subsubappendix}
\noindent\begin{minipage}{\textwidth}
\vskip 0.12in
\centering{\Large\bf Appendix\par}
\vskip 0.12in
\end{minipage}
\par\vskip 0.05in
\section{Specifications and the \sys{} Suite Dataset}
\label{app:consistency-specs}

Of the seven specifications evaluated in \cref{sec:eval}, one is Chapar's published causal-consistency spec~\cite{chapar} and the other six are \sys{} suite specs we wrote with a formal-verification and distributed-systems expert ($1{,}018$ lines of \coq{} across five spec files; RYW+MW is the joint composition of RYW and MW with no separate spec; \cref{tab:eval-specs-summary}). Formal definitions are in \cref{app:suite}, which also gives two reference implementations of \sys{} suite CC\del{ (vector-clock and dependency-map variants) and}\rev{: a per-client vector-clock store and}\rev{ a Lloyd--Freedman-style protocol~\cite{LloydFreedman13_2}}\del{ the Lloyd--Freedman protocol (the COPS causal-consistency protocol)} as a systems-scale reference.

\begin{table}[h]
\centering
\footnotesize
\setlength{\tabcolsep}{6pt}
\renewcommand{\arraystretch}{1.05}
\caption{The seven specifications. Chapar CC is a published spec; the six \sys{} suite specs are expert-written for this paper. \emph{Property} \rev{summarizes}\del{summarises} the consistency guarantee; \emph{LoC} is lines of \coq{} for the spec definition.}
\label{tab:eval-specs-summary}
\begin{tabular}{@{}l l p{0.50\linewidth} r@{}}
\toprule
Specification    & Family       & Property                                                                              & LoC \\
\midrule
Chapar CC        & Chapar       & cross-client causal consistency, defined over execution traces                        & n/a \\
\midrule
Read-Your-Writes & \sys{} suite & a client reads its own most recent write                                              & $159$ \\
Monotonic Writes & \sys{} suite & a client's writes are observed in order                                               & $158$ \\
Monotonic Reads  & \sys{} suite & a client's reads respect the order of observed writes                                 & $185$ \\
RYW+MW           & \sys{} suite & composition of RYW and MW (no separate spec)                                          & n/a \\
CC  & \sys{} suite & cross-client causal consistency via explicit dependency sets                          & $248$ \\
LCC              & \sys{} suite & causal consistency \rev{parameterized}\del{parameterised} by per-message labels and per-client interest masks  & $268$ \\
\bottomrule
\end{tabular}
\end{table}

\paragraph{Common shape.}
Every spec is a \coq{} \texttt{Module Type} with the same seven method signatures (\texttt{getReq}, \texttt{getGuard}, \texttt{get}, \texttt{getRes}, \texttt{putReq}, \texttt{putGuard}, \texttt{put}), a \texttt{State} type, an \texttt{Update} type, and an operational semantics on client-replica configurations. \sys{} consumes this shape directly.

\paragraph{Generality.}
\sys{} is not tuned to any individual spec: the same agent, prompts, and tools handle every spec we evaluate. Any specification matching the \texttt{Module Type} shape above is a candidate target for the same loop.
\section{Synthesis Results: Per-Spec Artifacts}
\label{app:synth-results}

All seven artifacts pass \coq{}'s kernel with no \texttt{Admitted}, \texttt{Axiom}, \texttt{Hypothesis}, or \texttt{Parameter}. The three CC specs (Chapar CC: $3{,}037$ lines / $79$ lemmas; \sys{} suite CC: $3{,}807$ / $121$; LCC: $3{,}924$ / $128$) are \del{$4$--$6\times$ }\rev{$3$--$6\times$} larger than the four session specs because their proofs span all replicas, not just sender and receiver.

\begin{table}[h]
\centering
\footnotesize
\setlength{\tabcolsep}{4pt}
\renewcommand{\arraystretch}{1.05}
\caption{Per-spec implementation choice and proof artifact size. Lemmas $=$ \texttt{Qed} count; lines via~\texttt{wc~-l}.}
\label{tab:synth-artifacts}
\begin{tabularx}{\linewidth}{@{}l >{\raggedright\arraybackslash}X r r r r r@{}}
\toprule
Specification & \multicolumn{1}{l}{Implementation\rev{ choice}} & \multicolumn{4}{c}{Lines} & Lemmas \\
\cmidrule(lr){3-6}
 & & \rev{Total} & \rev{Impl.} & \rev{Proof} & \rev{Proof share} & \\
\midrule
Chapar CC & per-key store entry with sender clock + dep vector & 3{,}037 & \rev{100} & \rev{2{,}937} & \rev{97\%} & 79 \\
RYW       & per-key (val, client, ts) assoc-list & 622 & \rev{76} & \rev{546} & \rev{88\%} & 25 \\
MW        & per-key (val, client$\to$ts) assoc-list & 722 & \rev{95} & \rev{627} & \rev{87\%} & 30 \\
MR        & per-(key, client) assoc-list, replace-on-write & 1{,}063 & \rev{102} & \rev{961} & \rev{90\%} & 30 \\
RYW+MW    & per-key (val, client, client$\to$ts) assoc-list & 958 & \rev{96} & \rev{862} & \rev{90\%} & 28 \\
\sys{} suite CC & vector clock + per-key VC map + balanced-tree store & 3{,}807 & \rev{126} & \rev{3{,}681} & \rev{97\%} & 121 \\
LCC       & vector clock + per-key VC map + balanced-tree store + per-client label-set filter & 3{,}924 & \rev{126} & \rev{3{,}798} & \rev{97\%} & 128 \\
\bottomrule
\end{tabularx}
\end{table}

\paragraph{Implementation rationale.}

\textbf{Chapar CC.} The $79$ lemmas split: $11$ per-method refinement (one per \texttt{get}/\texttt{put}/\texttt{update}/\texttt{guard}/\texttt{init}), $11$ reachable-state invariants, $7$ causality preservation, $39$ clock and per-key store algebra, and $11$ execution-model helpers.

\textbf{Session guarantees (RYW, MW, MR, RYW+MW).} All four share the same assoc-list skeleton; each parameterizes the per-entry payload to its consistency property: RYW carries (val, client, ts) to detect own writes, MW extends to (val, client$\to$ts) to enforce writer order, MR uses one entry per (key, client) updated in place, RYW+MW combines both. The proof technique is shared too: each session uses heavy case-analysis via \texttt{Nat.eq\_dec} on key and client equality ($80$--$100$ splits per spec), varying only in which payload field the splits land on. Each Get scans only the distinct keys touched, not the full Put history. MR is the hardest session spec because its Put proof splits into four sub-cases (same vs different key, same vs different client), each closed by case-analysis on \texttt{Nat.eq\_dec}.

\textbf{\sys{} suite CC.} A vector clock, a per-key vector-clock map, and a balanced-tree store; the balanced tree keeps Get logarithmic in distinct keys rather than linear in prior Puts. Of the $121$ lemmas: $8$ are the step-by-step simulation, $32$ are refinement-relation invariants for the state components (cell, replica, snapshot) and their preservation under updates, $19$ are vector-clock algebra; the remaining $62$ are update, lookup, owner, and dependency-tracking algebra. Get's proof closes by applying the relation that ties the implementation's vector clocks plus per-client owner lists (per-key tracking of which client most recently wrote each entry) to the spec's dependency sets.

\textbf{LCC.} A vector clock, a per-key vector-clock map, and a balanced-tree store, with dependency entries tagged with a label (a $4$-tuple of client, timestamp, key, label) and projections at Get filtered by the requesting client's interest mask. The proof closes by step-by-step simulation; the simulation relation ties the implementation's label-tagged vector-clock representation to the spec's label-filtered dependency sets. Of the $128$ lemmas: $6$ are the step-by-step simulation, $25$ are refinement-relation invariants and extensions for state components (cells, replicas, snapshots, key-snapshots), $19$ are vector-clock algebra; the remaining $78$ are update, lookup, owner, and dependency-tracking algebra.

\paragraph{Synthesis trajectories.}
The artifacts above are not the first design \sys{} attempted. In each case, the proof stalled on a goal that resisted splitting under the spec's natural state representation; the ISA's reloader abandoned the branch\del{, and the proposer pivoted to}\rev{ and chose} a representation under which the goal decomposed.

\textbf{Chapar CC.} The hardest case in any causal-consistency proof shows that whenever a replica delivers a remote message, it has already observed every update the message transitively depends on. \sys{}' first attempt kept the global state representation closest to the spec, under which the delivery lemma had to range over every in-flight message at every replica together, and the DSA could not split it. After the search stalled, the reloader abandoned that branch and \del{the proposer revised the design to }\rev{a fresh DSA rebuilt on} a per-key store with per-entry sender, sender-clock, and dependency vector. The proposer then split the delivery invariant into three sub-lemma classes (per-message, per-store-entry, per-replica-clock), each closing under the new representation. The representation was chosen by what split the proof, not by performance; the per-put performance win reported in \cref{sec:eval-perf} is a consequence.

\textbf{Monotonic reads.} Both agent baselines, and \sys{}' first attempt, stalled on the function-from-key-to-value representation. After the reloader abandoned that branch, the agent revised the design to a flat list of per-(key, client) entries updated in place, after which the put proof split into four sub-cases on key and client equality.

\textbf{\sys{} suite CC and LCC.} The winning move is to break the simulation goal into per-message and per-store-entry sub-lemmas, each closing against the per-key store and label-tagged dependency entries respectively. The baselines try the combined goal in one shot and \rev{exhaust}\del{time out within} the budget.

\rev{\paragraph{Proof-only baseline.} \cref{tab:proof-only} details the proof-only baseline from \cref{sec:eval-synthesis}. Given the implementation, Claude Code reaches a majority of runs on MR ($0/3 \to 2/3$) and RYW+MW ($1/3 \to 3/3$). \sys{} closes all seven because it discovers implementation and proof jointly. A finished implementation records the final state representation but not the trial and error behind it; in \sys{}, that search closes~the~proof.}

\begin{table}[ht]
\centering
\footnotesize
\setlength{\tabcolsep}{6pt}
\renewcommand{\arraystretch}{1.05}
\caption{\rev{Proof-only baseline: successes from $3$ runs per spec; \emph{Total} counts specs passing at least $2/3$. \emph{Spec only} is the \cref{tab:eval-synthesis} setting, where the agent writes both implementation and proof; with \emph{spec + \sys{} impl.}, Claude Code also receives the implementation \sys{} synthesized and writes only the proof.}}
\label{tab:proof-only}
\vspace{2pt}
\rev{%
\begin{tabular}{l cc c c}
\toprule
 & \multicolumn{2}{c}{Spec only} & Spec + \sys{} impl. & Spec only \\
\cmidrule(lr){2-3} \cmidrule(lr){4-4} \cmidrule(lr){5-5}
Specification & Codex & Claude Code & Claude Code & \sys{} \\
\midrule
RYW             & 3/3 & 3/3 & 3/3 & 3/3 \\
MW              & 2/3 & 3/3 & 3/3 & 3/3 \\
MR              & 0/3 & 0/3 & 2/3 & 3/3 \\
RYW+MW          & 0/3 & 1/3 & 3/3 & 3/3 \\
Chapar CC       & 0/3 & 0/3 & 0/3 & 3/3 \\
\sys{} suite CC & 0/3 & 0/3 & 0/3 & 2/3 \\
LCC             & 0/3 & 0/3 & 0/3 & 3/3 \\
\midrule
Total           & 2/7 & 2/7 & 4/7 & 7/7 \\
\bottomrule
\end{tabular}}
\end{table}

\rev{\paragraph{Failure modes.} \cref{tab:failure-modes} details the failure modes from \cref{sec:eval-synthesis}, how \sys{} responds to each, and the ablation that isolates it.}

\begin{table}[ht]
\centering
\footnotesize
\setlength{\tabcolsep}{4pt}
\renewcommand{\arraystretch}{1.1}
\caption{\rev{Failure modes, how \sys{} responds, and the ablation (\cref{tab:eval-ablation}) that isolates each. \sys{} recovers from the first three; the fourth ends the run at the budget.}}
\label{tab:failure-modes}
\vspace{2pt}
\rev{%
\begin{tabularx}{\linewidth}{@{}>{\raggedright\arraybackslash}p{1.55cm} >{\raggedright\arraybackslash}X >{\raggedright\arraybackslash}X >{\raggedright\arraybackslash}X@{}}
\toprule
Failure mode & Example & How \sys{} responds & Ablation (\cref{tab:eval-ablation}) \\
\midrule
Tactical stall & On Chapar CC the delivery invariant would not close as a single piece. & The proposer splits it into smaller lemmas (per-message, per-store-entry, per-replica-clock), each of which then closes (\cref{app:guide}). & Without the proposer: Chapar CC, MR, \sys{} suite CC, LCC fall to $0/3$. \\
\addlinespace
Strategic dead-end & On Chapar CC the initial global-state design left the delivery lemma unprovable. & The reloader drops that design and the DSA rebuilds on a per-key store, reaching a proof the first design could not (\cref{app:resetter}). & Without the reloader: Chapar CC, \sys{} suite CC, LCC fall to $0/3$; MR to $1/3$. \\
\addlinespace
Vacuous solution & On \sys{} suite CC one run made the put guard always return false, passing \coq{} vacuously (\cref{sec:eval-ablations}). & The audit rejects these before they count as solved, so the ISA returns to design for a genuine implementation and proof (\cref{app:auditor}). & Without the audit: MR, \sys{} suite CC, LCC fall to $0/3$. \\
\addlinespace
Budget exhaustion & The hardest specs hit the per-spec time cap before the proof closes. & The run is reported as a failure at the cap. & Not an ablation: $1$ of $21$ KV runs and \rev{$41$ of $778$ tasks on seven of the ten benchmarks}\del{$41$ of $778$ tasks} hit the cap (\cref{app:benchmarks}). \\
\bottomrule
\end{tabularx}}
\end{table}

\rev{\paragraph{Memory across runs.} \cref{tab:skill-card} details the skill-card experiment from \cref{sec:discussion}. Each card is stored under its spec type, and a new spec receives the most similar card in its prompt. The card only guides the search: \coq{} still checks every proof, so soundness is unaffected.}

\begin{table}[ht]
\centering
\footnotesize
\setlength{\tabcolsep}{6pt}
\renewcommand{\arraystretch}{1.05}
\caption{\rev{RYW+MW with and without the skill card from RYW (same model and budget). The no-card row is the \sys{} result in \cref{tab:eval-synthesis}.}}
\label{tab:skill-card}
\vspace{2pt}
\rev{%
\begin{tabular}{l c c c}
\toprule
 & Pass rate$\uparrow$ & Hours$\downarrow$ & Cost (\$)$\downarrow$ \\
\midrule
No card       & 3/3 & 5 & 93 \\
Card from RYW & 3/3 & \textbf{3} & \textbf{62} \\
\bottomrule
\end{tabular}}
\end{table}

\FloatBarrier
\section{Implementing Distributed-KV Specifications with Coding Agents \\(No Proof)}
\label{app:llm-only-synthesis}

\paragraph{Setup.}
We test Codex (\texttt{gpt-5.4}) on four KV-store consistency properties (RYW, MW, MR, and CC). For each property, Codex generates 100 candidate implementations under two prompt conditions, then picks its single best candidate. That selected implementation is the agent's output for that (property, condition).

\emph{Setting (1) Spec-Given.} Codex receives the formal \coq{} specification and the module-type signature.

\emph{Setting (2) NL-Only (``vibe coding'').} Codex receives only a one-paragraph natural-language description of the property (reproduced verbatim below) and the module-type signature.

Each of the 100 generation sessions runs independently and can compile, edit, and test its own code. The selector pass is also Codex, with the same compile and test access. The selected implementation must use bounded state, cannot copy any reference implementation, and cannot use \texttt{Admitted}, \texttt{Axiom}, \texttt{Hypothesis}, or \texttt{Parameter} (so the agent cannot stub out unfilled work). The agent does not write any refinement proof.

We evaluate the selected implementation in two stages.
\emph{Stage~1 (test).} We run the implementation on an adversarial multi-client scenario; if it accepts an operation the abstract specification rejects on this scenario, Stage~1 fails.
\emph{Stage~2 (proof).} If Stage~1 passes, \sys{} attempts the refinement proof against the abstract specification. If \sys{} cannot close the proof within budget, Stage~2 fails.
We split into two stages because proof attempts are much more expensive than scenario testing. If Stage~1 fails, Stage~2 fails too: an implementation that breaks the spec on a single trace cannot be proven correct on every trace.

\paragraph{Property statements (LLM-visible in Setting~2).}
\begin{description}
\item[RYW (Read-Your-Writes).] Each client must observe its own previous writes. If a client writes a value to a key and later reads that key in the same session, the read must return the client's value or a more recent one.
\item[MW (Monotonic Writes).] Writes by the same client are applied at every replica in the order the client issued them.
\item[MR (Monotonic Reads).] Successive reads by the same client observe non-decreasing replica states: once a client has seen a value, later reads must not appear older.
\item[CC (Causal Consistency).] If operation $A$ causally precedes operation $B$ (either $A$ comes before $B$ in the same client's session, or $B$ reads a value written by $A$), every client that sees $B$ must also have seen $A$.
\end{description}

\begin{table}[h]
\centering
\caption{Outcome of best-of-$N{=}100$ implementation-only synthesis with Codex. \textbf{1} = passed, \textbf{0} = failed. Stage~1: the selected implementation runs through a fixed adversarial multi-client scenario without violating the abstract specification on that scenario. Stage~2: \sys{} builds a refinement proof of the implementation against the abstract specification within budget. Cost is the \texttt{gpt-5.4} token cost for the 100 generations plus the selection pass.}
\label{tab:llm-only-synthesis}
\small
\begin{tabular}{l|ccc|ccc}
\toprule
 & \multicolumn{3}{c|}{Setting (1) Spec-Given} & \multicolumn{3}{c}{Setting (2) NL-Only} \\
Spec & Stage~1 pass & Stage~2 pass & Cost & Stage~1 pass & Stage~2 pass & Cost \\
\midrule
RYW & 1 & 1 & \$30.50 & 1 & 0 & \$25.77 \\
MW  & 1 & 0 & \$32.10 & 1 & 0 & \$27.33 \\
MR  & 0 & 0 & \$35.20 & 0 & 0 & \$29.91 \\
CC  & 0 & 0 & \$47.80 & 0 & 0 & \$40.79 \\
\bottomrule
\end{tabular}
\end{table}

\paragraph{Result and takeaway.}
Best-of-$N=100$ produces a verified implementation on 1 of 4 properties when given the formal specification (Setting~1, RYW), and on 0 of 4 from natural language alone (Setting~2). The two simpler session properties (RYW and MW) consistently pass the scenario test in both settings, but only RYW with the formal spec also passes the refinement proof. The harder properties (MR and CC) fail even the scenario test in both settings: the textbook patterns Codex selects (vector-clock dominance, last-write-wins) do not implement the per-key or causal-order behavior these specs demand. Two failure modes appear: (a) implementations that fail on a single adversarial trace, and (b) implementations that pass the trace but cannot be proven correct on every reachable state. \sys{} avoids both by integrating the refinement proof into synthesis rather than running it as a post-hoc check; in the main evaluation it closes the proof on all 7 specifications.
\section{Additional Performance Results and Throughput Breakdown}
\label{app:perf}
\label{app:perf-harness}

\sys{}' gains over the reference implementations all share one mechanism: \sys{}' state representations bound per-op cost by the spec, while the references' per-op cost grows with workload size. \cref{app:perf-curves} measures this on all five \sys{} suite specs with cluster-runnable implementations (RYW, MR, MW, RYW+MW, \sys{} suite CC), shows Chapar CC at the top of the figure for comparison, and adds an illustrative LCC row (labeled causal consistency: each client subscribes to a set of topic labels and only sees causality enforced on those). Across the \sys{} suite, throughput tracks consistency strength (RYW $>$ MW $>$ RYW+MW $>$ MR $>$ LCC $>$ \sys{} suite CC at put $=50\%$) because stricter specs require more checking work per operation. Chapar CC and \sys{} suite CC are both causal-consistency variants but their reference implementations are designed differently (Chapar's reads one position out of a flat list of per-write dependencies, while \sys{} suite's compares an entire per-client clock vector and atomically updates the writer's slot on every Put), so the two CC rows show different per-op work at the same correctness level. \cref{app:perf-cc,app:perf-mr} decompose the gap on Chapar CC and Monotonic~Reads.

\paragraph{Setup details (extending \cref{sec:eval-setup}).}
Cluster zone: \texttt{us-central1-f}. Each VM is an \texttt{e2-standard-2} (2 vCPU, 8\,GB RAM) on Debian 12. Toolchain: OCaml 4.13.1 (dune 3.10) and \rev{Coq~8.18.0}\del{\coq 8.18.0}. Throughput is the sum across the $\text{NW}=4$ workers of $N / t_w$, where $t_w$ is each worker's wall-clock time and $N$ is its ops count, taking the median over three runs per cell; the put-rate sweep uses $N=1000$ (4{,}000 cluster ops total per run) and the scaling sweep varies $N \in \{1{,}000, 2{,}000, 5{,}000, 20{,}000\}$. Per-operation latency is bracketed inside the runtime around \texttt{put\_method}/\texttt{get\_method} and merged across workers; p99 is the 99th percentile of the pooled per-op latency distribution. Peak memory is the worker maximum of \texttt{Maximum resident set size} from \texttt{/usr/bin/time -v}, taking the median over three runs. Workload parameters: $\texttt{key\_range}=50$, $\texttt{val\_range}=100{,}000$, fixed seed $42$. The cluster uses a Chapar-style UDP runtime (\texttt{runtime\_doctex.ml}) adapted to the seven-method doctex \texttt{AlgDef} interface; we patched a socket bug for OCaml 4.13 / Debian 12.

\subsection{Per-protocol curves: all five \sys{} suite specs}
\label{app:perf-curves}

\begin{figure*}[p]
\centering
\includegraphics[width=0.95\linewidth]{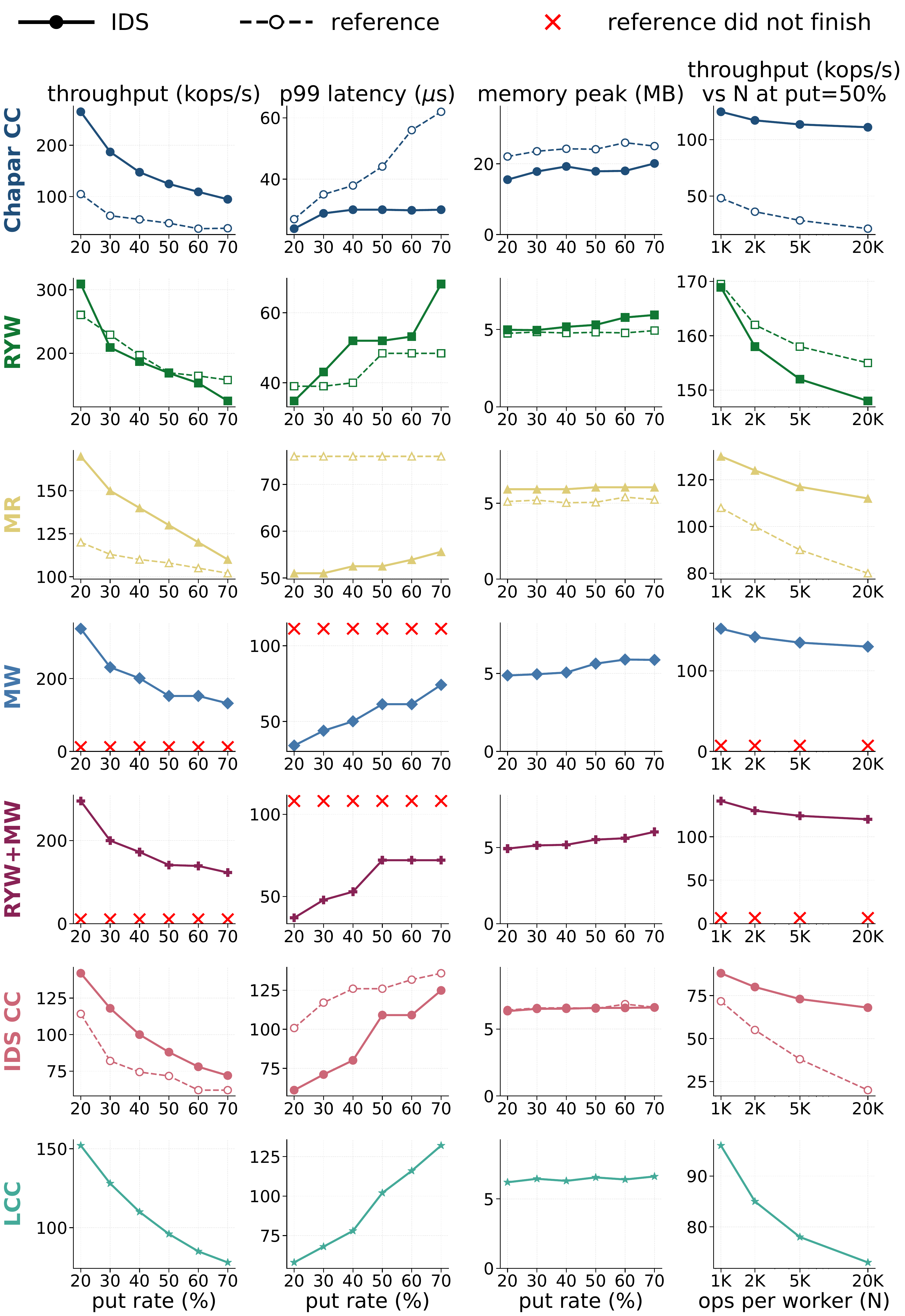}
\caption{Per-protocol performance. Top row is Chapar CC (Chapar paper protocol) shown for comparison; the next five rows are the \sys{} suite specs (RYW, MR, MW, RYW+MW, \sys{} suite CC); the bottom LCC row is a single-line illustrative trajectory of labeled causal consistency. Columns: throughput (kops/s), p99 latency ($\mu$s), memory peak (MB), and throughput vs ops-per-worker $N$ at put $=50\%$. Per-protocol $y$-axes are dedicated so each row's range is visible. Solid $=$ \sys{}; dashed $=$ reference (function-as-map). Red $\times$ marks runs that hit the $180$\,s wall-clock cap; on MW and RYW+MW the reference times out at every put rate (the reference's read-then-write check on a function-as-map is the cause; see \rev{\cref{app:perf-curves}}\del{below}).}
\label{fig:eval-perf-appendix}
\end{figure*}

\paragraph{Result.}
Read-then-write specs cause the reference to exceed our $180$\,s wall-clock cap at every put rate measured: in \cref{fig:eval-perf-appendix} the reference's MW and RYW+MW rows show red $\times$ at every cell, with no measurable throughput. \sys{} runs cleanly throughout: $152$ kops/s on MW and $139$ kops/s on RYW+MW at put $=60\%$. The mechanism is a closure chain: the reference stores per-replica state as a function from key to value (a function-as-map); after extraction to OCaml, each prior \texttt{Put} adds a nested closure that a \texttt{Get} must unwrap. The read-then-write check re-evaluates this chain on every \texttt{Put}; combined with UDP packet reordering and the protocol's no-retry compare-and-swap, throughput collapses to zero. RYW has no read-then-write check, so its reference is competitive with \sys{} across the put-rate sweep: within \rev{$21\%$}\del{roughly $20\%$} in either direction depending on workload mix, with \sys{} winning at low put rate (\rev{put $=20\%$}\del{pct$=20$}) and the reference faster at high put rate (\rev{put $=60$--$70\%$}\del{pct$=60$--$70$}). The reversal is a per-Put cost trade: \sys{}' assoc-list rewrites its spine on every \texttt{Put} ($O(K)$ cell allocations, $K$ = unique keys touched), while the reference's function-as-map prepends a single closure layer per \texttt{Put} ($O(1)$ heap cell). At low put rate the assoc-list's $O(K)$ \texttt{Get} lookup beats the reference's closure-chain walk; at high put rate the per-\texttt{Put} allocation cost dominates. MR shows the same chain-walk cost on \texttt{Get} (per-layer slope quantified in \cref{app:perf-mr}); \sys{} suite CC follows the same mechanism with \sys{}' balanced-tree store keeping Get logarithmic in distinct keys. Memory: \sys{}' assoc-list pays $0.1$--$1.0$ MB more on RYW (per-entry overhead) and runs in the same $5$--$6$ MB envelope as the reference on the other specs where the reference completes; on MW and RYW+MW the reference produces no memory measurement (it never completes a run). All memory traces are at $N=1{,}000$ ops/worker; \sys{} suite CC's vector-clock state is $O(\text{\#clients})$, so its trace is flat across the put-rate sweep at this workload size. At substantially larger $N$ or more clients, the message-bus and per-cell deps would push memory upward. The rightmost column (throughput vs $N$ at put $=50\%$) confirms the mechanism on every spec: the reference's throughput drops as $N$ grows because each new \texttt{Put} extends the chain, while \sys{}' assoc-list and balanced-tree structures keep throughput approximately flat. Across the put-rate sweep, both implementations' throughput falls and p99 latency rises as put-rate increases because each \texttt{Put} is broadcast to every replica, so more writes mean more in-flight messages and longer per-op queue waits.

\paragraph{Course correction on MW.}
Without performance feedback, \sys{} ships the first verifying design: a balanced-tree store. With performance feedback, the agent runs both candidates in its microbench harness and converges on the assoc-list, which the cluster benchmark harness confirms at $152$k ops/s at put $=60\%$ (\cref{fig:eval-perf-appendix}). Performance feedback changes the search policy, not the correctness criterion; both candidates are verified \sys{} suite refinements.

\subsection{Throughput breakdown: Chapar CC}
\label{app:perf-cc}

\begin{figure*}[t]
\centering
\includegraphics[width=\linewidth]{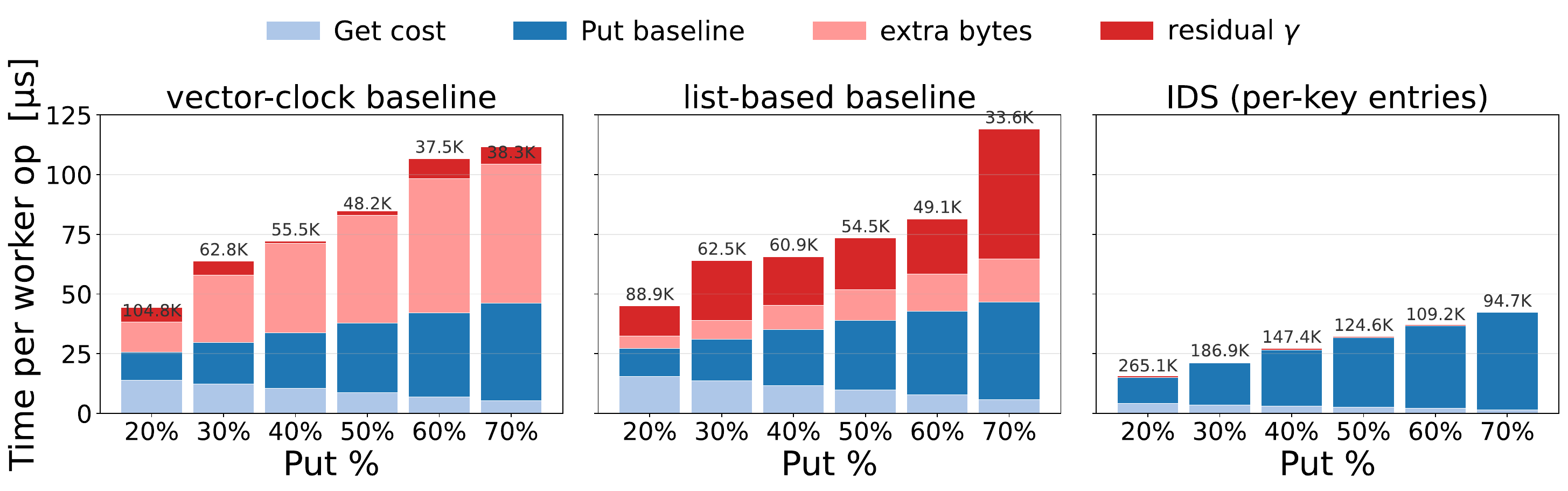}
\caption{Per-component decomposition of Chapar's two published \rev{references}\del{baselines} (vector-clock, list-based) and \sys{}' Chapar implementation (per-key store entries, used here as internal pivot). Bar height equals $4 / \text{(published cluster throughput)}$ in $\mu$s per worker op. Each bar splits into local Get cost, the per-key-entries Put baseline (same for all), per-algorithm extra wire bytes, and a per-algorithm residual $\gamma$. Segments sum to the published value at every cell by construction\rev{; where $\gamma<0$ (e.g., vector-clock at $20$, $50$, and $70\%$), the red segment shows $|\gamma|$ and is subtracted from the bar top}.}
\label{fig:chapar-decomp}
\end{figure*}

\paragraph{Setup.}
We attribute each algorithm's per-op time to four pieces: its Get cost (varies by algorithm), a shared Put baseline (per-key entries: $58.4\,\mu$s, the smallest of the three), a wire-bytes term proportional to extra bytes per Put ($\beta = 18\,\mu$s per byte, fit from vector-clock vs per-key entries), and a CPU residual $\gamma$ that captures whatever the wire-bytes term does not explain. Marshaled bytes per Put on the cluster: $44.81$ (vector-clock), $41.05$ (list-based), $39.62$ (per-key entries).

\paragraph{Result.}
Vector-clock loses on wire bytes; list-based loses on CPU. \textbf{Vector-clock} ships $+5.2$ bytes per Put over per-key entries, mapping to $+47\,\mu$s of per-op work at put $=50\%$; the residual $\gamma$ is $-1.75\,\mu$s, well below the bytes term. Almost the entire gap is wire bytes. \textbf{List-based} ships only $+1.4$ extra bytes per Put on average, yet $\gamma = +21.6\,\mu$s at put $=50\%$. The list-based replica stores the full history of delivered messages; receivers walk this list on every delivery (causality check), and \texttt{Get} walks it again to find the most-recent value for the requested key. Vector-clock and per-key entries do constant-time element-wise lookups.

\subsection{Throughput breakdown: Monotonic Reads}
\label{app:perf-mr}

\begin{figure*}[t]
\centering
\includegraphics[width=\linewidth]{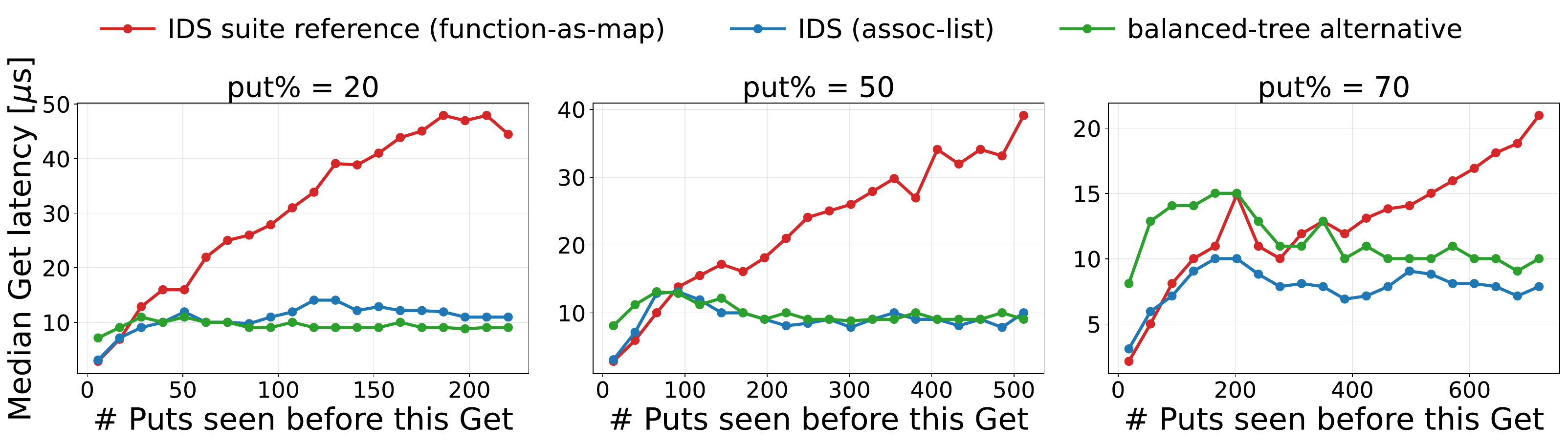}
\caption{\texttt{Get} latency vs.\ closure depth (number of \texttt{Put}s preceding each \texttt{Get} in the same worker trace) for the three Monotonic Reads implementations at three put rates. The \sys{} suite reference's \texttt{Get} latency grows linearly with depth (slope $0.21\,\mu$s per closure layer at put $=20\%$, $R^2 = 0.95$). \sys{}' assoc-list slope is an order of magnitude smaller; the balanced-tree variant's slope is within statistical noise at every put rate.}
\label{fig:mr-depth}
\end{figure*}

\paragraph{Setup.}\looseness=-1
Three MR implementations on the same runtime and identical wire format ($92.83$ B per \texttt{Put} on average): the \sys{} suite reference (function-as-map), \sys{} (assoc-list), and a balanced-tree alternative. Throughput differences are CPU-side. The balanced tree is a control: it tests whether \sys{}' win comes from the assoc-list specifically or from any structure that does not walk a closure chain on \texttt{Get}.

\paragraph{Where the cost lands.}
The reference pays on \texttt{Get}; the refinements pay on \texttt{Put}. At put $=50\%$:
\begin{center}
\begin{tabular}{lrr}
\toprule
implementation & $T_{\mathrm{get}}$ ($\mu$s) & $T_{\mathrm{put}}$ ($\mu$s) \\
\midrule
\sys{} suite reference (function-as-map) & $24.83$ & $15.40$ \\
\sys{} (assoc-list)                & $10.81$ & $16.81$ \\
balanced-tree alternative          & $11.40$ & $21.12$ \\
\bottomrule
\end{tabular}
\end{center}
Regressing \texttt{Get} latency against closure depth (number of \texttt{Put}s preceding a \texttt{Get} in the same worker trace) confirms the reference's chain walk: slope $0.21\,\mu$s/layer at put $=20\%$ ($R^2 = 0.95$, \cref{fig:mr-depth}). Both refinements stay flat across every put rate measured ($|\text{slope}| < 0.025\,\mu$s/layer): bounded data structure, no chain to walk. The balanced tree's extra Put cost comes from rebalance.

\paragraph{N-scaling at put $=50\%$.}
The reference's throughput drops as ops-per-worker $N$ grows; \sys{} stays~nearly~flat:
\begin{center}
\begin{tabular}{lrrrr}
\toprule
$N$ & $1{,}000$ & $2{,}000$ & $5{,}000$ & $20{,}000$ \\
\midrule
\sys{} suite reference (function-as-map) & $108{,}000$ & $100{,}000$ & $90{,}000$ & $80{,}000$ \\
\sys{} (assoc-list)                & $130{,}000$ & $124{,}000$ & $117{,}000$ & $112{,}000$ \\
\bottomrule
\end{tabular}
\end{center}
At $N=1{,}000$ \sys{} is $1.20\times$ faster; at $N=20{,}000$ the gap widens to $1.40\times$. Throughput drop from $N=1{,}000$ to $N=20{,}000$: reference $1.35\times$, \sys{} $1.16\times$. The reference's per-\texttt{Get} closure walk grows linearly with prior \texttt{Put}s, so its per-op cost rises with $N$; the assoc-list's lookup is bounded by the number of distinct keys in the workload (50 here), so its per-op cost is $N$-independent and only message-bus pressure causes the mild decline.

\section{Results on Proof, Annotation, and Code-and-Proof Benchmarks}
\label{app:benchmarks}

\sys{} beats every prior method on the \del{four }\rev{seven} proof and annotation benchmarks, and reaches $\mathbf{176/189}$ on VERINA against prior SOTA $38$ and agentic baselines Codex $136$, Claude Code~$149$~(\mbox{\cref{tab:bench-results}}).

\begin{table}[t]
\centering
\footnotesize
\setlength{\tabcolsep}{5pt}
\renewcommand{\arraystretch}{1.05}
\caption{Pass counts on verification benchmarks. Bold $=$ best per \del{column }\rev{row}. The first \del{four }\rev{seven} are proof-only (impl and spec given); on these, \sys{} reduces to its DSA component since the ISA has no design space to search. The last three are \del{concurrent }\rev{joint} code-and-proof tasks. Per-benchmark slices, prior-SOTA references, and CloverBench note in \cref{app:benchmarks}.}
\label{tab:bench-results}
\begin{tabular}{l r c c c c}
\toprule
\del{Method }\rev{Benchmark} & \rev{Tasks} & Prior SOTA & Codex & Claude Code & \textbf{\sys{}} (ours) \\
\midrule
DafnyBench     & \rev{100} & 52  & 78  & 76  & \textbf{88} \\
miniCodeProps  & \rev{100} & 49  & 80  & 86  & \textbf{100} \\
Verus-Bench    & \rev{150} & 137 & 148 & 148 & \rev{\textbf{149}} \\
CoqStoq        & \rev{100} & 28  & 46  & 51  & \textbf{97} \\
\rev{VerKer}         & \rev{60}  & \rev{18}  & \rev{22}  & \rev{26}  & \rev{\textbf{39}} \\
\rev{VeriSoftBench}  & \rev{500} & \rev{205} & \rev{380} & \rev{401} & \rev{\textbf{432}} \\
\rev{VeruSAGE-Bench} & \rev{849} & \rev{797} & \rev{801} & \rev{805} & \rev{\textbf{838}} \\
\midrule
VERINA         & \rev{189} & 38  & 136 & 149 & \textbf{176} \\
AlgoVeri       & \rev{77}  & 31  & 51  & 56  & \textbf{65} \\
CloverBench    & \rev{62}  & --  & \textbf{62}  & \textbf{62}  & \textbf{62} \\
\bottomrule
\end{tabular}
\end{table}

\subsection{Proof and annotation benchmarks}

\paragraph{Setup.}
The model receives a complete task specification (signature, preconditions, postconditions, and code where applicable) and must produce verifier-acceptable output: proof annotations on DafnyBench~\cite{dafnybench} and Verus-Bench~\cite{autoverus}, or a complete proof on miniCodeProps~\cite{minicodeprops}\rev{,}\del{ and} CoqStoq~\cite{rango}\rev{, VerKer~\cite{tu2026agentic}, VeriSoftBench~\cite{xin2026verisoftbench}, and VeruSAGE-Bench~\cite{yang2025verusage}} (modality in \cref{tab:bench-modalities}). For each benchmark we use the slice the prior SOTA method published its number on:
\begin{itemize}
\item \textbf{DafnyBench}: $100$ hardest tasks; prior SOTA is DafnyPro~\cite{dafnypro} (from the DafnyBench paper).
\item \textbf{miniCodeProps}: $100$ tasks from the sorting split; prior SOTA is COPRA~\cite{copra}.
\item \textbf{Verus-Bench}: full $150$ tasks; prior SOTA is AutoVerus~\cite{autoverus}.
\item \textbf{CoqStoq}: $100$ hardest tasks; prior SOTA is Rango~\cite{rango}.
\rev{\item \textbf{VerKer}: $60$ Linux-kernel tasks from AutoRocq's case study, scored with AutoRocq's own unmodified script; prior SOTA is AutoRocq's strongest variant (using CoqHammer)~\cite{tu2026agentic}.}%
\rev{\item \textbf{VeriSoftBench}: full $500$ Lean~4 proof obligations from $23$ formal-methods repositories, with the same $15$-minute budget per task for \sys{}, Codex, and Claude Code (\sys{}: about \$$569$ in total); prior SOTA is non-agentic multi-pass sampling with \rev{Gemini 3 Pro}\del{Gemini-3 Pro}, which does not report its cost. We use VeriSoftBench rather than PutnamBench~\cite{tsoukalas2024putnambench} or LeanEval~\cite{lean_eval_2026}, which target mathematical theorem proving.}%
\rev{\item \textbf{VeruSAGE-Bench}: full $849$ tasks from eight real-world Verus projects, each with the target proof function body removed; prior SOTA is Opus~4.5.}
\end{itemize}

\paragraph{Result.}
\sys{} reaches $88/100$ on DafnyBench, $100/100$ on miniCodeProps, $149/150$ on Verus-Bench, \del{and }$97/100$ on CoqStoq\rev{, $39/60$ on VerKer, $432/500$ on VeriSoftBench, and $838/849$ on VeruSAGE-Bench} (\cref{tab:bench-results}), beating the strongest prior on all \del{four }\rev{seven}. The largest absolute gap is on CoqStoq ($+69$\rev{\%} over Rango); \sys{} hits a perfect $100/100$ on miniCodeProps, and on Verus-Bench all three LLM methods are within \del{$1\%$ }\rev{$2\%$} of saturation (\sys{} $149$, Codex $148$, Claude Code $148$ of $150$). \rev{VeruSAGE-Bench is also near saturation (Sonnet~4.5 reaches $81\%$, Opus~4.5 $93.9\%$), yet \sys{} still leads with $98.7\%$. On VeriSoftBench, \sys{} substantially improves on the non-agentic prior SOTA ($432$ vs.\ $205$ of $500$).} Failures reflect per-task budget caps and proofs requiring tactics or library-specific lemmas the agent did not discover within budget (\cref{app:proof-method}).

\subsection{Code-and-proof benchmarks}

\paragraph{Setup.}
A second, recent line of benchmarks evaluates joint code-and-proof synthesis on single-function algorithmic tasks (sorting, search, list manipulation, basic arithmetic like factorial/gcd/Fibonacci): the model receives a natural-language description, signature, and spec, and produces both the code and a proof of correctness. These tasks are strictly simpler than the distributed-system specs of \cref{sec:eval-synthesis}: no concurrency, no replication, no multi-step operational semantics.
\begin{itemize}
\item \textbf{VERINA}~\cite{verina}: $189$ Lean tasks; prior SOTA is iterative refinement with Lean compiler feedback ($64$ rounds), reported in the VERINA paper.
\item \textbf{AlgoVeri}~\cite{algoveri}: $77$ problems, each expressed in Dafny, Verus, and Lean; prior SOTA $31/77$ (\rev{Gemini 3 Flash}\del{Gemini-3 Flash}, Dafny track, reported in the AlgoVeri paper).
\item \textbf{CloverBench}~\cite{clover}: $62$ Dafny tasks. We omit a Prior-SOTA cell because Clover (the only prior published method) evaluates a different task: its consistency-checker filters pre-existing correct triples (an $87\%$ acceptance rate, with $13\%$ false negatives where it rejects a known-correct triple). Our column reports end-to-end synthesis: given the spec, the model generates code and annotations, and the Dafny verifier accepts.
\end{itemize}

\begin{table}[t]
\centering
\footnotesize
\setlength{\tabcolsep}{6pt}
\renewcommand{\arraystretch}{1.15}
\caption{What each cross-language benchmark gives the model and what the model must produce. \cmark{} = given as input; \xmark{} = the model must generate it. ``Spec'' covers preconditions and postconditions (or the theorem statement, where applicable). For DafnyBench the implementation body is given; only Dafny verification \emph{hints} (loop invariants, assertions, decreases clauses) are stripped. miniCodeProps\rev{, VeriSoftBench,}\del{ and} CoqStoq\rev{, and VerKer} give the spec as a Lean / \del{Coq }\rev{\coq} theorem statement.}
\label{tab:bench-modalities}
\providecommand{\cmark}{\ding{51}}
\providecommand{\xmark}{\ding{55}}
\begin{tabular}{lllccccl}
\toprule
\multirow{2}{*}{Benchmark} & \multirow{2}{*}{Language} & \multirow{2}{*}{Verifier} & \multicolumn{4}{c}{Given to model} & \multirow{2}{*}{Task} \\
\cmidrule(lr){4-7}
& & & NL desc. & Signature & Spec & Code & \\
\midrule
DafnyBench~\cite{dafnybench}    & Dafny       & Dafny     & \xmark & \cmark & \cmark & \cmark & annotations \\
miniCodeProps~\cite{minicodeprops} & Lean 4   & Lean      & \xmark & \cmark & \cmark & \cmark & proof \\
Verus-Bench~\cite{autoverus}    & Rust        & Verus     & \xmark & \cmark & \cmark & \cmark & annotations \\
CoqStoq~\cite{rango}            & \del{Coq }\rev{\coq}  & \del{Coq }\rev{\coq} & \xmark & \cmark & \cmark & \cmark & proof \\
\rev{VerKer~\cite{tu2026agentic}} & \rev{\coq} & \rev{\coq} & \rev{\xmark} & \rev{\cmark} & \rev{\cmark} & \rev{\cmark} & \rev{proof} \\
\rev{VeriSoftBench~\cite{xin2026verisoftbench}} & \rev{Lean 4} & \rev{Lean} & \rev{\xmark} & \rev{\cmark} & \rev{\cmark} & \rev{\cmark} & \rev{proof} \\
\rev{VeruSAGE-Bench~\cite{yang2025verusage}} & \rev{Rust} & \rev{Verus} & \rev{\xmark} & \rev{\cmark} & \rev{\cmark} & \rev{\cmark} & \rev{proof} \\
CloverBench~\cite{clover}       & Dafny       & Dafny     & \cmark & \cmark & \cmark & \xmark & code + annot. \\
\rev{AlgoVeri~\cite{algoveri}} & \rev{Dafny} & \rev{Dafny} & \rev{\cmark} & \rev{\cmark} & \rev{\cmark} & \rev{\xmark} & \rev{\textbf{code + proof}} \\
{VERINA}~\cite{verina}  & Lean 4 & Lean & \cmark & \cmark & \cmark & \xmark & \textbf{code + proof} \\
\bottomrule
\end{tabular}
\end{table}

\paragraph{Result.}
\sys{} reaches $\mathbf{176/189}$ on VERINA, well above prior SOTA $38$ and agentic baselines Codex $136$, Claude Code $149$; $\mathbf{65/77}$ on AlgoVeri's Dafny track, $2\times$ the published \rev{Gemini 3 Flash}\del{Gemini-3 Flash} baseline of $31/77$; and all three LLM methods saturate at $62/62$ on CloverBench. \rev{Across DafnyBench, miniCodeProps, Verus-Bench, CoqStoq, VERINA, AlgoVeri, and CloverBench, $41$ of the $778$ tasks hit the per-task cap: $12$ on DafnyBench, $13$ on VERINA, $12$ on AlgoVeri, $3$ on CoqStoq, and $1$~on~Verus-Bench.}

\rev{\subsection{Distributed system: IronFleet lock}
\label{app:ironfleet}
IronFleet~\cite{ironfleet} is a suite of verified distributed systems in Dafny, built over roughly $3.7$ person-years. We take its lock and remove all implementations and proofs (lemmas, asserts, loop invariants), keeping the spec unmodified. \sys{} regenerates and proves all $15$ files with no axioms or assumptions.}
\section{\sys{} Architecture Details}
\label{app:architecture}

This appendix expands on the components introduced in \cref{sec:design} and ablated in \cref{sec:eval-ablations}. The architecture instantiates the two-level CEGIS-style refinement of \cref{sec:design}. The inner loop is the DSA (\cref{app:worker}): forward moves \emph{Define}, \emph{Prove}, \emph{Decompose} extend the partial (code, proof) candidate, and the structured \coq{} diagnostic from a failed step (\cref{app:feedback}) is a concrete witness that the candidate cannot close as written, used by the agent to choose \emph{Repair} or \emph{Revert}. The outer loop is the ISA: a DSA's stalled proof attempts and bench failures (\cref{app:bench-signal}) drive the proposer (\cref{app:guide}) on tactical stalls and the reloader (\cref{app:resetter}) on strategic dead-ends. The audit step (\cref{app:auditor}) is orthogonal to refinement: it is a deterministic safety gate, not a counterexample signal.

\providecolor{specbg}{HTML}{F7F8FA}
\providecolor{specrule}{HTML}{C8CDD3}
\providecommand{\promptbox}[1]{%
  {\setlength{\fboxsep}{6pt}\setlength{\fboxrule}{0.3pt}%
   \noindent\fcolorbox{specrule}{specbg}{\begin{minipage}{0.96\linewidth}\small #1\end{minipage}}}%
}

\subsection{Proof methodology}
\label{app:proof-method}

\looseness=1 Each spec is a \coq{} \texttt{Module Type} carrying a \texttt{State} type, an \texttt{Update} type, seven method signatures (\texttt{getReq}, \texttt{getGuard}, \texttt{get}, \texttt{getRes}, \texttt{putReq}, \texttt{putGuard}, \texttt{put}), and an operational semantics on client--replica configurations. The implementation \sys{} produces is an \texttt{AlgDef} module providing the same seven methods at concrete state types, plus a \texttt{Refinement} module proving forward simulation: every step the implementation can take corresponds to a step the abstract spec can take, on related states. From the \texttt{Refinement} we obtain a \texttt{TraceInclusion} theorem: every observable trace of the implementation is also a trace of the abstract spec.

The tactic vocabulary is standard \coq{}: \texttt{inversion}, \texttt{destruct}, \texttt{subst}, \texttt{rewrite}, \texttt{apply}, \texttt{econstructor; eauto}, \texttt{simpl}, \texttt{lia}, \texttt{congruence}, with case-analysis on \texttt{Nat.eq\_dec} for equality decisions on identifiers and timestamps. No \texttt{firstorder}/\texttt{intuition} shortcuts; no \texttt{Admitted}, no \texttt{Axiom}, no \texttt{Hypothesis}, no \texttt{Parameter}. Closure is verified by \coq{}'s kernel.

\subsection{DSA: deductive synthesis}
\label{app:worker}

The DSA is a Codex (or Claude Code) session per workspace performing the deductive-synthesis loop of \cref{sec:design-worker}. At each search-tree node the agent moves forward with one of three primitives:
\begin{itemize}[leftmargin=*,itemsep=2pt,topsep=2pt,parsep=0pt]
\item \emph{Define}: declare or extend a state/message type or function body, with \texttt{admit} placeholders for parts not yet filled in (e.g.\ adding a \texttt{Cell} record to the per-key store with \texttt{cell\_deps} stubbed).
\item \emph{Prove}: close an open proof obligation by replacing \texttt{Admitted} with a tactic sequence ending in \texttt{Qed}, or extend an existing tactic block.
\item \emph{Decompose}: state a helper lemma as \texttt{Admitted} and use it to advance the proof of an existing lemma; the helper is discharged in a later iteration.
\end{itemize}

After each forward step \texttt{make} runs and \coq{}'s type-checker is the only oracle. \rev{The definitions and lemmas the proof depends on stay compiled in separate files, so \texttt{make} recompiles only the working file the step edits: about $4$ seconds on Chapar, versus $12$--$17$ seconds for a full build.} On error, the agent backtracks with one of two primitives:
\begin{itemize}[leftmargin=*,itemsep=2pt,topsep=2pt,parsep=0pt]
\item \emph{Repair}: re-attempt the failed step with a different tactic (e.g.\ \texttt{induction} instead of \texttt{destruct}; \texttt{lia} instead of \texttt{omega}).
\item \emph{Revert}: rewind to an earlier search-tree node and choose a different forward primitive (typically when the same compile error recurs three or more times).
\end{itemize}

The agent's initial brief is read from \texttt{CLAUDE.md} in the workspace and instantiates the \del{four briefings of }\rev{DSA of} \cref{sec:design-worker}\rev{ in four briefings} (methodology, auditability, design, tactical playbook). The brief shown below is for the \coq{} setting (our \sys{} suite specs and Chapar CC, plus CoqStoq\del{ and miniCodeProps}\rev{ and VerKer}); for cross-language benchmarks we swap the verifier and surface tools to match the target language: Dafny (DafnyBench, AlgoVeri-Dafny, CloverBench\rev{, IronFleet lock}), Verus (Verus-Bench\rev{, VeruSAGE-Bench}), and Lean~4 (VERINA\rev{, miniCodeProps, VeriSoftBench}), with success criterion adapted accordingly (e.g.\ \texttt{dafny verify} accepts, \texttt{lake build} succeeds, \texttt{verus} reports zero errors).

\promptbox{%
\textbf{Role: Deductive Synthesis Agent.} Given a specification, incrementally produce an implementation and a machine-checked proof that it satisfies the specification. The type-checker is the only oracle and accepts a partial state (with deferred holes) as a valid intermediate.

\textbf{Schema.} Synthesis is a finite sequence of well-typed steps; the file must type-check (possibly with deferred holes) after every step. Take one forward move:
\begin{itemize}[leftmargin=1.4em,itemsep=1pt,topsep=1pt]
\item \emph{Define} (refinement): introduce or extend a \texttt{Definition}, \texttt{Record}, \texttt{Inductive}, \texttt{Fixpoint}, or \texttt{Module}; bodies may contain \texttt{admit} or invoke \texttt{Admitted} lemmas.
\item \emph{Prove} (closure): replace an \texttt{Admitted} lemma with a tactic sequence ending in \texttt{Qed}, or extend an existing partial proof.
\item \emph{Decompose} (lemma introduction): state a helper lemma as \texttt{Admitted}, use it to advance the current proof, and discharge it later. This is the deferred-hole device of deductive synthesis.
\end{itemize}
On error, backtrack: \emph{Repair} (different tactic on the same goal, e.g.\ \texttt{induction} instead of \texttt{destruct}, \texttt{lia} instead of \texttt{omega}) or \emph{Revert} (rewind to an earlier state and try a different forward move; trigger when the same error recurs three or more times).

\textbf{Invariants.} (I1)~The file type-checks after every step. (I2)~Every \texttt{Admitted} is a deferred hole, not a permanent assumption: all \texttt{Admitted} must be closed before reporting \emph{done}. (I3)~The specification is fixed. (I4)~No \texttt{Axiom}, \texttt{Hypothesis}, \texttt{Parameter}, or unguarded recursion.

\textbf{Design.} State types should be concrete (records, bounded lists, indexed nats), not function types whose \texttt{override} builds closure chains under repeated update. State must not grow unboundedly with the number of operations: traces, logs, and accumulated dependency lists belong in the proof context, not the runtime state.

\textbf{Termination.} Report \emph{done} when (a)~the file type-checks with zero \texttt{Admitted} and zero \texttt{admit}, and (b)~the audit step passes (no vacuous proofs; no specification edits; non-trivial method bodies). The coordinator additionally extracts the implementation and runs the benchmark harness; if extraction or execution fails, the run reverts.

\textbf{Tactical playbook.}
\begin{itemize}[leftmargin=1.4em,itemsep=1pt,topsep=1pt]
\item Build simple to complex: prove small structural facts first, then combine.
\item When stuck on a subgoal, read it carefully; the goal often names the missing helper.
\item Use \texttt{admit} (lowercase) to skip subgoals temporarily; the final file must have zero \texttt{Admitted} and~zero~\texttt{admit}.
\end{itemize}
}

\rev{The $-$G ablation (\cref{sec:eval-ablations}) removes the tactical playbook, that is, the tactic and decomposition guidance, keeping the task constraints and the \coq{} feedback loop. The playbook is generic proof-engineering advice, not domain-specific algorithms, state representations, or proof content; with only tactics adapted per language, it applies across Dafny, Lean, Verus, and \coq{}.}

\cref{app:progression} walks through an end-to-end illustration of this schema on a small target (\texttt{all\_less\_than}): each of the five synthesis steps is annotated with its move (\emph{Define}, \emph{Prove}, \emph{Decompose}), with every intermediate listing type-checking under the kernel.

\subsection{Audit step}
\label{app:auditor}

The audit step is a deterministic Python script that gates closure with kernel-level verdicts (no model outputs). It encapsulates the cheating patterns we have observed during runs.

\textbf{Static checks.}
\begin{itemize}[leftmargin=*,itemsep=2pt,topsep=2pt]
\item \texttt{Admitted} count is zero.
\item After stripping comments, \texttt{Axiom}, \texttt{Hypothesis}, and \texttt{Parameter} counts are zero.
\item \texttt{make clean \&\& make} exits zero.
\item \rev{The specification is unchanged.}
\end{itemize}

\textbf{Vacuous-proof patterns.} The script also matches the closed file against \del{four }\rev{five} patterns:
\begin{itemize}[leftmargin=*,itemsep=2pt,topsep=2pt]
\item A \texttt{put} that returns a fixed constant the spec accepts on every input.
\item An always-true \texttt{getGuard} that admits any read.
\item \rev{An always-false \texttt{putGuard} under which the theorem holds trivially.}
\item A forward-simulation theorem stated over an empty domain so that it holds vacuously.
\item A state-only stub whose \texttt{Update} drops the payload and returns the initial state on every operation.
\end{itemize}

A candidate that fails any check is sent to the ISA; only candidates passing all checks count as audit-clean closures.

\subsection{Proposer (ISA tactical role)}
\label{app:guide}

After the coordinator counts no progress across $10$ consecutive polls, the proposer fires as a one-shot LLM session. It receives the work file, every \texttt{Admitted} lemma's surrounding $17$-line context, and the most recent \texttt{make} error with $4$ lines of trailing context when present, and writes its output to \texttt{GUIDANCE.md} in the workspace.

\promptbox{%
\textbf{Role.} You are an expert \coq{} proof advisor. Read the work file and focus on the proof goals and error context below. Do NOT edit the \texttt{.v} file.

\textbf{For each Admitted proof, follow this checklist:}
\begin{enumerate}[leftmargin=1.4em,itemsep=1pt,topsep=1pt]
\item \textbf{Read the lemma statement.} What does it claim? What are the hypotheses? What is the conclusion~type?
\item \textbf{Identify the proof structure needed:}
\texttt{forall x, P x} $\to$ \texttt{intros x};
inductive type (list, trace, nat) $\to$ \texttt{induction} on it;
step/transition $\to$ \texttt{destruct} on the step type, then case analysis;
simulation $\to$ invariant + induction on trace length;
two functions equal $\to$ \texttt{functional\_extensionality}.
\item \textbf{Identify missing helper lemmas.} For each subgoal you cannot close directly, state the EXACT helper (name, signature, informal meaning). Suggest: prove it separately; \texttt{Admit} it in the main proof; close it.
\item \textbf{If the proof is $>30$ lines, decompose:} by cases (\texttt{X\_case\_C1}, \texttt{X\_case\_C2}, \dots\ for each constructor); by subgoal (if $A \wedge B$, prove $A$ and $B$ separately); extract common patterns repeating $3$+ times.
\item \textbf{Suggest specific tactic sequences,} not vague directions.
\end{enumerate}

\textbf{For compile errors,} diagnose using a decision tree (e.g., \emph{``Cannot unify X with Y''} $\to$ type mismatch, use \texttt{Check expr.}; \emph{``Tactic failure''} $\to$ try \texttt{lia} instead of \texttt{omega}; \emph{``No matching clauses''} $\to$ incomplete pattern~match).

\textbf{Output.} Write \texttt{GUIDANCE.md} with one section per Admitted proof or error: (a) the diagnosis, (b) the exact tactic sequence to try, (c) any helper-lemma signatures.
}

On Chapar, the proposer chose three sub-lemma classes (per-message, per-store-entry, per-replica-clock) by \rev{recognizing}\del{recognising} that the global delivery invariant decomposes cleanly along these axes, none of which \coq{}'s automation closes alone.

\subsection{Reloader (ISA strategic role)}
\label{app:resetter}

The reloader escalates every $20$ polling cycles when no progress is detected: Level $1$ at stall cycle $20$, Level $2$ at $40$, Level $3$ at $60$, and so on. Each level is a one-shot LLM session \rev{that reads}\del{reading} \texttt{DESIGN\_LOG.md}, \texttt{STATUS.md}, and every prior level's output, and writes \texttt{META\_GUIDANCE\_L\{N\}.md}. The \del{strategist }\rev{coordinator} consumes the output to rewrite its plan and spawn a fresh worker on the new design.

\promptbox{%
\textbf{Role.} You are a level-$N$ formal verification architect. Read \texttt{DESIGN\_LOG.md}, \texttt{STATUS.md}, and every prior \texttt{META\_GUIDANCE\_L\{1..N{-}1\}.md}: the history of all approaches tried and why each failed.

\textbf{Synthesize a NEW approach that avoids ALL recorded failures.} Think across these dimensions:
\begin{enumerate}[leftmargin=1.4em,itemsep=1pt,topsep=1pt]
\item \textbf{Data representation.} Concrete types for state and messages: records, bounded lists, nats, not functions.
\item \textbf{Abstraction function.} How concrete state maps to abstract state. This is the KEY creative choice; if previous attempts failed on the simulation proof, the abstraction was wrong.
\item \textbf{Invariant.} What property holds at every reachable state. The invariant must be (a)~true initially, (b)~preserved by every step, (c)~strong enough to prove the final theorem.
\item \textbf{Proof decomposition.} Which helper lemmas are needed, stated explicitly with signatures.
\item \textbf{Minimum information.} For each function that takes a collection, could the caller pass a smaller collection and still get a correct result? The most precise design passes only what is needed per call.
\end{enumerate}

\textbf{Output.} Write \texttt{META\_GUIDANCE\_L\{N\}.md} as a CONCRETE blueprint a worker can directly implement: exact \texttt{Record} types, method signatures, abstraction function, invariant, and lemma decomposition. Do NOT write vague advice; do NOT spawn workers.
}

On Chapar, the reloader chose a per-key store entry recording sender, sender clock, and dependency vector; alternative layouts (e.g.\ a single global per-replica state) leave the delivery lemma unprovable from local information.

\subsection{\coq{} feedback}
\label{app:feedback}

Each \texttt{make} invocation captures stdout/stderr; on non-zero exit, the coordinator extracts every \texttt{Error:} line plus $2$ lines of preceding context and $4$ lines of trailing context as a single error block. The block is included verbatim in the worker's next prompt alongside the proof goal, the local hypothesis context, and the tactic backtrace where applicable. No \rev{normalization}\del{normalisation} or \rev{summarization}\del{summarisation} is applied: the structured \coq{} diagnostic carries the \rev{localization}\del{localisation} signal. Replacing this stream with a binary accept/reject collapses the search (\cref{sec:eval-ablations} reports $\le 1/3$ closure on every \coq{} spec under $-$VF).

\subsection{Performance feedback}
\label{app:bench-signal}

When a worker reaches a verifier-clean state with a design hash distinct from prior runs, the coordinator invokes the benchmark harness asynchronously. The harness extracts the implementation to OCaml, deploys it on the cluster (\cref{app:perf-harness}), and writes results to \texttt{PERF\_RESULTS.md} in the workspace. Throughput targets are calibrated per-spec from the published reference; implementations whose throughput falls below the target are flagged and trigger the reloader on the next escalation.
\section{Synthesis Progression: \texttt{all\_less\_than}}
\label{app:progression}

This appendix walks through one full progression of \sys{}' deductive-synthesis loop on a small target. Step~0 and the Final State are complete \coq{} files that compile; the listings for intermediate Steps~1--5 show only the parts that change from the previous step, with not-yet-filled bodies stubbed as \texttt{Parameter}. \rev{These stubs are only for exposition: in \sys{} runs, Invariant I4 forbids \texttt{Parameter}, and the DSA leaves unfilled bodies as \texttt{admit} or \texttt{Admitted} (\cref{app:worker}).} The progression on real distributed-system specs follows the same pattern at much larger scale (\cref{sec:eval-synthesis}); the small target keeps each listing short enough to read end-to-end.

The five synthesis steps below are an instance of the deductive-synthesis schema of \cref{app:worker}. Step 1 is a \emph{Decompose} (split the goal into a \texttt{nil} and a \texttt{cons} helper lemma). Step 2 combines \emph{Define} (fill the \texttt{nil} body) with \emph{Prove} (close the \texttt{nil} helper). Step 3 is another \emph{Decompose} (split the \texttt{cons} helper on the head). Step 4 \rev{combines \emph{Define} (the recursive call) with \emph{Prove} (close the \texttt{then} helper)}\del{is \emph{Define} (the recursive call)}. Step~5 combines \emph{Define} (fill the \texttt{else} branch) with \emph{Prove} (close the \texttt{else} helper). The file type-checks at every step (Invariant I1); \texttt{Admitted} placeholders carry the deferred holes forward and are all discharged by the Final state.

The target is \texttt{all\_less\_than :\ list nat -> nat -> bool}, returning \texttt{true} iff every element of the input list is strictly less than the bound \texttt{n}. The correctness statement is the iff-equivalence~with~\texttt{Forall}:
\begin{quote}
  \texttt{all\_less\_than l n = true} $\Leftrightarrow$ \texttt{Forall (fun x => x < n) l}.
\end{quote}

\subsection{Step 0: bare specification}
The implementation is a \texttt{Parameter}; the correctness lemma is \texttt{Admitted}. The file type-checks but proves nothing.

\begin{lstlisting}[label={lst:state0}, caption={Step 0: spec only.}]
Require Import Coq.Lists.List.

Parameter all_less_than : list nat -> nat -> bool.

Lemma all_less_than_correct : forall (l : list nat) (n : nat),
  all_less_than l n = true <-> Forall (fun x => x < n) l.
Proof.
Admitted.
\end{lstlisting}

\subsection{Step 1: case-split on the input list}
The agent commits to recursing on \texttt{l}. The implementation becomes a \texttt{Definition} with a \texttt{match} on \texttt{nil}/\texttt{cons}; the branch bodies are stubbed as \texttt{Parameter}s. The correctness lemma is split into a \texttt{\_nil} helper and a \texttt{\_cons} helper, both \texttt{Admitted}; the main lemma now closes by \texttt{induction} on \texttt{l}, applying each helper at its leaf.

\begin{lstlisting}
Parameter nil_body : bool.
Parameter cons_body : nat -> list nat -> bool.

Definition all_less_than (l : list nat) (n : nat) : bool :=
  match l with
  | nil => nil_body
  | x :: xs => cons_body x xs
  end.

Lemma all_less_than_correct_nil : forall (n : nat),
  all_less_than nil n = true <-> Forall (fun x => x < n) nil.
Proof. Admitted.

Lemma all_less_than_correct_cons :
  forall (x : nat) (xs : list nat) (n : nat),
  all_less_than (x :: xs) n = true <-> Forall (fun x => x < n) (x :: xs).
Proof. Admitted.

Lemma all_less_than_correct : forall (l : list nat) (n : nat),
  all_less_than l n = true <-> Forall (fun x => x < n) l.
Proof.
  intros l n. induction l as [| x xs IH].
  - apply all_less_than_correct_nil.
  - apply all_less_than_correct_cons.
Qed.
\end{lstlisting}

\subsection{Step 2: fill the \texttt{nil} branch and close the \texttt{nil} helper}
The agent fills \texttt{nil\_body} with \texttt{true} (replacing the \texttt{Parameter}) and proves \texttt{\_nil}: both sides of the iff are vacuously satisfied (\texttt{true = true}; \texttt{Forall \_ nil} holds by the empty constructor). The \texttt{cons} body is still a \texttt{Parameter}.

\begin{lstlisting}
Definition all_less_than (l : list nat) (n : nat) : bool :=
  match l with
  | nil => true
  | x :: xs => cons_body x xs
  end.

Lemma all_less_than_correct_nil : forall (n : nat),
  all_less_than nil n = true <-> Forall (fun x => x < n) nil.
Proof. intros. simpl. split; intros; constructor. Qed.
\end{lstlisting}

\subsection{Step 3: introduce the \texttt{if} branch and split the \texttt{cons} helper}
The agent commits to comparing \texttt{x} against \texttt{n} via \texttt{x <?\ n}, leaving both branches as \texttt{Parameter} stubs. The \texttt{\_cons} helper is split into a \texttt{\_then} sub-helper (under \texttt{x < n}) and an \texttt{\_else} sub-helper (under \texttt{$\neg$(x < n)}), and \texttt{\_cons} now proves itself by case-splitting on \texttt{Nat.ltb\_spec} and applying the appropriate sub-helper.\rev{ The main lemma now passes the induction hypothesis to \texttt{\_cons}.}

\begin{lstlisting}
Parameter then_body else_body : nat -> list nat -> nat -> bool.

Definition all_less_than (l : list nat) (n : nat) : bool :=
  match l with
  | nil => true
  | x :: xs => if x <? n then then_body x xs n else else_body x xs n
  end.

Lemma all_less_than_correct_cons_then :
  forall (x : nat) (xs : list nat) (n : nat),
  x < n -> (all_less_than xs n = true <-> Forall (fun x => x < n) xs) ->
  (all_less_than (x :: xs) n = true <-> Forall (fun x => x < n) (x :: xs)).
Proof. Admitted.

Lemma all_less_than_correct_cons_else :
  forall (x : nat) (xs : list nat) (n : nat),
  ~ (x < n) -> (all_less_than xs n = true <-> Forall (fun x => x < n) xs) ->
  (all_less_than (x :: xs) n = true <-> Forall (fun x => x < n) (x :: xs)).
Proof. Admitted.

Lemma all_less_than_correct_cons :
  forall (x : nat) (xs : list nat) (n : nat),
  all_less_than xs n = true <-> Forall (fun x => x < n) xs ->
  all_less_than (x :: xs) n = true <-> Forall (fun x => x < n) (x :: xs).
Proof.
  intros. destruct (Nat.ltb_spec x n) as [Hlt | Hge].
  - apply all_less_than_correct_cons_then; assumption.
  - apply all_less_than_correct_cons_else; [apply Nat.le_ngt; assumption | assumption].
Qed.

Lemma all_less_than_correct : forall (l : list nat) (n : nat),
  all_less_than l n = true <-> Forall (fun x => x < n) l.
Proof.
  intros l n. induction l as [| x xs IH].
  - apply all_less_than_correct_nil.
  - apply all_less_than_correct_cons. assumption.
Qed.
\end{lstlisting}

\subsection{Step 4: fill the \texttt{then} branch with the recursive call}
The agent fills \texttt{then\_body} with \texttt{all\_less\_than xs n}: the implementation now actually recurses, so \texttt{Definition} is promoted to \texttt{Fixpoint}. The \texttt{\_then} helper proves cleanly: under \texttt{x < n}, applying \texttt{Nat.ltb\_lt} rewrites the \texttt{if}-condition to \texttt{true}, and \texttt{Forall} on \texttt{x ::\ xs} reduces to \texttt{x < n} on the head plus \texttt{Forall} on \texttt{xs}, both available.

\begin{lstlisting}
Fixpoint all_less_than (l : list nat) (n : nat) : bool :=
  match l with
  | nil => true
  | x :: xs => if x <? n then all_less_than xs n else else_body x xs n
  end.

Lemma all_less_than_correct_cons_then :
  forall (x : nat) (xs : list nat) (n : nat),
  x < n -> (all_less_than xs n = true <-> Forall (fun x => x < n) xs) ->
  (all_less_than (x :: xs) n = true <-> Forall (fun x => x < n) (x :: xs)).
Proof.
  intros x xs n Hlt H_ind. simpl.
  apply Nat.ltb_lt in Hlt. rewrite Hlt.
  split; intros.
  - constructor; [apply Nat.ltb_lt; assumption | tauto].
  - inversion H. tauto.
Qed.
\end{lstlisting}

\subsection{Step 5: fill the \texttt{else} branch and close the \texttt{else} helper}
The agent's first attempt is to recurse: \texttt{else\_body := all\_less\_than xs n} (``skip the out-of-bound element, keep checking''). The implementation type-checks, but the \texttt{\_else} obligation does not: under $\neg(x < n)$ the goal becomes \texttt{all\_less\_than xs n = true} $\Leftrightarrow$ \texttt{Forall (fun y => y < n) (x ::\ xs)}, which fails on \texttt{xs = nil}, \texttt{x = 5}, \texttt{n = 3} (LHS is \texttt{true}; RHS demands \texttt{5 < 3}). \coq{} rejects the proof. The agent backtracks and replaces the \texttt{else} body with \texttt{false}: the function now returns false on any out-of-bound element, the \texttt{\_else} helper closes (LHS becomes \texttt{false = true}, closed by \texttt{discriminate}; RHS gives \texttt{x < n} from inversion of \texttt{Forall}, contradicting $\neg(x < n)$).

\begin{lstlisting}
(* Attempt 1: else_body := all_less_than xs n  (rejected: iff fails) *)
(* Attempt 2: else_body := false               (accepted)            *)

Lemma all_less_than_correct_cons_else :
  forall (x : nat) (xs : list nat) (n : nat),
  ~ (x < n) -> (all_less_than xs n = true <-> Forall (fun x => x < n) xs) ->
  (all_less_than (x :: xs) n = true <-> Forall (fun x => x < n) (x :: xs)).
Proof.
  intros x xs n Hge _. simpl.
  apply Nat.le_ngt in Hge. apply Nat.ltb_ge in Hge. rewrite Hge.
  split.
  - discriminate.
  - intros HF. inversion HF; subst.
    apply Nat.ltb_lt in H1. congruence.
Qed.
\end{lstlisting}

\subsection{Final state}

After five steps every \texttt{Parameter} stub and \texttt{Admitted} placeholder is closed. \coq{}'s kernel accepts the file, verifying \texttt{all\_less\_than\_correct} on every input.

\begin{lstlisting}[label={lst:state5}, caption={Step 5 (final): implementation and proof, fully verified.}]
Require Import Coq.Lists.List.
Require Import Coq.Arith.PeanoNat.

Fixpoint all_less_than (l : list nat) (n : nat) : bool :=
  match l with
  | nil => true
  | x :: xs => if x <? n then all_less_than xs n else false
  end.

Lemma all_less_than_correct_nil : forall (n : nat),
  all_less_than nil n = true <-> Forall (fun x => x < n) nil.
Proof. intros. simpl. split; intros; constructor. Qed.

Lemma all_less_than_correct_cons_then :
  forall (x : nat) (xs : list nat) (n : nat),
  x < n -> (all_less_than xs n = true <-> Forall (fun x => x < n) xs) ->
  (all_less_than (x :: xs) n = true <-> Forall (fun x => x < n) (x :: xs)).
Proof.
  intros x xs n Hlt H_ind. simpl.
  apply Nat.ltb_lt in Hlt. rewrite Hlt.
  split; intros.
  - constructor; [apply Nat.ltb_lt; assumption | tauto].
  - inversion H. tauto.
Qed.

Lemma all_less_than_correct_cons_else :
  forall (x : nat) (xs : list nat) (n : nat),
  ~ (x < n) -> (all_less_than xs n = true <-> Forall (fun x => x < n) xs) ->
  (all_less_than (x :: xs) n = true <-> Forall (fun x => x < n) (x :: xs)).
Proof.
  intros x xs n Hge _. simpl.
  apply Nat.le_ngt in Hge. apply Nat.ltb_ge in Hge. rewrite Hge.
  split.
  - discriminate.
  - intros HF. inversion HF; subst.
    apply Nat.ltb_lt in H1. congruence.
Qed.

Lemma all_less_than_correct_cons :
  forall (x : nat) (xs : list nat) (n : nat),
  all_less_than xs n = true <-> Forall (fun x => x < n) xs ->
  all_less_than (x :: xs) n = true <-> Forall (fun x => x < n) (x :: xs).
Proof.
  intros. destruct (Nat.ltb_spec x n) as [Hlt | Hge].
  - apply all_less_than_correct_cons_then; assumption.
  - apply all_less_than_correct_cons_else; [apply Nat.le_ngt; assumption | assumption].
Qed.

Lemma all_less_than_correct : forall (l : list nat) (n : nat),
  all_less_than l n = true <-> Forall (fun x => x < n) l.
Proof.
  intros l n. induction l as [| x xs IH].
  - apply all_less_than_correct_nil.
  - apply all_less_than_correct_cons. assumption.
Qed.
\end{lstlisting}

The progression on real distributed-system specs follows the same pattern at much larger scale: dozens of helper lemmas, hundreds of lines of implementation, and tactics that range over inductive simulation arguments rather than single-step rewrites. \cref{sec:design} describes how \sys{} drives this process; the case-study trajectory in \cref{sec:eval} shows it on the causal-consistency specification.
\section{Limitations, Broader Impacts, and Ethics}
\label{app:limitations-impacts}

The main paper presents \sys{}' design and evaluation; this appendix addresses the limitations of the work, the broader impacts of verified-synthesis tooling, and ethical considerations.

\paragraph{Limitations.}
\sys{} requires a formal \del{\coq{} \texttt{Module Type} }\rev{specification} as input; we do not synthesize specifications from natural language\rev{; automating specification development is an orthogonal open problem (\cref{sec:discussion})}. Compute cost is $2$--$11$ hours and \$52--\$155 per closed spec, substantially below the months-to-years of expert proof effort it replaces but still meaningful. We evaluate \rev{system-level synthesis }on distributed key-value-store consistency\rev{ and the IronFleet lock, and proof synthesis alone also on the Linux kernel (VerKer) and on Rust systems such as OS kernels, memory allocators, and serialization code (VeruSAGE-Bench)}; generality \rev{of the former }to other verified-synthesis domains (compilers, OS kernels, cryptographic protocols) is conjectured, not measured. \rev{IronFleet's larger IronRSL and IronKV are future work.} \rev{For the key-value stores, \sys{} verifies at the protocol level: \coq{} checks the implementation and its refinement proof. As in Verdi~\cite{verdi} and Chapar~\cite{chapar}, the trusted computing base is the \coq{} kernel, the extraction, the OCaml toolchain, the distributed runtime (\cref{sec:eval-setup,app:perf-harness}), and the hardware. Our synthesis method is independent of this boundary; extending down the stack, via CertiCoq~\cite{anand2017certicoq} for verified compilation, together with validation of the runtime, is a promising direction.}

\paragraph{Broader impacts.}
Distributed systems are common in production software (databases, message queues, cloud services), but formal verification of these systems has remained out of reach for most production code because hand-written proofs take person-years of expert effort. \sys{} reduces that effort to hours of compute, putting verified distributed software within reach of production teams. Kernel-checked proofs make \sys{}' outputs safer than unverified LLM-generated code: the verifier rules out silent bugs against the specification. As with all formal verification, mis-specification remains a residual risk.

\paragraph{Ethics.}
The work involves no human subjects and no scraped data; the verified outputs introduce no dual-use risks beyond those already present in unverified LLM code generation.
\section{\sys{} Suite}
\label{app:suite}

\sys{} suite is the benchmark dataset of six distributed key-value-store specifications we release alongside the system; together with Chapar's published causal-consistency spec~\cite{chapar}, they form the seven specifications evaluated in \cref{sec:eval}. Each specification fixes (i) the abstract spec state and operations the application sees, (ii) the concrete replica state and operations the implementation runs, and (iii) the consistency property the implementation must satisfy. This appendix gives the formal definition of every spec, plus the reference implementations we benchmark in \cref{sec:eval}. The framework setup below introduces the notation: client programs (\cref{fig:prog-def}), the seven-method \texttt{AlgDef} interface (\cref{fig:alg-def}), the configuration syntax (\cref{fig:configuration-syntax}), the concurrent operational semantics (\cref{fig:updated-conc-op-sem}), the refinement hierarchy (\cref{fig:refinement-hierarchy}), and a relaxed baseline spec (\cref{fig:relaxed-specification}).

\paragraph{Summary.}
\Cref{tab:suite-summary} lists the six \sys{} suite specifications plus Chapar's published CC. Each is then introduced in plain English and immediately followed by its formal definition and \mbox{reference}~\mbox{implementation}.

\begin{table}[h]
\centering
\caption{The six \sys{} suite specifications, plus Chapar's published CC for reference.}
\label{tab:suite-summary}
\small
\setlength{\tabcolsep}{6pt}
\renewcommand{\arraystretch}{1.15}
\begin{tabular}{@{}llp{0.45\linewidth}@{}}
\toprule
spec & abbreviation & informal guarantee \\
\midrule
Read-Your-Writes                          & RYW    & A client's read sees that client's own most recent write. \\
Monotonic Reads                           & MR     & A client's reads never go back in time relative to writes the client has already seen. \\
Monotonic Writes                          & MW     & Writes from one client are applied in order at every replica. \\
Read-Your-Writes $+$ Monotonic Writes     & RYW+MW & Composition of RYW and MW; an implementation must refine both. \\
Causal Consistency                        & CC     & A read returns a value whose causal predecessors have all been observed at the responding replica. \\
Labeled Causal Consistency                & LCC    & CC \rev{parameterized}\del{parameterised} by per-message topic labels: each client subscribes to a subset of labels and only sees causality enforced on those. \\
Chapar Causal Consistency                 & Chapar CC & Chapar's published CC specification~\cite{chapar}; see the original paper for its formal definition. \\
\bottomrule
\end{tabular}
\end{table}

\providecolor{specbg}{HTML}{F7F8FA}
\providecolor{specrule}{HTML}{C8CDD3}
\newcommand{\specfig}[1]{%
  {\setlength{\fboxsep}{4pt}\setlength{\fboxrule}{0.3pt}%
   \fcolorbox{specrule}{specbg}{#1}}%
}

\begin{figure}[h]
\centering
\specfig{%
\begin{minipage}{0.90\linewidth}
  \figuredefstextsize
  \setlength{\abovedisplayskip}{0pt}%
  \setlength{\abovedisplayshortskip}{0pt}%
  \setlength{\belowdisplayskip}{0pt}%
  \setlength{\belowdisplayshortskip}{0pt}%
   \begin{mathpar}
      \begin{array}{rcl@{\qquad}l}
      k
         & : &
            K  %
         & 
         \mbox{Key}
      \\
      v
         & : &
            V ⊇ K
         & 
         \mbox{Value}
      \\
      x
         & : &
            V
         & 
         \mbox{Variable}
      \\
      i
         & &
            
         & 
         \mbox{Unique Identifiers}
      \\
      s : S
         & ::= &
            \sPut[i]{k}{v} ; \ s
         &
         \mbox{Statement}
      \\ & | &
         \sGet[i]{x}{k}; \ s
         &
      \\ & | &
         \sSkip
         &
      \\ & | &
         &
         \mbox{Extended Syntax for Internal Statements}
      \\ & | &
         ⊘ \, \sGet[i]{x}{k}; \ s
         &
         \mbox{Blocked Get}
   \\
   c
      & : &
         C
      & 
      \mbox{Clients}
   \\
      a 
         & : &
            A = C ↦ S
      &
      \mbox{Application}            
      \end{array}
   \end{mathpar}
\end{minipage}%
}
   \caption{\label{fig:prog-def}Client Programs}
\end{figure}

\begin{figure}[h]
\centering
\specfig{%
\resizebox{0.92\linewidth}{!}{%
\begin{minipage}{1.05\linewidth}
  \figuredefstextsize
  \setlength{\abovedisplayskip}{0pt}%
  \setlength{\abovedisplayshortskip}{0pt}%
  \setlength{\belowdisplayskip}{0pt}%
  \setlength{\belowdisplayshortskip}{0pt}%
   \begin{mathpar}
      \begin{array}{lr}
         𝐈 = (\CState, \cInit, \RState, \rInit,
         & \note{Implementation}
      \\ \phantom{𝐈 = (}
         \getReq, \getGuard, \get, \getRes, 
      \\ \phantom{𝐈 = (}         
         \putReq, \putGuard, \lput)
      \\
         C
         & 
         \note{Clients}
      \\
         \CState : \Type
         & \note{Client State}
      \\
         \cInit : C → \CState
         & \note{Client Initial State}
      \\
         R
         &
         \note{Replicas}         
      \\
         \RState : \Type
         & \note{Replica State}
      \\
         \rInit : R → V → \RState
         & \note{Replica Initial State}
      \\
         \GetReqPayload, \GetResPayload: \Type
         & \note{Get Payload Types}
      \\
         \getReq : (K) (C, \CState) → 
         (\GetReqPayload, \CState)
         & \note{Get Request (at client)}
      \\
         \getGuard : (K) (\GetReqPayload) (C, R, \RState) → \Bool
         & \note{Get Guard (at replica)}
      \\
         \get : (K) (\GetReqPayload) (C, R, \RState) → 
         (V × \GetResPayload, \RState)
         & \note{Get (at replica)}
      \\
         \getRes : (K, V) (\GetResPayload) (C, \CState) → \CState
         & \note{Get Response (at client)}
      \\
         \PutReqPayload: \Type
         & \note{Put Payload Type}
      \\
         \putReq : (K, V) (C, \CState) → 
         (\PutReqPayload × \CState)
         & \note{Put Request (at client)}
      \\
         \putGuard : (K, V) (\PutReqPayload) (C, R, \RState) → \Bool
         & \note{Put Guard (at replica)}
      \\
         \lput : (K, V) (\PutReqPayload) (C, R, \RState) → \RState
         & \note{Put (at replica)}
      \end{array}
   \end{mathpar}
\end{minipage}%
}}
   \caption{\label{fig:alg-def}
      Key-Value Store Implementation Interface}
\end{figure}

\clearpage
\begin{figure}[t]
\centering
\specfig{%
\begin{minipage}{0.90\linewidth}
  \figureruletextsize
  \setlength{\abovedisplayskip}{0pt}%
  \setlength{\abovedisplayshortskip}{0pt}%
  \setlength{\belowdisplayskip}{0pt}%
  \setlength{\belowdisplayshortskip}{0pt}%
\figuredefstextsize
\centering
$
\begin{array}{rcl@{\qquad}l}
   𝐈 & = &
      (\CState, \cInit, \RState, \rInit, 
      & \note{Implementation}
   \\ 
      & & \phantom{(}
      \getReq, \getGuard, \get, \getRes, 
   \\   
      & & \phantom{(}
      \putReq, \putGuard, \lput)
   \\
   W  : 𝓦 
      & ≔ &
         〈𝓒, 𝓡, 𝓝 〉
      &
      \mbox{World}
   \\
   c
      & : &
         C
      & 
      \mbox{Clients}
   \\
   σ
      & : &
         \CState
      & 
      \mbox{Client State}
   \\
   𝓒
      & : &
         C ↦ 〈\CState, S〉
      &
      \mbox{Client Programs}
   \\
   r
      & : &
         R
      &
      \mbox{Replicas}
   \\
   ς
      & : &
         \RState
      & 
      \mbox{Replica State}         
   \\
   𝓡 
      & : &
         R ↦ \RState
      & 
      \mbox{Replica States}
   \\
   𝓝
      & : &
         \mathbb{PM}(M)
      &
      \mbox{Network}
   \\
   M
      & : &
      C × R × \GetReq(I, K, \GetReqPayload)
      & 
      \mbox{Message}
   \\
      & | &
      R × C × \GetRes(I, K, V, \GetResPayload)
   \\
      & | &
      C × R × \PutReq(K, V, \PutReqPayload)      
   \\
   l
      & ::= &
      c ▹ \sGet[i]{}{k}
      & 
      \mbox{Label}
   \\ & | &
      r ▹ \sGet[i]{}{k}: v
   \\ & | &
      c ▹ \sGet[i]{}{k}: v
   \\ & | &
      c ▹ \sPut{k}{v}
   \\ & | &
      r ▹ \sPut{k}{v}
      &
   \\
   h
      & ::= &
         l ⃰
      & 
      \mbox{History}
   \\
   \ext(h)
      & ::= &
         h \ | \ \{ c ▹ \sGet[i]{}{k}: v, \ c ▹ \sPut{k}{v} \}
      & 
      \mbox{External History}               
   \end{array}
$

\ \\
\ \\

$
   \begin{array}{rcl}
   ○ & : & \Unit
   \\
   \mathbb{PM}(S)
      & \defeq &
      \mbox{The multiset powerset of the set $S$}
   \\
      W₀(a)_{𝐈}
      & \defeq &
      〈[\overline{c ↦ 〈\cInit(c), a(c)〉}_{c ∈ C}],
         [\overline{r ↦ \rInit(r, v₀)}_{r ∈ R}],
         ∅〉
   \\
      v_0
      & \defeq &
      \mbox{The initial value}
   \end{array}
$
\end{minipage}%
}

\caption{The State of the Key-\rev{V}\del{v}alue Store Operational Semantics}
\label{fig:configuration-syntax}

\end{figure}

\clearpage
\begin{figure}
\centering
\specfig{%
\begin{minipage}{0.90\linewidth}
  \figureruletextsize
  \setlength{\abovedisplayskip}{0pt}%
  \setlength{\abovedisplayshortskip}{0pt}%
  \setlength{\belowdisplayskip}{0pt}%
  \setlength{\belowdisplayshortskip}{0pt}%
  \begin{mathpar}
      \indrule[Get-Req] {
         c ≠ c₀
         \\\\ 
         \getReq (k) (c, σ) ⇝ ⃰  〈p, σ'〉
      } {
         \world{
            𝓒 [c ↦ 〈σ, \sGet[i]{x}{k}; s〉]
         }{
            𝓡 
         }{
            𝓝
         }
         \xrightarrow{c \ ▹ \ \sGet[i]{}{k}}_{𝐈}
         \world{
            𝓒 [c ↦ 〈σ', ⊘ \, \sGet[i]{x}{k}; s〉]
         }{
            𝓡 
         }{
            𝓝  ∪ \{ \overline{〈c, r, \GetReq(i, k, p)〉 }_{r ∈ R}  \}
         }
      }
      
      \indrule[Get] {
         \getGuard (k) (p) (c, r, ς) ⇝ ⃰  \true
         \\\\
         \get (k) (p) (c, r, ς) ⇝ ⃰  〈v, p', ς'〉
      } {
         \world{
            𝓒 
         }{
            𝓡 [r ↦ ς]
         }{
            𝓝 ∪ \{〈 c, r, \GetReq(i, k, p) 〉 \}
         }
         \xrightarrow{r \ ▹ \ \sGet[i]{}{k} : v}_{𝐈}
         \world{
            𝓒
         }{
            𝓡 [r ↦ ς']
         }{
            𝓝  ∪ \{ 〈 r, c, \GetRes( i, k, v, p' ) 〉\}
         }
      }
   
      \indrule[Get-Res] {
         \getRes (k, v) (p) (c, σ) ⇝ ⃰  σ'
      } {
         \world{
            𝓒 [c ↦ 〈σ, ⊘ \, \sGet[i]{x}{k}; s〉]
         }{
            𝓡 
         }{
            𝓝 ∪ \{〈r, c, \GetRes(i, k, v, p)〉 \}
         }
         \xrightarrow{c \ ▹ \ \sGet[i]{}{k} : v}_{𝐈}
         \world{
            𝓒 [c ↦ 〈σ', s[v / x]〉]
         }{
            𝓡 
         }{
            𝓝  
         }
      }

      \indrule[Put-Req] {
         c ≠ c₀
         \\\\      
         \putReq (k, v) (c, σ) ⇝ ⃰ 〈p, σ'〉
      } {
         \world{
            𝓒 [c ↦ (σ, \sPut{k}{v}; s)]
         }{
            𝓡 
         }{
            𝓝 
         }
         \xrightarrow{c \ ▹ \ \sPut{k}{v}}_{𝐈}
         \world{
            𝓒 [c ↦ (σ', s)]
         }{
            𝓡 
         }{
            𝓝 ∪ \{ \overline{〈c, r, \PutReq( k, v, p )〉 }_{r ∈ R}  \}
         }
      }
   
      \indrule[Put] {
         \putGuard (k, v) (p) (c, r, ς) ⇝ ⃰  \true
         \\\\
         \lput (k, v) (p) (c, r, ς) ⇝ ⃰  ς'
      } {
         \world{
            𝓒 
         }{
            𝓡 [r ↦ ς]
         }{
            𝓝  ∪ \{ 〈c, r, \PutReq( k, v, p ) 〉\}
         }
         \xrightarrow{r \ ▹ \ \sPut{k}{v}}_{𝐈}
         \world{
            𝓒 
         }{
            𝓡 [r ↦ ς']
         }{
            𝓝 
         }
      }
   \end{mathpar}
\end{minipage}%
}
   \caption{\label{fig:updated-conc-op-sem}
      Key-\rev{V}\del{v}alue Store Operational Semantics $→_{𝐈}$
      for the implementation      
      $𝐈 =$ $(\CState,$ $\cInit,$ $\RState,$ $\rInit,$
         $\getReq,$ $\getGuard,$ $\get,$ $\getRes,$
         $\putReq,$ $\putGuard,$ $\lput)$.         
   }
\end{figure}

The $\sGet{}{}$ operations are synchronous. 
Therefore, the semantics implicitly preserves the order of $\sGet{}{}$s and succeeding $\sGet{}{}$s and $\sPut{}{}$s.

\begin{definition}[Trace Inclusion]
An implementation $𝐈₂$ trace-includes another $𝐈₁$, written $𝐈₂ ⊑ 𝐈₁$, if
for all $W₂$, and $h₂$, 
if $W₀(a)_{𝐈₂} \xrightarrow{h₂}_{𝐈₂} ⃰  W₂$,
then there \rev{exist}\del{exists} $h₁$ and $W₁$ such that
$W₀(a)_{𝐈₁} \xrightarrow{h₁}_{𝐈₁} ⃰  W₁$
and
$\ext(h₁) = \ext(h₂)$.
\end{definition}

\begin{definition}[Convergence]
An implementation $𝐈$ is convergent if
for all $𝓡$,  %
if $W₀(a)_{𝐈} \xrightarrow{}_{𝐈} ⃰  〈\_ , 𝓡 , ∅ 〉$,
then for all $k$, $p$, $c$, $r₁$, $r₂$ and $v$,
if
$\getGuard (k) (p) (c, r₁, 𝓡 [r₁]) ⇝ ⃰  \true$ and
$\get (k) (p) (c, r₁, 𝓡 [r₁]) ⇝ ⃰  〈v, \_, \_〉$
then
$\getGuard (k) (p) (c, r₂, 𝓡 [r₂]) ⇝ ⃰  \true$ and
$\get (k) (p) (c, r₂, 𝓡 [r₂]) ⇝ ⃰  〈v, \_, \_〉$.
\end{definition}

\clearpage
\begin{figure}[h]
  \centering
  \begin{tikzpicture}[
    every node/.style={font=\large},
    arr/.style={-{Triangle[open, length=3mm]}, thick},
  ]

  \node[draw] (CC1)    at (-1.5, 0)  {$\mathbf{I}_{\CC_1}$};
  \node[draw] (CC2)    at ( 1.5, 0)  {$\mathbf{I}_{\CC_2}$};

  \node[draw] (RYWMW)  at (-5,  2.5) {$\mathbf{I}_{\RYW\text{-}\MW}$};
  \node[draw] (CCstar) at ( 0,  2.5) {$\mathbf{I}_{\CC}^{*}$};

  \node[draw] (RYW)    at (-4, 5)    {$\mathbf{I}_{\RYW}$};
  \node[draw] (MW)     at ( 0, 5)    {$\mathbf{I}_{\MW}$};
  \node[draw] (MR)     at ( 4, 5)    {$\mathbf{I}_{\MR}$};
  \node[draw] (LCCstar) at ( 2, 3.5)   {$\mathbf{I}_{\LCC}^{*}$};

  \node[draw] (RYWstar) at (-4, 7.5) {$\mathbf{I}_{\RYW}^{*}$};
  \node[draw] (MWstar)  at ( 0, 7.5) {$\mathbf{I}_{\MW}^{*}$};
  \node[draw] (MRstar)  at ( 4, 7.5) {$\mathbf{I}_{\MR}^{*}$};

  \node[draw] (Rel)     at ( 0, 10)  {$\mathbf{I}_{\mathrm{Rel}}$};

  \draw[arr] (CC1)     -- (CCstar);
  \draw[arr] (CC2)     -- (CCstar);

  \draw[arr] (RYWMW.north) to[out=90, in=225] (RYWstar.south west);

  \draw[arr] (RYWMW)   -- (MWstar);

  \draw[arr] (CCstar)  -- (RYWstar);
  \draw[arr] (CCstar)  -- (MRstar);
  \draw[arr] (CCstar)  -- (LCCstar);

  \draw[arr] (RYW)     -- (RYWstar);
  \draw[arr] (MW)      -- (MWstar);
  \draw[arr] (MR)      -- (MRstar);

  \draw[arr] (RYWstar) -- (Rel);
  \draw[arr] (MWstar)  -- (Rel);
  \draw[arr] (MRstar)  -- (Rel);

  \end{tikzpicture}

  \vspace{0.5em}
  {\small
  \begin{tabular}{@{}ll@{}}
    $\mathbf{I}_{\mathrm{Rel}}$        & Relaxed Specification (Fig.~\ref{fig:relaxed-specification}) \\
    $\mathbf{I}_{\RYW}^{*}$           & \rev{Read-Your-Writes}\del{Read Your Writes} Specification (Fig.~\ref{fig:read-your-writes-spec}) \\
    $\mathbf{I}_{\RYW}$               & \rev{Read-Your-Writes}\del{Read Your Writes} Implementation (Fig.~\ref{fig:read-your-writes-impl}) \\
    $\mathbf{I}_{\MR}^{*}$            & Monotonic Reads Specification (Fig.~\ref{fig:monotonic-reads-spec}) \\
    $\mathbf{I}_{\MR}$                & Monotonic Reads Implementation (Fig.~\ref{fig:monotonic-reads-impl}) \\
    $\mathbf{I}_{\MW}^{*}$            & Monotonic Writes Specification (Fig.~\ref{fig:monotonic-writes-spec}) \\
    $\mathbf{I}_{\MW}$                & Monotonic Writes Implementation (Fig.~\ref{fig:monotonic-writes-impl}) \\
    $\mathbf{I}_{\RYW\text{-}\MW}$    & \rev{Read-Your-Writes}\del{Read Your Writes} and Monotonic Writes Implementation (Fig.~\ref{fig:ryw-mw-impl}) \\
    $\mathbf{I}_{\LCC}^{*}$           & Labeled Causal Consistency \rev{Specification}\del{Spec} (Fig.~\ref{fig:lcc-spec}) \\
    $\mathbf{I}_{\CC}^{*}$            & Causal Consistency \rev{Specification}\del{Spec} (Fig.~\ref{fig:causal-cons-spec}) \\
    $\mathbf{I}_{\CC_1}$              & Causal Consistency Implementation 1 (Fig.~\ref{fig:causal-map-95}) \\
    $\mathbf{I}_{\CC_2}$              & Causal Consistency Implementation 2 (Fig.~\ref{fig:LloydFreedman-protocol}) \\
  \end{tabular}
  }

  \caption{Refinement Hierarchy}
  \label{fig:refinement-hierarchy}
\end{figure}

\clearpage
\begin{figure}
\centering
\specfig{%
\begin{minipage}{0.90\linewidth}
   \figurealgtextsize
   \setlength{\abovedisplayskip}{0pt}%
   \setlength{\abovedisplayshortskip}{0pt}%
   \setlength{\belowdisplayskip}{0pt}%
   \setlength{\belowdisplayshortskip}{0pt}%
   \begin{mathpar}
   \begin{array}{|lr|}
      \hline
         𝐈_{\Rel} \ (\name{Relaxed Specification})
      & \\ \hline
         \CState ≔ \Unit
         & \bnote{Client State}
      \\
         \cInit(c) ≔ \rev{○}\del{⊥}
         & \bnote{Client Initial State}
      \\
         \RState ≔ K ↦ V
         & \bnote{Replica State}
      \\
         \rInit(r, v₀) ≔ [\overline{ k ↦ v₀}]
         & \bnote{Replica Initial State}
      \\
      & \\
         \GetReqPayload ≔ \GetResPayload ≔ \Unit
         & \note{Get payload types}
      \\
         \getReq \ (\_) (\_, σ)  ≔ 
         & \note{At client: } \bnote{Get Request}
      \\ \t
         \ret〈○, \rev{σ}\del{p}〉
         & \note{No payload}
      \\
         \getGuard \ (\_) (\_) (\_, \_, \_) ≔
         & \note{At replica: } \bnote{Get Guard}
      \\ \t
         \ret \true
         & \note{}
      \\
         \get \ (k) (\_) (\_, \_, ς)  ≔
         & \note{At replica: } \bnote{Get}
      \\ \t
         \llet \rev{v}\del{〈v, \_〉}← ς (k)
         & \note{}
      \\ \t
         \ret〈 v, ○, ς 〉
         & \note{No payload}
      \\
         \getRes \ (\_, \_) (\_) (\_, σ) ≔
         & \note{At client: } \bnote{Get Response}
      \\ \t
         \ret σ
         & \note{}
      \\
      & \\
         \PutReqPayload ≔ \Unit
         & \note{Put payload types} 
      \\
         \putReq \ (k, \_) (\_, σ) ≔ 
         & \note{At client: }\bnote{Put Request}
      \\ \t
         \ret〈○, σ〉
         & \note{No payload}
      \\
         \putGuard \ (\_, \_) (\_) (\_, \_, \_) ≔ 
         & \note{At replica: } \bnote{Put Guard}
      \\ \t
         \ret \true
         & \note{}          
      \\
         \lput \ (k, v) (\_) (\_, \_, ς) ≔ 
         & \note{At replica: } \bnote{Put}
      \\ \t
         \ret ς [k ↦ v]
      & \\ \hline
   \end{array}
   \end{mathpar}
\end{minipage}%
}
   \caption{Relaxed Specification}
   \label{fig:relaxed-specification}
\end{figure}

\subsection{Read-Your-Writes (RYW)}
A client's read sees its own most recent write. The session tracks every put the client has issued; the get-guard requires the responding replica to have applied every put in that set (\cref{fig:read-your-writes-spec}). The reference implementation stores per-replica state as a function from key to\del{ timestamp}\rev{ (value, client, timestamp)} (\cref{fig:read-your-writes-impl}).

\clearpage
\begin{figure}
   \figurealgtextsize
   \setlength{\abovedisplayskip}{0pt}%
   \setlength{\abovedisplayshortskip}{0pt}%
   \setlength{\belowdisplayskip}{0pt}%
   \setlength{\belowdisplayshortskip}{0pt}%
   \begin{adjustbox}{max width=\linewidth,center}$\begin{array}{|lr|}
      \hline
         𝐈_{\RYW} ⃰  \ (\name{\rev{Read-Your-Writes}\del{Read Your Writes} Specification}) 
      & \\ \hline
         T ≔ \Nat
         & \note{Timestamp}
      \\
         \CState ≔ 
         & \bnote{Client State}
      \\ \t
         \Set[\rev{T × K}\del{K × T}]
         & \note{Session puts. ($T$ uniquely identifies puts in a client.)}
      \\
         \cInit(c) ≔
         & \bnote{Client Initial State}
      \\ \t
         ∅
         & \note{No initial puts}
      \\
         \RState ≔
         & \bnote{Replica State}
      \\ \t
         K ↦ (V × C × T \, × 
         & \note{Map from each key $k$ to its value $v$, its origin $〈c, t〉$,}
      \\ \t \phantom{K ↦ (}
         \Set[T])
         & \note{and timestamps of puts on $k$ before it.}
      \\
         \rInit(r, v₀) ≔ 
         & \bnote{Replica Initial State}
      \\ \t
         [\overline{ k ↦ 〈v₀, c₀, 0, ∅〉}]
         & \note{}
      \\
      & \\
         \GetReqPayload ≔ \Set[T]
         & \note{Get payload types}
      \\
         \GetResPayload ≔ \Unit
      & \\
         \getReq \ (k) (\_, p)  ≔ 
         & \note{At client: } \bnote{Get Request}
      \\ \t
         \ret〈p|ₖ, p〉
         & \note{Session past puts as request payload, State unchanged}
      \\
         \getGuard \ (k)(p)(c, \_, ς) ≔
         & \note{At replica: } \bnote{Get Guard}
      \\ \t
         \llet〈\_, c', t', p'〉← ς (k)
      & \\ \t
         \ret c = c' ⇒ p ⊆ p' ∪ \{ t'\}
         & \note{If stored value is from requesting client, his puts not missed.}
      \\
         \get \ (k) (\_) (\_, \_, ς)  ≔
         & \note{At replica: } \bnote{Get}
      \\ \t
         \llet〈v, \_, \_, \_〉← ς (k)
         & \note{}
      \\ \t
         \ret〈 v, ○, ς 〉
         & \note{No payload}
      \\
         \getRes \ (\_, \_) (\_) (\_, p) ≔
         & \note{At client: } \bnote{Get Response}
      \\ \t
         \ret p
         & \note{State unchanged.}
      \\
      & \\
         \PutReqPayload ≔ T × \Set[T]
         & \note{Put payload types}
      \\
         \putReq \ (k, \_) (\self, p) ≔ 
         & \note{At client: }\bnote{Put Request}
      \\ \t
         \llet t ← |p| + 1
         & \note{Current put timestamp}
      \\ \t
         \llet p' ← p ∪ \{〈t, k〉\}
         & \note{The put is added to session puts.}
      \\ \t
         \ret〈〈t, p|ₖ〉, p'〉
         & \note{Session puts are sent as request payload.}
      \\
         \putGuard \ (\_, \_) (\_) (\_, \_, \_) ≔ 
         & \note{At replica: } \bnote{Put Guard}
      \\ \t
         \ret \true
         & \note{}          
      \\
         \lput \ (k, v) (〈t, p〉) (c, \_, ς) ≔ 
         & \note{At replica: } \bnote{Put}
      \\ \t
         \ret ς [k ↦ 〈v, c, t, p〉]
      & \\
      & \\
      p|ₖ ≔ \{ t \ | \ 〈t, k〉∈ p \}
      & \\ \hline
   \end{array}$\end{adjustbox}
   \caption{\rev{Read-Your-Writes}\del{Read Your Writes} Specification}
   \label{fig:read-your-writes-spec}
\end{figure}

\clearpage
\begin{figure}
   \figurealgtextsize
   \setlength{\abovedisplayskip}{0pt}%
   \setlength{\abovedisplayshortskip}{0pt}%
   \setlength{\belowdisplayskip}{0pt}%
   \setlength{\belowdisplayshortskip}{0pt}%
   \begin{adjustbox}{max width=\linewidth,center}$\begin{array}{|lr|}
      \hline
         𝐈_{\RYW} \ (\name{\rev{Read-Your-Writes}\del{Read Your Writes} Implementation}) 
      & \\ \hline
         T ≔ \Nat
         & \note{Timestamp}
      \\
         \CState ≔ 
         & \bnote{Client State}
      \\ \t
         K ↦ T
         & \note{Latest put \rev{per key ($K × T$ is unique within a client).}\del{for each key. ($T$ uniquely identifies puts in a client.)}}
      \\
         \cInit(c) ≔
         & \bnote{Client Initial State}
      \\ \t
         [ \overline{k ↦ 0} ]
         & \note{No initial puts}
      \\
         \RState ≔
         & \bnote{Replica State}
      \\ \t
         K ↦ (V × C × T) 
         & \note{Map from each key $k$ to its value $v$, and its origin $〈c, t〉$}
      \\
         \rInit(r, v₀) ≔ 
         & \bnote{Replica Initial State}
      \\ \t
         [\overline{ k ↦ 〈v₀, c₀, 0〉}]
         & \note{}
      \\
      & \\
         \GetReqPayload ≔ T
         & \note{Get payload types}
      \\
         \GetResPayload ≔ \Unit
      & \\
         \getReq \ (k) (\_, p)  ≔ 
         & \note{At client: } \bnote{Get Request}
      \\ \t
         \ret〈p(k), p〉
         & \note{Timestamp of latest put as request payload, State unchanged}
      \\
         \getGuard \ (k)(t)(c, \_, ς) ≔
         & \note{At replica: } \bnote{Get Guard}
      \\ \t
         \llet〈\_, c', t'〉← ς (k)
      & \\ \t
         \ret  c = c'  ⇒  t = t' 
         & \note{If stored value is from requesting client, his puts not missed.}
      \\
         \get \ (k) (\_) (\_, \_, ς)  ≔
         & \note{At replica: } \bnote{Get}
      \\ \t
         \llet〈v, \_, \_〉← ς (k)
         & \note{}
      \\ \t
         \ret〈 v, ○, ς 〉
         & \note{No payload}
      \\
         \getRes \ (\_, \_) (\_) (\_, p) ≔
         & \note{At client: } \bnote{Get Response}
      \\ \t
         \ret p
         & \note{State unchanged.}
      \\
      & \\
         \PutReqPayload ≔ T
         & \note{Put payload types}
      \\
         \putReq \ (k, \_) (\self, p) ≔ 
         & \note{At client: }\bnote{Put Request}
      \\ \t
         \llet t ← p(k) + 1
         & \note{Current put timestamp}
      \\ \t
         \llet p' ← p [ k ↦ t ]
         & \note{The put timestamp is stored as the latest for the key.}
      \\ \t
         \ret〈t, p'〉
         & \note{Latest put timestamp sent as request payload.}
      \\
         \putGuard \ (\_, \_) (\_) (\_, \_, \_) ≔ 
         & \note{At replica: } \bnote{Put Guard}
      \\ \t
         \ret \true
         & \note{}          
      \\
         \lput \ (k, v) (t) (c, \_, ς) ≔
         & \note{At replica: } \bnote{Put}
      \\ \t
         \ret ς [k ↦ 〈v, c, t〉]
      & \\ \hline
   \end{array}$\end{adjustbox}
   \caption{\rev{Read-Your-Writes}\del{Read Your Writes} Implementation}
   \label{fig:read-your-writes-impl}
\end{figure}

\subsection{Monotonic Reads (MR)}
A client's reads respect the order of writes the client has previously observed. The session keeps the set of writes seen so far; subsequent reads must return values applied no earlier than every write in that set (\cref{fig:monotonic-reads-spec}). The reference implementation stores per-replica state as a function from key to (value, client, timestamp); the\del{ get-guard checks every entry against the client's session set on every read}\rev{ session keeps, per key and writing client, the latest timestamp read, and the get-guard requires the stored write to be no older than the session's timestamp for its writing client} (\cref{fig:monotonic-reads-impl}).

\clearpage
\begin{figure}
   \figurealgtextsize
   \setlength{\abovedisplayskip}{0pt}%
   \setlength{\abovedisplayshortskip}{0pt}%
   \setlength{\belowdisplayskip}{0pt}%
   \setlength{\belowdisplayshortskip}{0pt}%
   \begin{adjustbox}{max width=\linewidth,center}$\begin{array}{|lr|}
      \hline
         𝐈_{\MR} ⃰  \ (\name{Monotonic Reads Specification})
      & \\ \hline
         T ≔ \Nat
         & \note{Timestamp}
      \\
         \CState ≔ 
         & \bnote{Client State}
      \\ \t
         \Set[T × K] \, ×
         & \note{Session puts. (\rev{$T$ uniquely identifies puts in a client}\del{$C × T$ uniquely identifies puts}.)}
      \\ \t 
         \Set[C × T × K]
         & \note{Session dependencies.}         
      \\
         \cInit(c) ≔
         & \bnote{Client Initial State}
      \\ \t
         \rev{〈∅, ∅〉}\del{∅}
         & \note{No initial dependencies}
      \\
         \RState ≔
         & \bnote{Replica State}
      \\ \t
         K ↦ (V × C × T \, ×
         & \note{Map from each key $K$ to its value $V$, its origin $C × T$}
      \\ \t \phantom{K ↦ (}
         \Set[T  × K])
         & \note{and puts before it}
      \\
         \rInit(r, v₀) ≔ 
         & \bnote{Replica Initial State}
      \\ \t
         [\overline{ k ↦ 〈v₀, c₀, 0, ∅〉}]
         & \note{}
      \\
      & \\
         \GetReqPayload ≔ \Set[C × T]
         & \note{Get payload types}
      \\
         \GetResPayload ≔ C × T
         & \note{}
      \\
         \getReq \ (k) (\_, 〈p, d〉)  ≔ 
         & \note{At client: } \bnote{Get Request}
      \\ \t
         \ret〈d|ₖ, 〈p, d〉〉
         & \note{Session deps for $k$ sent as payload, State unchanged}
      \\
         \getGuard \ (k) (d) (c, \_, ς) ≔
         & \note{At replica: } \bnote{Get Guard}
      \\ \t
         \llet〈\_, c', t', d'〉← ς (k)
      & \\ \t
         \ret d|_{c'} ⊆ d'|ₖ ∪ \{ t' \}
         & \note{Stored put from $c'$ doesn't miss observed puts from $c'$.}
      \\
         \get \ (k) (\_) (\_, \_, ς)  ≔
         & \note{}
      \\ \t
         \llet〈v, c, t, d〉← ς (k)
      & \\ \t
         \ret〈 v, 〈c, t〉, ς 〉
         & \note{The stored put returned as a dependency.}
      \\
         \getRes \ (k, \_) (〈c, t〉) (\_, 〈p, d〉) ≔
         & \note{At client: } \bnote{Get Response}
      \\ \t
         \ret 〈p, d ∪ \{ 〈c, t, k〉\}〉
         & \note{Returned dependency added to session dependencies.}
      \\
      & \\
         \PutReqPayload ≔ T × \Set[T × K]
         & \note{Put payload types}
      \\
         \putReq \ (k, \_) (\_, 〈p, d〉) ≔ 
         & \note{At client: }\bnote{Put Request}
      \\ \t
         \llet t ← | p | + 1
         & \note{Current put timestamp}
      \\ \t
         \llet p' ← p ∪ \{〈t, k〉\}
         & \note{The put is added to session puts.}
      \\ \t
         \ret〈〈t, p〉, 〈p', d〉〉
         & \note{Session puts are sent as request payload.}
      \\
         \putGuard \ (\_, \_) (\_) (\_, \_, \_) ≔ 
         & \note{At replica: } \bnote{Put Guard}
      \\ \t
         \ret \true
         & \note{}          
      \\
         \lput \ (k, v) (〈t, p〉) (c, \_, ς) ≔ 
         & \note{At replica: } \bnote{Put}
      \\ \t
         \ret ς [k ↦ 〈v, c, t, p〉]
      & \\
      & \\
      d|ₖ ≔ \{〈c, t〉 \ | \ 〈c, t, k〉∈ d \}
      & \\
      d|ₖ ≔ \{ t \ | \ 〈t, k〉∈ d \}
      & \\
      d|_c  ≔ \{ t \ | \ 〈c, t〉∈ d \}
      & \\ \hline
   \end{array}$\end{adjustbox}
   \caption{Monotonic Reads Specification}
   \label{fig:monotonic-reads-spec}
\end{figure}

\clearpage
\begin{figure}
   \figurealgtextsize
   \setlength{\abovedisplayskip}{0pt}%
   \setlength{\abovedisplayshortskip}{0pt}%
   \setlength{\belowdisplayskip}{0pt}%
   \setlength{\belowdisplayshortskip}{0pt}%
   \begin{adjustbox}{max width=\linewidth,center}$\begin{array}{|lr|}
      \hline
         𝐈_{\MR} \ (\name{Monotonic Reads Implementation})
      & \\ \hline
         T ≔ \Nat
         & \note{Timestamp}
      \\
         \CState ≔ 
         & \bnote{Client State}
      \\ \t
         T \ ×
         & \note{Session latest put timestamp}
      \\ \t
         K ↦ C ↦ T
         & \note{Largest timestamp read for the key from the client.}         
      \\
         \cInit(c) ≔
         & \bnote{Client Initial State}
      \\ \t
         \rev{〈0, ∅〉}\del{〈∅, ∅〉}
         & \note{No initial puts and dependencies}
      \\
         \RState ≔
         & \bnote{Replica State}
      \\ \t
         K ↦ (V × C × T)
         & \note{Map from each key $K$ to its value $V$, its origin $C × T$}
      \\
         \rInit(r, v₀) ≔ 
         & \bnote{Replica Initial State}
      \\ \t
         [\overline{ k ↦ 〈v₀, c₀, 0〉}]
         & \note{}
      \\
      & \\
         \GetReqPayload ≔ \Set[C × T]
         & \note{Get payload types}
      \\
         \GetResPayload ≔ C × T
         & \note{}
      \\
         \getReq \ (k) (\_, 〈t, d〉)  ≔ 
         & \note{At client: } \bnote{Get Request}
      \\ \t
         \ret〈d(k), 〈t, d〉〉
         & \note{Session deps for $k$ sent as payload, State unchanged}
      \\
         \getGuard \ (k) (d) (c, \_, ς) ≔
         & \note{At replica: } \bnote{Get Guard}
      \\ \t
         \llet〈\_, c', t'〉← ς (k)
      & \\ \t
         \ret d(c') ≤  t'
         & \note{Stored put from $c'$ doesn't miss observed puts from $c'$.}
      \\
         \get \ (k) (\_) (\_, \_, ς)  ≔
         & \note{}
      \\ \t
         \llet〈v, c, t〉← ς (k)
      & \\ \t
         \ret〈 v, 〈c, t〉, ς 〉
         & \note{The stored put returned as a dependency.}
      \\
         \getRes \ (k, \_) (〈c, t〉) (\_, 〈\rev{t'}\del{t}, d〉) ≔
         & \note{At client: } \bnote{Get Response}
      \\ \t
         \ret〈\rev{t'}\del{t}, d[〈k,c〉 ↦ t] 〉
         & \note{Returned dependency advances session dependencies.}
      \\
      & \\
         \PutReqPayload ≔ T
         & \note{Put payload types}
      \\
         \putReq \ (k, \_) (\_, 〈t, d〉) ≔ 
         & \note{At client: }\bnote{Put Request}
      \\ \t
         \llet t' ← t + 1
         & \note{Current put timestamp}
      \\ \t
         \ret〈t', 〈t', d〉〉
         & \note{}
      \\
         \putGuard \ (\_, \_) (\_) (\_, \_, \_) ≔ 
         & \note{At replica: } \bnote{Put Guard}
      \\ \t
         \ret \true
         & \note{}          
      \\
         \lput \ (k, v) (t) (c, \_, ς) ≔ 
         & \note{At replica: } \bnote{Put}
      \\ \t
         \ret ς [k ↦ 〈v, c, t〉]
      & \\ \hline
   \end{array}$\end{adjustbox}
   \caption{Monotonic Reads Implementation}
   \label{fig:monotonic-reads-impl}
\end{figure}

\subsection{Monotonic Writes (MW)}
A single client's writes are observed in order at every replica that observes them. Each replica records, per\del{ writing client, the latest applied timestamp}\rev{ key, the set of puts already applied}; the put-guard rejects any out-of-order delivery (\cref{fig:monotonic-writes-spec}). The reference implementation\del{ prepends the timestamp of each new write to a per-key list with no de-duplication, so receivers can re-check ordering on every delivery}\rev{ instead keeps, per key, the latest applied timestamp from each writing client; its put-guard admits only the next put from that client} (\cref{fig:monotonic-writes-impl}).

\clearpage
\begin{figure}
   \figurealgtextsize
   \setlength{\abovedisplayskip}{0pt}%
   \setlength{\abovedisplayshortskip}{0pt}%
   \setlength{\belowdisplayskip}{0pt}%
   \setlength{\belowdisplayshortskip}{0pt}%
   \begin{adjustbox}{max width=\linewidth,center}$\begin{array}{|lr|}
      \hline
         𝐈_{\MW} ⃰  \ (\name{Monotonic Writes Specification}) 
      & \\ \hline
         T ≔ \Nat
         & \note{Timestamp}
      \\
         \CState ≔ 
         & \bnote{Client State}
      \\ \t
         \Set[T × K]
         & \note{Session puts. (\rev{$T$ uniquely identifies puts in a client}\del{$C × T$ uniquely identifies puts}.)}
      \\
         \cInit(c) ≔
         & \bnote{Client Initial State}
      \\ \t
         ∅
         & \note{No initial dependencies}
      \\
         \RState ≔
         & \bnote{Replica State}
      \\ \t
         K ↦ (V × \Set[C × T])
         & \note{Map from each key $K$ to its value $V$, and puts already applied.}
      \\
         \rInit(r, v₀) ≔ 
         & \bnote{Replica Initial State}
      \\ \t
         [\overline{ k ↦ 〈v₀, ∅〉}]
         & \note{}
      \\
      & \\
         \GetReqPayload ≔ \GetResPayload ≔ \Unit
         & \note{Get payload types}
      \\
         \getReq \ (\_) (\_, p)  ≔ 
         & \note{At client: } \bnote{Get Request}
      \\ \t
         \ret〈○, p〉
         & \note{\rev{No payload}\del{Session past puts as request payload}, State unchanged}
      \\
         \getGuard \ (\_) (\_) (\_, \_, \_) ≔
         & \note{At replica: } \bnote{Get Guard}
      \\ \t
         \ret \true
         & \note{}
      \\
         \get \ (k) (\_) (\_, \_, ς)  ≔
         & \note{At replica: } \bnote{Get}
      \\ \t
         \llet〈v, \_〉← ς (k)
         & \note{}
      \\ \t
         \ret〈 v, ○, ς 〉
         & \note{No payload}
      \\
         \getRes \ (\_, \_) (\_) (\_, p) ≔
         & \note{At client: } \bnote{Get Response}
      \\ \t
         \ret p
         & \note{State unchanged.}
      \\
      & \\
         \PutReqPayload ≔ T × \Set[T]
         & \note{Put payload types}
      \\
         \putReq \ (k, \_) (\_, p) ≔ 
         & \note{At client: }\bnote{Put Request}
      \\ \t
         \llet t ← |p| + 1
         & \note{Current put timestamp}
      \\ \t
         \llet p' ← p ∪ \{〈t, k〉\}         
         & \note{The put is added to session puts.}
      \\ \t
         \ret〈〈t, p|ₖ〉, p'〉
         & \note{Session puts for the key are sent as request payload.}
      \\
         \putGuard \ (k, \_) (〈t, p〉) (c, \_, ς) ≔ 
         & \note{At replica: } \bnote{Put Guard}
      \\ \t
         \llet〈\_, p'〉← ς (k)
      & \\ \t
         \ret p ⊆ p'|_c
         & \note{Previous puts of session already applied.}
      \\
         \lput \ (k, v) (〈t, \_〉) (c, \_, ς) ≔ 
         & \note{At replica: } \bnote{Put}
      \\ \t
         \llet〈\_, p'〉← ς (k)
         & \note{}
      \\ \t
         \ret ς [k ↦ 〈v, p' ∪ \{〈c,t〉\}\rev{〉}]
      & \\
      & \\
      p|ₖ ≔ \{ t \ | \ 〈t, k〉∈ p \}
      & \\
      p|_c  ≔ \{ t \ | \ 〈c, t〉∈ p \}
      & \\ \hline
   \end{array}$\end{adjustbox}
   \caption{Monotonic Writes Specification}
   \label{fig:monotonic-writes-spec}
\end{figure}

\clearpage
\begin{figure}
   \figurealgtextsize
   \setlength{\abovedisplayskip}{0pt}%
   \setlength{\abovedisplayshortskip}{0pt}%
   \setlength{\belowdisplayskip}{0pt}%
   \setlength{\belowdisplayshortskip}{0pt}%
   \begin{adjustbox}{max width=\linewidth,center}$\begin{array}{|lr|}
      \hline
         𝐈_{\MW} \ (\name{Monotonic Writes Implementation}) 
      & \\ \hline
         T ≔ \Nat
         & \note{Timestamp}
      \\
         \CState ≔ 
         & \bnote{Client State}
      \\ \t         
         K ↦ T
         & \note{Latest timestamp of puts for the key.}
      \\
         \cInit(c) ≔
         & \bnote{Client Initial State}
      \\ \t
         \overline{k ↦ 0}
         & \note{No initial puts}
      \\
         \RState ≔
         & \bnote{Replica State}
      \\ \t
         K ↦ (V \ ×
         & \note{Map from each key $K$ to its value $V$, and}
      \\ \t \phantom{K ↦ (}
         C ↦ T)
         & \note{latest puts applied from sessions.}
      \\
         \rInit(r, v₀) ≔ 
         & \bnote{Replica Initial State}
      \\ \t
         [\overline{ k ↦ 〈v₀, \overline{c ↦ 0}〉}]
         & \note{}
      \\
      & \\
         \GetReqPayload ≔ \GetResPayload ≔ \Unit
         & \note{Get payload types}
      \\
         \getReq \ (\_) (\_, σ)  ≔ 
         & \note{At client: } \bnote{Get Request}
      \\ \t
         \ret〈○, σ〉
         & \note{\rev{No payload}\del{Session past puts as request payload}, State unchanged}
      \\
         \getGuard \ (\_) (\_) (\_, \_, \_) ≔
         & \note{At replica: } \bnote{Get Guard}
      \\ \t
         \ret \true
         & \note{}
      \\
         \get \ (k) (\_) (\_, \_, ς)  ≔
         & \note{At replica: } \bnote{Get}
      \\ \t
         \llet〈v, \_〉← ς (k)
         & \note{}
      \\ \t
         \ret〈 v, ○, ς 〉
         & \note{No payload}
      \\
         \getRes \ (\_, \_) (\_) (\_, σ) ≔
         & \note{At client: } \bnote{Get Response}
      \\ \t
         \ret σ
         & \note{State unchanged.}
      \\
      & \\
         \PutReqPayload ≔ T
         & \note{Put payload types}
      \\
         \putReq \ (k, \_) (\_, p) ≔ 
         & \note{At client: }\bnote{Put Request}
      \\ \t
         t ← p(k)
         & \note{}
      \\ \t
         \llet p' ← p [ k ↦ t + 1]
         & \note{The timestamp of the latest put for key is advanced.}
      \\ \t
         \ret〈t, p'〉
         & \note{Latest put timestamp for the key is sent as request payload.}
      \\
         \putGuard \ (k, \_) (t) (c, \_, ς) ≔ 
         & \note{At replica: } \bnote{Put Guard}
      \\ \t
         \llet〈\_, p'〉← ς (k)
      & \\ \t
         \ret  p'(c) = t
         & \note{Previous puts of session already applied.}
      \\
         \lput \ (k, v) (t) (c, \_, ς) ≔ 
         & \note{At replica: } \bnote{Put}
      \\ \t
         \llet〈\_, p'〉← ς (k)
         & \note{}
      \\ \t
         \ret ς [k ↦ 〈v, p' [c ↦ t + 1]〉]
         & \note{}
      \\ \hline
   \end{array}$\end{adjustbox}
   \caption{Monotonic Writes Implementation}
   \label{fig:monotonic-writes-impl}
\end{figure}

\subsection{Read-Your-Writes $+$ Monotonic Writes (RYW+MW)}
Composition of Read-Your-Writes and Monotonic Writes. RYW+MW is the conjunction of RYW and MW rather than an independent property; an implementation satisfies it iff it refines both, so correctness is the joint refinement obligation. The reference combines RYW's read-side check with MW's compare-and-swap on every put (\cref{fig:ryw-mw-impl}).

\clearpage
\begin{figure}
   \figurealgtextsize
   \setlength{\abovedisplayskip}{0pt}%
   \setlength{\abovedisplayshortskip}{0pt}%
   \setlength{\belowdisplayskip}{0pt}%
   \setlength{\belowdisplayshortskip}{0pt}%
   \begin{adjustbox}{max width=\linewidth,center}$\begin{array}{|lr|}
      \hline
         \multicolumn{2}{|l|}{𝐈_{\RYW\text{-}\MW} \ (\name{\rev{Read-Your-Writes}\del{Read Your Writes} and Monotonic Writes Implementation}) }         
      \\ \hline
         T ≔ \Nat
         & \note{Timestamp}
      \\
         \CState ≔ 
         & \bnote{Client State}
      \\ \t         
         K ↦ T
         & \note{Latest timestamp of puts for the key.}
      \\
         \cInit(c) ≔
         & \bnote{Client Initial State}
      \\ \t
         \overline{k ↦ 0}
         & \note{No initial puts}
      \\
         \RState ≔
         & \bnote{Replica State}
      \\ \t
         K ↦ (V × C \ × 
         & \note{Map from each key $K$ to its value $V$, and its originating sessions $C$,}
      \\ \t \phantom{K ↦ (}
         C ↦ T)
         & \note{latest puts applied from sessions}
      \\
         \rInit(r, v₀) ≔ 
         & \bnote{Replica Initial State}
      \\ \t
         [\overline{ k ↦ 〈v₀, \rev{c₀}\del{0}, \overline{c ↦ 0}〉}]
         & \note{}
      \\
      & \\
         \GetReqPayload ≔ T
         & \note{Get payload types}
      \\
         \GetResPayload ≔ \Unit
      & \\
         \getReq \ (k) (\_, p)  ≔ 
         & \note{At client: } \bnote{Get Request}
      \\ \t
         \ret〈p(k), p〉
         & \note{Timestamp of latest put as request payload, State unchanged}
      \\
         \getGuard \ (k)(t)(c, \_, ς) ≔
         & \note{At replica: } \bnote{Get Guard}
      \\ \t
         \llet〈\_, c', p'〉← ς (k)
      & \\ \t
         \ret  c = c'  ⇒  p'(c') = t
         & \note{If stored value is from requesting client, his puts not missed.}
      \\
         \get \ (k) (\_) (\_, \_, ς)  ≔
         & \note{At replica: } \bnote{Get}
      \\ \t
         \llet〈v, \_, \_〉← ς (k)
         & \note{}
      \\ \t
         \ret〈 v, ○, ς 〉
         & \note{No payload}
      \\
         \getRes \ (\_, \_) (\_) (\_, p) ≔
         & \note{At client: } \bnote{Get Response}
      \\ \t
         \ret p
         & \note{State unchanged.}
      \\
      & \\
         \PutReqPayload ≔ T
         & \note{Put payload types}
      \\
         \putReq \ (k, \_) (\_, p) ≔ 
         & \note{At client: }\bnote{Put Request}
      \\ \t
         t ← p(k)
         & \note{}
      \\ \t
         \llet p' ← p [ k ↦ t + 1 ]
         & \note{The timestamp of the latest put for key is advanced.}
      \\ \t
         \ret〈t, p'〉
         & \note{Latest put timestamp for the key is sent as request payload.}
      \\
         \putGuard \ (k, \_) (t) (c, \_, ς) ≔ 
         & \note{At replica: } \bnote{Put Guard}
      \\ \t
         \llet〈\_, \_, p'〉← ς (k)
      & \\ \t
         \ret p'(c) = t
         & \note{Previous puts of session already applied.}
      \\
         \lput \ (k, v) (t) (c, \_, ς) ≔ 
         & \note{At replica: } \bnote{Put}
      \\ \t
         \llet〈\_, \rev{\_, }p'〉← ς (k)
         & \note{}
      \\ \t
         \ret ς [k ↦ 〈v, c, p' [c ↦ t + 1]〉]
         & \note{}
      \\ \hline
   \end{array}$\end{adjustbox}
   \caption{\rev{Read-Your-Writes}\del{Read Your Writes} and Monotonic Writes Implementation}
   \label{fig:ryw-mw-impl}
\end{figure}

\subsection{Causal Consistency (CC) and Labeled Causal Consistency (LCC)}
\textbf{Causal Consistency (CC).}
Cross-client causal consistency: every read returns a value whose causal predecessors have all been observed at the responding replica. Dependencies are tracked as explicit sets in the spec state (\cref{fig:causal-cons-spec}), which the implementation must bridge to its concrete representation (typically a per-client vector clock plus a compare-and-swap on the writer's own slot during put). Two reference implementations are given: a per-client vector-clock store (\cref{fig:causal-map-95}) and a Lloyd--Freedman-style protocol (\cref{fig:LloydFreedman-protocol}).

\noindent\textbf{Labeled Causal Consistency (LCC).}
Causal consistency \rev{parameterized}\del{parameterised} by per-message topic labels and per-client interest masks. Every put is tagged with a label, and every client declares which labels it cares about; the get-guard requires the responding replica to have observed every dependency whose label is in the requesting client's mask (\cref{fig:lcc-spec}). To our knowledge this is the first \coq{} \rev{formalization}\del{formalisation} of LCC and the first verified implementation.

\clearpage
\begin{figure}
   \figurealgtextsize
   \setlength{\abovedisplayskip}{0pt}%
   \setlength{\abovedisplayshortskip}{0pt}%
   \setlength{\belowdisplayskip}{0pt}%
   \setlength{\belowdisplayshortskip}{0pt}%
   \begin{adjustbox}{max width=\linewidth,center}$\begin{array}{|lr|}
      \hline
         𝐈_{\CC} ⃰  \ (\name{Causal Consistency \rev{Specification}\del{Spec}}) 
      & \\ \hline
         T ≔ \Nat
         & \note{Timestamp}
      \\
         \CState ≔ 
         & \bnote{Client State}
      \\ \t
         〈\Set[C × T × K],
         & \note{Session dependencies from puts.}
      \\ \t \phantom{〈}
         \Set[\Set[C × T × K]] 〉
         & \note{Session dependencies from gets}
      \\ \t
         & \note{($C × T$ uniquely identifies puts.)}         
      \\
         \cInit(c) ≔
         & \bnote{Client Initial State}
      \\ \t
         〈∅, ∅〉
         & \note{No initial dependencies}
      \\
         \RState ≔
         & \bnote{Replica State}
      \\ \t
         K ↦ (V × C × T \, × 
         & \note{Map from each key $K$ to its value $V$, its origin $C × T$,}
      \\ \t \phantom{K ↦ (}         
         \Set[C × T × K])
         & \note{and set of dependencies}
      \\
         \rInit(r, v₀) ≔ 
         & \bnote{Replica Initial State}
      \\ \t
         [\overline{ k ↦ 〈v₀, c₀, 0, ∅〉}]
         & \note{Empty initial dependencies}
      \\
      & \\
         \GetReqPayload ≔ 〈\Set[C × T × K],
         & \note{Get payload types}
      \\ \phantom{\GetReqPayload ≔ 〈}
         \Set[\Set[C × T × K]]\>
         & \note{}
      \\
         \GetResPayload ≔ \Set[C × T  × K]
      & \\
         \getReq \ (k) (\_, 〈p, G〉)  ≔ 
         & \note{At client: } \bnote{Get Request}
      \\ \t
         \ret〈〈p|ₖ, G|ₖ〉, 〈p, G〉〉 
         & \note{Session deps sent as request payload, State unchanged}
      \\
         \getGuard \ (k) (〈p, G〉) (c, \_, ς) ≔
         & \note{At replica: } \bnote{Get Guard}
      \\ \t
         \llet〈\_, c', t', d'〉← ς (k)
         & \note{}
      \\ \t
         \ret \lforall (λ d. \ ¬ (d'|ₖ ∪ \{〈c', t', k〉\} ⊂ d)) \ G \ ∧
         & \note{Stored value isn't behind any previously read value.}
      \\ \t \phantom{\ret}
         c = c' ⇒ p ⊆ d'|ₖ ∪ \{〈c', t', k〉\}
         & \note{Read your writes.}
      \\
         \get \ (k) (\_) (\_, \_, ς)  ≔
         & \note{At replica: } \bnote{Get}
      \\ \t
         \llet〈v, c, t, d〉← ς (k)
      & \\ \t
         \llet d' ← \lif (c ≠ c₀) \ d ∪ \{〈c, t, k〉\} \ \lelse ∅
         & \note{The dependencies returned to the session}
      \\ \t
         \ret〈 v, d', ς 〉
         & \note{are the entry and its dependencies}
      \\
         \getRes \ (\_, \_) (d') (\_, 〈p, G〉) ≔
         & \note{At client: } \bnote{Get Response}
      \\ \t
         \ret 〈p, G ∪ \{ d' \}〉
         & \note{Received dependencies added to session dependencies.}
      \\
      & \\
         \PutReqPayload ≔ T × \Set [ C × T × K ]
         & \note{Put payload types}
      \\
         \putReq \ (k, \_) (\self, 〈p, G〉) ≔ 
         & \note{At client: }\bnote{Put Request}
      \\ \t
         \llet t ← |p| + 1
         & \note{Current put timestamp}
      \\ \t
         \llet p' ← p ∪ \{〈 \self, t, k 〉\}
         & \note{The put is added to session dependencies.}
      \\ \t
         \ret〈〈t, p ∪ (⋃\, G)〉, 〈p', G〉〉 
         & \note{Session dependencies are sent as request payload.}
      \\
         \putGuard \ (k, \_) (〈t, \_〉) (c, \_, ς) ≔ 
         & \note{At replica: } \bnote{Put Guard}
      \\ \t
         \llet〈\_, \_, \_, d'〉← ς (k)
      & \\ \t
          \ret〈c, t, k〉∉ d'
          & \note{The new put is not causally before the current stored put.}
      \\
         \lput \ (k, v) (〈t, d〉) (c, \_, ς) ≔ 
         & \note{At replica: } \bnote{Put}
      \\ \t
         \ret ς [k ↦ 〈v, c, t, d〉]
      & \\
      & \\
         d|ₖ ≔ \{〈c, t, k〉\ | \ 〈c, t, k'〉∈ d  ∧ k = k' \}
      & \\
         D|ₖ ≔ \{ d|ₖ \ | \ d ∈ D \}
      & \\ \hline
   \end{array}$\end{adjustbox}
   \caption{Causal Consistency \rev{Specification}\del{Spec}}
   \label{fig:causal-cons-spec}
\end{figure}

\clearpage
\begin{figure}
   \figurealgtextsize
   \setlength{\abovedisplayskip}{0pt}%
   \setlength{\abovedisplayshortskip}{0pt}%
   \setlength{\belowdisplayskip}{0pt}%
   \setlength{\belowdisplayshortskip}{0pt}%
   \begin{adjustbox}{max width=\linewidth,center}$\begin{array}{|lr|}
      \hline
         \multicolumn{2}{|l|}{𝐈_{\LCC} ⃰  \ (\name{Labeled Causal Consistency \rev{Specification}\del{Spec}})}
      \\ \hline
         \L
         & \note{Label type}
      \\
         \lab : V → \L
         & \note{Maps each value to its label}
      \\
         \clientLabs : C → \Set[\L]
         & \note{Maps each client to its set of relevant labels}
      \\
         T ≔ \Nat
         & \note{Timestamp}
      \\
         \CState ≔
         & \bnote{Client State}
      \\ \t
         〈\Set[C × T × K × \L],
         & \note{Session dependencies from puts (with labels)}
      \\ \t \phantom{〈}
         \Set[\Set[C × T × K × \L]] 〉
         & \note{Session dependencies from gets (with labels)}
      \\
         \cInit(c) ≔
         & \bnote{Client Initial State}
      \\ \t
         〈∅, ∅〉
         & \note{No initial dependencies}
      \\
         \RState ≔
         & \bnote{Replica State}
      \\ \t
         K ↦ (V × C × T \, ×
         & \note{Map from each key $K$ to its value $V$, its origin $C × T$,}
      \\ \t \phantom{K ↦ (}
         \Set[C × T × K × \L])
         & \note{and set of label-annotated dependencies}
      \\
         \rInit(r, v₀) ≔
         & \bnote{Replica Initial State}
      \\ \t
         [\overline{ k ↦ 〈v₀, c₀, 0, ∅〉}]
         & \note{Empty initial dependencies}
      \\
      & \\
         \GetReqPayload ≔ 〈\Set[C × T × K × \L]\rev{,}
         & \note{Get payload types}
      \\ \phantom{\GetReqPayload ≔ 〈}
         \Set[\Set[C × T × K × \L]]\rev{\>}
      & \\
         \GetResPayload ≔ \Set[C × T × K × \L]
      & \\
         \getReq \ (k) (\self, 〈p, G〉)  ≔
         & \note{At client: } \bnote{Get Request}
      \\ \t
         \llet L ← \clientLabs(\self)
      & \\ \t
         \ret〈〈p|ₖ|_{L}, G|ₖ|_{L}〉, 〈p, G〉〉
         & \note{Label-filtered deps sent as payload; state unchanged.}
      \\
         \getGuard \ (k) (〈p, G〉) (c, \_, ς) ≔
         & \note{At replica: } \bnote{Get Guard}
      \\ \t
         \llet〈v', c', t', d'〉← ς (k)
      & \\ \t
         \llet L ← \clientLabs(c)
      & \\ \t
         \ret \lforall (λ d. \ 
         & \note{Stored value not behind any previously}
      \\ \t \ \ \ \
         ¬ ((d'|ₖ ∪ \{〈c', t', k, \lab(v')〉\})|_{L} ⊂ d)) \ G
         & \note{read value, for client's labels.}
      \\ \t
         ∧ \ c = c' ⇒ p ⊆ d'|ₖ ∪ \{〈c', t', k, \lab(v')〉\}
         & \note{Read your writes.}         
      \\
         \get \ (k) (\_) (\_, \_, ς)  ≔
         & \note{At replica: } \bnote{Get}
      \\ \t
         \llet〈v, c, t, d〉← ς (k)
      & \\ \t
         \llet d' ← \lif (c ≠ c₀) \ d ∪ \{〈c, t, k, \lab(v)〉\} \ \lelse ∅
         & \note{Label of stored value included in returned dependency.}
      \\ \t
         \ret〈 v, d', ς 〉
         & \note{}
      \\
         \getRes \ (\_, \_) (d') (\_, 〈p, G〉) ≔
         & \note{At client: } \bnote{Get Response}
      \\ \t
         \ret 〈p, G ∪ \{ d' \}〉
         & \note{Received dependencies added to session.}
      \\
      & \\
         \PutReqPayload ≔ T × \Set [ C × T × K × \L ]
         & \note{Put payload types}
      \\
         \putReq \ (k, v) (\self, 〈p, G〉) ≔
         & \note{At client: }\bnote{Put Request}
      \\ \t
         \llet t ← |p| + 1
         & \note{Current put timestamp}
      \\ \t
         \llet p' ← p ∪ \{〈 \self, t, k, \lab(v) 〉\}
         & \note{Put with its label added to session dependencies.}
      \\ \t
         \ret〈〈t, p ∪ (⋃\, G)〉, 〈p', G〉〉
         & \note{All session dependencies sent as request payload.}
      \\
         \putGuard \ (k, v) (〈t, \_〉) (c, \_, ς) ≔
         & \note{At replica: } \bnote{Put Guard}
      \\ \t
         \llet〈\_, \_, \_, d'〉← ς (k)
      & \\ \t
          \ret〈c, t, k, \lab(v)〉∉ d'
          & \note{The new put has not already been applied.}
      \\
         \lput \ (k, v) (〈t, d〉) (c, \_, ς) ≔
         & \note{At replica: } \bnote{Put}
      \\ \t
         \ret ς [k ↦ 〈v, c, t, d〉]
      & \\
      & \\
         d|ₖ ≔ \{〈c, t, k, l〉\ | \ 〈c, t, k', l〉∈ d  ∧ k = k' \}
      & \\
         D|ₖ ≔ \{ d|ₖ \ | \ d ∈ D \}
      & \\
         d|_{L} ≔ \{〈c, t, k, l〉\ | \ 〈c, t, k, l〉∈ d  ∧ l ∈ L \}
         & \note{Filter dependencies to the label set $L$}
      \\
         D|_{L} ≔ \{ d|_{L} \ | \ d ∈ D \}
      & \\ \hline
   \end{array}$\end{adjustbox}
   \caption{Labeled Causal Consistency \rev{Specification}\del{Spec}. 
   Causal consistency — but per topic. Every put is tagged with a label (think: Kafka topic, Slack channel, pub/sub subject). Every client declares which labels it cares about. Causal delivery for a client is enforced only for labels the client is interested in.}
   \label{fig:lcc-spec}
\end{figure}

\clearpage
\begin{figure}
   \figurealgtextsize
   \setlength{\abovedisplayskip}{0pt}%
   \setlength{\abovedisplayshortskip}{0pt}%
   \setlength{\belowdisplayskip}{0pt}%
   \setlength{\belowdisplayshortskip}{0pt}%
   \begin{adjustbox}{max width=\linewidth,center}$\begin{array}{|lr|}
      \hline
         \multicolumn{2}{|l|}{𝐈_{\CC_1} \ (\name{Causal Consistency Implementation 1})}
      \\ \hline
         T ≔ \Nat
         & \note{Timestamp}
      \\
         \CState ≔ 
         & \bnote{Client State}
      \\ \t
         C ↦ T
         & \note{Session vector-clock}
      \\
         \cInit(c) ≔
         & \bnote{Client Initial State}
      \\ \t
         [\overline{ c ↦ 0 }]
         & \note{No initial dependencies}
      \\
         \RState ≔
         & \bnote{Replica State}
      \\ \t
         〈K ↦ V, \
         & \note{Store}  %
      \\ \t \phantom{〈}
          C ↦ T〉          
         & \note{Received vector-clock (one for all keys)}
      \\
         \rInit(r, v₀) ≔ 
         & \bnote{Replica Initial State}
      \\ \t
         〈[\overline{ k ↦ v₀}], [\overline{ c ↦ 0 }]〉
         & \note{Empty initial dependencies}
      \\
      & \\
         \GetReqPayload ≔ C ↦ T
         & \note{Get payload types}
      \\
         \GetResPayload ≔ C ↦ T
      & \\
         \getReq \ (\_) (\_, d)  ≔ 
         & \note{At client: } \bnote{Get Request}
      \\ \t
         \ret〈d, d〉
         & \note{Session vector-clock (1) sent as request payload (2) remains unchanged}
      \\
         \getGuard \ (k)(d)(\_, \_, ς) ≔
         & \note{At replica: } \bnote{Get Guard}
      \\ \t
         \llet〈\_, r〉← ς
      & \\ \t
         \ret r ≥ d
         & \note{Pointwise comparison of vector clocks. Store is ahead of session}
      \\
         \get \ (k) (\_) (\_, \_, ς)  ≔
         & \note{At replica: } \bnote{Get}
      \\ \t
         \llet〈s, r〉← ς
      & \\ \t
         \llet v← s (k)
      & \\ \t
         \ret〈 v, r, ς 〉
         & \note{}
      \\
         \getRes \ (\_, \_) (r) (\_, d) ≔
         & \note{At client: } \bnote{Get Response}
      \\ \t
         \ret \max(d, r)
         & \note{Pointwise max of vector-clocks}
      \\
      & \\
         \PutReqPayload ≔ C ↦ T
         & \note{Put payload types}
      \\
         \putReq \ (\_, \_) (\self, d) ≔ 
         & \note{At client: }\bnote{Put Request}
         \\ \t         
         \llet d' ← d [ \self ↦ d(\self) + 1 ]
         & \note{Session dependencies advanced.}
      \\ \t
         \ret〈d, d'〉
         & \note{Session vector-clock is sent as request payload.}
      \\
         \putGuard \ (\_, \_) (d) (c, \_, ς) ≔ 
         & \note{At replica: } \bnote{Put Guard}
      \\ \t
         \llet〈\_, r〉← ς
      & \\ \t
         \ret r ≥ d \, ∧ \, r(c) = d(c)
         & \note{Store satisfies the dependencies. The new put advances the store.}
      \\
         \lput \ (k, v) (d) (c, \_, ς) ≔ 
         & \note{At replica: } \bnote{Put}
      \\ \t
         \llet〈s, r〉← ς
      & \\ \t
         \llet r' ← r[c ↦ d(c) + 1]
         & \note{Received advanced}
      \\ \t    
         s' ← s [ k ↦ v ]
      & \\ \t
         \ret 〈s', r'〉
      & \\ \hline
   \end{array}$\end{adjustbox}
      \caption{Causal Consistency Implementation 1
      (Inspired by \cite{AhamadNeiger95}).}
   \label{fig:causal-map-95}
\end{figure}

\clearpage
\begin{figure}
   \figurealgtextsize
   \setlength{\abovedisplayskip}{0pt}%
   \setlength{\abovedisplayshortskip}{0pt}%
   \setlength{\belowdisplayskip}{0pt}%
   \setlength{\belowdisplayshortskip}{0pt}%
   \begin{adjustbox}{max width=\linewidth,center}$\begin{array}{|lr|}
      \hline
            \multicolumn{2}{|l|}{𝐈_{\CC_2}  \ (\name{Causal Consistency Implementation 2})}
      \\ \hline
         T ≔ \Nat
         & \note{Timestamp}
      \\
         \CState ≔ 
         & \bnote{Client State}
      \\ \t
         T × \List [C × T]
         & \note{Put timestamp, and Session dependencies}
      \\
         \cInit(c) ≔
         & \bnote{Client Initial State}
      \\ \t
         〈0, ∅〉
         & \note{Timestamp zero. No initial dependencies.}
      \\
         \RState ≔
         & \bnote{Replica State}
      \\ \t
         〈K ↦ C × T × V, \
         & \note{Store ($C × T$ uniquely identifies puts.)}
      \\ \t \phantom{〈}
          C ↦ T〉          
         & \note{Received vector-clock}
      \\
         \rInit (r, v₀) ≔ 
         & \bnote{Replica Initial State}
      \\ \t
         〈[\overline{ k ↦ 〈c₀, 0, v₀〉}], [\overline{ c ↦ 0 }]〉
         & \note{Nothing initially received}
      \\
      & \\
         \GetReqPayload ≔ \List [C × T]
         & \note{Get payload types}
      \\
         \GetResPayload ≔ C × T
      & \\
         \getReq \ (\_) (\_, σ)  ≔ 
         & \note{At client: } \bnote{Get Request}
      \\ \t
         \llet 〈\_, d〉← σ
      & \\ \t
         \ret〈d, σ〉
         & \note{Session dependencies sent as request payload}
      \\
         \getGuard \ (\_) (d) (\_, \_, ς) ≔
         & \note{At replica: } \bnote{Get Guard}
      \\ \t
         \llet〈\_, r〉← ς
      & \\ \t \ret \lforall (λ (c, t). \
            r(c) ≥ t)
            \ d
            & \note{Store satisfies the dependencies}
      \\
         \get \ (k) (\_) (\_, \_, ς)  ≔
         & \note{At replica: } \bnote{Get}
      \\ \t
         \llet〈s, \_〉← ς
      & \\ \t
         \llet〈c, t, v〉← \rev{s}\del{ς} (k)
      & \\ \t
         \ret〈 v, 〈c, t〉, ς 〉
         & \note{}
      \\
         \getRes \ (\_, \_) (〈c, t〉) (\_, σ) ≔
         & \note{At client: } \bnote{Get Response}
      \\ \t
         \llet 〈t', d〉 ← σ 
         & \note{}
      \\ \t
         \llet d' ← d :: 〈c, t〉
         & \note{}
      \\ \t
         \ret〈 t', d' 〉
         & \note{}
      \\
      & \\
         \PutReqPayload ≔ \List [C × T]
         & \note{Put payload types}
      \\
         \putReq \ (\_, \_) (c, σ) ≔
         & \note{At client: }\bnote{Put Request}
         \\ \t
         \llet〈t, d〉← σ 
         & \note{}
      \\ \t
         \llet t' ← t + 1
         & \note{}
      \\ \t
         \llet d' ← [〈c, t'〉]
         & \note{} 
      \\ \t
         \ret〈 d, 〈t', d'〉 〉
         & \note{Session dependencies sent as request payload.}
      \\
         \putGuard \ (\_, \_) (d) (c, \_, ς) ≔ 
         & \note{At replica: } \bnote{Put Guard}
      \\ \t
         \llet〈\_, r〉← ς
         & 
      \\ \t \ret \lforall (λ (c, t). \
         r(c) ≥ t)
         \ d \ \rev{∧}
         & \note{Store satisfies the dependencies}
      \\ \t \phantom{\ret}
         \llookup (c, d, 0) = r(c)
         & \note{The new put advances the store} 
      \\
         \lput \ (k, v) (\_) (c, \_, ς) ≔ 
         & \note{At replica: } \bnote{Put}
      \\ \t
         \llet〈s, r〉← ς
         & \note{}
      \\ \t
         \llet t ← r(c) + 1
         & \note{Current put timestamp}
      \\ \t
         \llet r' ← r [ c ↦ t ]
         & \note{}
      \\ \t    
         s' ← s [k ↦ 〈c, t, v〉]
      & \\ \t
         \ret 〈s', r'〉
      & \\ \hline
   \end{array}$\end{adjustbox}
      \caption{Causal Consistency Implementation 2
      (Inspired by \cite{LloydFreedman13_2}).}
   \label{fig:LloydFreedman-protocol}
\end{figure}

\subsection{Chapar Causal Consistency}
For Chapar CC we use Chapar's published causal-consistency specification~\cite{chapar}; we refer the reader to the original paper for its formal definition. Per-operation cost comparison appears in \cref{app:perf}.

\end{document}